\documentclass[
 twocolumn,
 superscriptaddress,
 nofootinbib,
 amsmath,
 amssymb,
 aps,
]{revtex4}

\usepackage{graphicx} 
\usepackage{dcolumn} 
\usepackage{bm} 

\usepackage{hyperref} 

\usepackage{xcolor}
\allowdisplaybreaks

\usepackage{mathtools}

\usepackage[whole]{bxcjkjatype}
\usepackage[linesnumbered, ruled, vlined]{algorithm2e}

\makeatletter
\newcommand{\vast}{\bBigg@{3.0}}
\newcommand{\Vast}{\bBigg@{4.0}}

\makeatother

\usepackage{coffeestains}

\usepackage{xr}
\begin{document}

\preprint{APS/123-QED}

\title{Berezinskii--Kosterlitz--Thouless transition in a context-sensitive random language model}

\author{Yuma Toji}

\affiliation{
	Graduate School of Information Science and Technology,
	Hokkaido University, Sapporo, Hokkaido 060-0814, Japan
}


\author{Jun Takahashi}

\affiliation{
	Institute for Solid State Physics, The University of Tokyo, 5-1-5 Kashiwanoha, Kashiwa, Chiba 277-8581, Japan
}


\author{Vwani Roychowdhury}


\affiliation{
	Henry Samueli School of Engineering and Applied Science, \\
	University of California, Los Angeles, California 90095
}

\author{Hideyuki Miyahara}

\email{miyahara@ist.hokudai.ac.jp, hmiyahara512@gmail.com}

\thanks{corresponding author.}

\affiliation{
	Graduate School of Information Science and Technology,
	Hokkaido University, Sapporo, Hokkaido 060-0814, Japan
}

\date{\today}

\begin{abstract}
	Several power-law critical properties involving different statistics in natural languages---reminiscent of scaling properties of physical systems at or near phase transitions---have been documented for decades.
	The recent rise of large language models has added further evidence and excitement by providing intriguing similarities with notions in physics such as scaling laws and emergent abilities.
	However, specific instances of classes of generative language models that exhibit phase transitions, as understood by the statistical physics community, are lacking.
	In this work, inspired by the one-dimensional Potts model in statistical physics, we construct a simple probabilistic language model that falls under the class of context-sensitive grammars, which we call the context-sensitive random language model, and numerically demonstrate an unambiguous phase transition in the framework of a natural language model.
	We explicitly show that a precisely defined order parameter---that captures symbol frequency biases in the sentences generated by the language model---changes from strictly zero to a strictly nonzero value (in the infinite-length limit of sentences), implying a mathematical singularity arising when tuning the parameter of the stochastic language model we consider.
	Furthermore, we identify the phase transition as a variant of the Berezinskii--Kosterlitz--Thouless (BKT) transition, which is known to exhibit critical properties not only at the transition point but also in the entire phase.
	This finding leads to the possibility that critical properties in natural languages may not require careful fine-tuning nor self-organized criticality, but are generically explained by the underlying connection between language structures and the BKT phases.
	From a statistical physics perspective, our paper captures a rare phenomenon of the BKT phase in a one-dimensional system, as the BKT phase is widely studied and believed to exist only in two-dimensional systems.
\end{abstract}


\maketitle


\section{Introduction}

Natural languages exhibit striking statistical regularities, such as Zipf's law, which describes a power-law relationship between word frequencies and their rank~\cite{Piantadosi_001}.
Similarly, it has been shown that the number of bits of information provided by a symbol about another symbol drops roughly as a power law as a function of their distance~\cite{Lin_001}.
These critical properties, observed across diverse languages, have inspired explanations rooted in statistical physics, drawing parallels between linguistic phenomena and phase transitions.
For example, the scaling behavior inherent in Zipf's law suggests underlying mechanisms that resemble those driving critical phenomena in physical systems~\cite{Simkin_001}.

Recent advances in generative language models, including large language models (LLMs)~\cite{OpenAI_001, Google_001, Naveed_001, Hadi_001}, have further highlighted intriguing connections between language and physics.
Notably, studies have uncovered scaling laws and emergent capabilities in these models, which apparently become evident as the system size increases~\cite{Kaplan_001, Wei_001, Chen_001}.
These findings have prompted the question of whether the statistical properties of natural languages and the behavior of their generative models can be understood as manifestations of genuine phase transitions in physical terms.

These observed phenomena in LLMs, combined with the fact that they become apparent when some parameters of the system are large enough (network depth, width, training data set size, etc.), evoke parallels to phase transitions in the field of statistical physics.
The fact that in critical (continuous) phase transitions many quantities, such as the correlation function, exhibit power-law scaling also fuels the approach of studying these emergent phenomena in languages from a statistical physics perspective.
The notion of phase transitions in the thermodynamic limit, i.e., in an idealized limit where some parameter governing the system size tends to infinity, enables us to study qualitative differences in the system in a rigorous way.

Such a formal qualitative study is essential because once we have a parameterized language model, it will always have a ``trivial regime'' corresponding to a trivial distribution of characters/words/tokens, and an objectively defined hallmark of nontriviality---such as phase transitions---will be a theoretically powerful tool to distinguish meaningful sentences from those generated in the trivial regimes.
Just as there is a clear and mathematically well-defined boundary between water and ice, the existence of a phase transition in language models could lead us to understand the intrinsic difference separating, e.g., a language-like structure that actually conveys information versus generated text that is just statistical gibberish.

\begin{figure*}[t]
	\centering
	\includegraphics[scale=0.35]{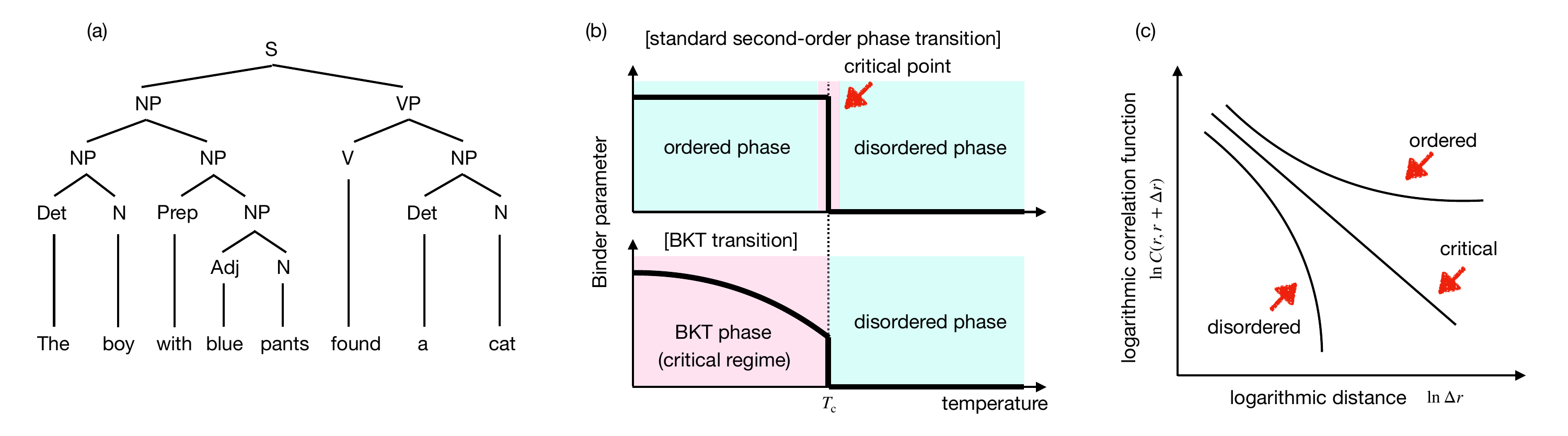}
	\caption{(a) Schematic of a CFG. This diagram shows the syntactic structure of ``The boy with blue pants found a cat." Natural languages are, however, context-sensitive and in this paper we define a simple CSG that demonstrates a phase transition. (b) We consider the relative frequencies of letters or symbols in the alphabet---as captured by the net magnetization of the corresponding physical system---as an order parameter in our CSG. The Binder parameter, which is the normalized kurtosis of the underlying order parameter [see Eq.~\ref{main_eq_def_Binder_parameter_001_001}], can capture phase transitions effectively, especially for one-dimensional models. A schematic of the Binder parameter is shown for the standard second-order phase transition (upper) and the BKT transition (lower). For the BKT transition the Binder parameter's dependence on temperature---which determines the probability with which a rule is applied in the underlying CSG---does not become a step function in the thermodynamic limit. (c) Correlation functions for the disordered, critical, and ordered phases. When a system is critical, the correlation function shows a polynomial decay.}
	\label{main_fig_schematic_phase_transition_001_001}
\end{figure*}

The potential existence of a spin-glass-like phase transition in probabilistic context-free grammars (CFGs) has recently attracted attention~\cite{De-Giuli_001, De-Giuli_002} [see Fig.~\ref{main_fig_schematic_phase_transition_001_001}(a) for a schematic diagram of CFGs].
However, detailed analyses of CFGs have revealed the absence of such claimed phase transitions in some naturally assumed thermodynamic limits~\cite{Nakaishi_001}.
While consensus is yet to be reached on this matter~\cite{Lalegani_001, Nakaishi_002}, it would be fair to say that numerical and analytical support for a true phase transition, i.e., a mathematically singular behavior in a clearly defined thermodynamic limit, is lacking.
Furthermore, as a probabilistic CFG is one of the simplest probabilistic models of natural language, this conundrum with the existence of phase transitions naturally raises the question of how complex a natural language model needs to be to exhibit phase transitions.

Phase transitions in statistical physics are precisely defined mathematical concepts that require careful analysis.
We provide a schematic figure of the types of phase transitions investigated in this paper in Fig.~\ref{main_fig_schematic_phase_transition_001_001}(b).
In particular, to demonstrate a phase transition, one should be able to construct an {\it order parameter}, which is strictly zero in the disordered phase and is nonzero in the ordered phase, thus allowing a rigorous characterization of different phases in a system.
This switch from zero to nonzero is inevitably accompanied by a mathematical singularity (a divergence or discontinuity in the derivative, etc.) of the order parameter.
Note that these features are only realized in the thermodynamic limit.
In practice, one can only observe finite-size systems and extrapolate the data to see consistency with the existence of phase transitions in the thermodynamic limit.
In our paper, we will consider the relative frequencies of letters in the alphabet as the order parameter, and explore whether it shows a second-order phase transition, where the derivative of the order parameter has a singularity as a function of temperature.
Temperature in the context of a language model represents the probability with which a particular rule is applied in text generation.

There are, however, different types of second-order phase transitions.
In a conventional transition, the critical behavior characterized by power-law scaling phenomena emerges only at the point of the phase transition.
That is, the system parameters have to be accurately tuned---so that the system is very close to the phase transition point---for any such scaling laws to be observed.
Both LLMs and natural languages, however, show robust scaling behaviors without requiring extensive fine-tuning of parameters.
This raises the possibility that for language models to exhibit phase transitions, they might need to go through a different type of phase transition, referred to as the Berezinskii--Kosterlitz--Thouless (BKT)~\cite{Berezinskii_001, Berezinskii_002, Kosterlitz_001, Kosterlitz_004, Kosterlitz_002} transition.
One of the most exceptional features of the BKT phase is that the system becomes critical
throughout the entire phase, and not only at the transition point [see Fig.~\ref{main_fig_schematic_phase_transition_001_001}(b)].
The presence of any BKT transition could explain the robust presence of scaling laws in language models.

The BKT transition, however, is most commonly discussed in two-dimensional systems with continuous symmetries.
However, since language models generate strings of text any reasonable physical modeling would lead to one-dimensional systems.
While the BKT transition is also realizable in long-range interacting one-dimensional systems, such one-dimensional phenomena are less studied and less well known.
One method to determine whether a phase transition is BKT
is to compute the kurtosis of the order parameter, which is referred to as the {\it Binder parameter} in the physics literature.
The Binder parameter also nicely captures phase transitions---showing a singularity in its derivative---while being able to distinguish between different types of phases: it is strictly 0 in the disordered phase, and is exactly 1 in a conventionally ordered phase, while having a nontrivial value (between 0 and 1) at critical points where correlations have power-law decays; see Fig.~\ref{main_fig_schematic_phase_transition_001_001}(b).
The Binder parameter becomes an especially powerful tool for studying our language model since the model turns out to exhibit a type of the BKT phase.

In this study, we construct a new language model that falls under the class of context-sensitive grammars (CSGs), which we call the context-sensitive random language model, by introducing context-sensitivity to the random language model~\cite{De-Giuli_001, De-Giuli_002}.
Our construction is based on the long-range interacting Potts model, and utilizes the additional complexity that CSGs possess from being one level higher in the Chomsky hierarchy~\cite{Jager_001, Deletang_001} compared to CFGs.

CSGs allow the application of a rewrite rule to depend on the symbols surrounding the one being rewritten. 
Such rules naturally introduce effective long-range couplings and capture linguistic phenomena---agreement, feature matching, and structural constraints---that CFGs cannot express. 
In this work, we extract and formalize these essential ingredients to construct a probabilistic, CSG-inspired random language model that is sufficiently expressive to generate extended correlations while remaining amenable to thermodynamic analysis.

The model dynamics combine three interacting processes:
(1) [Growth]
A nonterminal symbol expands via rules of the form $X \to YZ$, increasing string length and enabling a thermodynamic limit as length $\to \infty$. 
A growth parameter $q$ controls the balance between expansion and rewriting, with different regimes giving rise to qualitatively distinct global behaviors.
(2) [Context-sensitive rewrites with Metropolis acceptance]
A substring $\alpha_{-} X \alpha_{+}$ may be rewritten as $\alpha_{-} Y \alpha_{+}$, with acceptance probability $p = \min (1, \mathrm{e}^{- \frac{\Delta E}{k_\mathrm{B} T}})$ where $\Delta E$ is computed from an energy functional coupling symbol pairs at distance $|i-j|$ through a long-range kernel $|i-j|^{-(1+s)}$. This structure is analogous to a one-dimensional long-range Potts model but with crucial differences: the domain itself grows, interactions depend on context, and the rewriting operations are not simple spin flips but grammar-inspired substitutions.
(3) [Transition to terminal symbols] 
A nonterminal symbol may transition into a terminal symbol, after which it no longer grows or undergoes rewriting.
Note that the notion of nonterminal symbols has no direct counterpart in physics and is specific to language models.
For simplicity, we neglect this process in the present study; as a result, we can take the thermodynamic limit straightforwardly.

Languages have a one-dimensional structure naturally associated with them: namely, the direction from the beginning to the end of a sentence.
Together with the observation that the degrees of freedom in a sentence are essentially discrete (such as words or tokens), we focus our attention on the archetypal toy model in statistical physics with discrete degrees of freedom: the one-dimensional Potts model.
We focus on the order parameter that captures the relative frequency of the letters or symbols in the string.
We also compute the higher-order cumulants (variance and kurtosis), which correspond to the susceptibility and Binder parameter from a physics perspective, and are the most widely used measures for detecting phase transitions.
From the perspective of language models, e.g., the variance/susceptibility corresponds to how diverse the produced strings are, and therefore could be viewed as a measure of the average nontriviality of the sentences.
We compute the system-size dependence of the susceptibility and perform the finite-size scaling method~\cite{Ardourel_001, Hsieh_001, Zuo_001, Ueda_001}, which allows us to confidently conclude the existence of a phase transition in the thermodynamic (infinite-size) limit.
For example, we can observe a clear divergence of the susceptibility in the low-temperature BKT phase in our model, which indicates an entire parameter range where nontrivially correlated strings are produced.

In our numerical simulations, we first compute the system-size dependence of the quantity corresponding to magnetization.
This quantifies how much a specific symbol is favored in the generating process of a sentence.
We also compute the corresponding susceptibility and the Binder parameter, which play an important role in distinguishing between the standard second-order phase transition and the BKT transition~\cite{Hasenbusch_001, Tuan_001}.
We then perform the finite-size scaling method to confirm the existence of phase transitions.
In addition to these, we also investigate the correlation functions~\footnote{In addition to the correlation function, we also compute mutual information in Appendix~\ref{supp_sec_numerical_simulations_001_001}.} between symbols~\cite{Lin_001}.
These quantities play an essential role in discussing the scaling law in at least natural languages
as the power-law scaling in correlation functions serves as the smoking gun for critical phase transitions in physics [see Fig.~\ref{main_fig_schematic_phase_transition_001_001}(c) for correlation functions]~\footnote{Furthermore, we analyze the histogram of the magnetization and show typical configurations of symbols for parameters at which the histogram has a peak and relative frequencies of letters in the alphabet, which are often discussed in the context of natural language processing in Appendix~\ref{supp_sec_numerical_simulations_001_001}.}.

It should be noted that the context-sensitive random language model is {\it not} a mere translation of the physical Potts model into a language model setting.
The unique property of language models in general, absent in standard statistical mechanics models, is the increase in symbols over time, and there is an additional degree of freedom in how the symbols grow; see the parameters $q, t$ and the specifics of the $X\to YZ$ rule in Eq.~\ref{main_eq_rule_002_002}.
\textit{These properties of the context-sensitive random language model render the existence of a phase transition nontrivial}, and indeed we find that for certain settings of the language parameters, the phase transition can vanish.
Furthermore, we compare the context-sensitive random language model with multiple alphabets to the one-dimensional long-range Potts model~\cite{Bayong_001}.
As for phase transitions in one-dimensional models, it is well known that the Ising model---which is equivalent to the Potts model in the case of an alphabet of size two (i.e., two letters)---in one dimension shows no phase transition with short-range interactions but exhibits phase transitions with long-range interactions~\cite{Dyson_001, Frohlich_001, Dyson_002, Dyson_003, Thouless_001, Chang_001, Martinez-Herrera_001, Tomita_001}.
For the one-dimensional long-range Potts model with two-level spins ($\sigma = \pm 1$), which is also known as the Ising model, it is known to exhibit an ordered phase at low temperatures and to undergo a BKT transition~\cite{Anderson_003, Kosterlitz_003, Kosterlitz_004, Aizenman_001, Aizenman_002}.
However, such a BKT phase is observed when the decay exponent $s$ in the power-law interaction with distance is set to $s = 1$ in the exponent [see Eq.~\ref{main_eq_Delta_E_001_001}].
In our language model, we observe a BKT transition even when $s=0.9$, which is far from the setting studied in Refs.~\cite{Kosterlitz_003, Kosterlitz_004}.
We also observe a BKT transition in our systems with multi-level spins, i.e., alphabets with $K > 2$ letters.
\textit{We hypothesize that our language model, where the system size grows in size following a set of probabilistic rules, behaves differently from the physics models that inspired their conception and BKT transitions might be much more robust in our systems.}

The structure of this paper is as follows.
In Sec.~\ref{main_sec_language_model_001_001}, we describe in detail the language model we constructed, inspired by the Potts model.
Section~\ref{main_sec_methods_001_001} introduces the physical quantities computed to analyze the layer-wise behavior of the language model.
In Sec.~\ref{main_sec_numerical_simulations_001_001}, we present the results for a representative parameter setting, namely the case of $K=2$ and $X \to YZ$.
Based on these results, Sec.~\ref{main_sec_discussions_001_001} discusses the implications of the observed phase transition phenomenon in the context of language models.
Finally, we conclude the paper in Sec.~\ref{main_sec_conclusions_001_001} by summarizing the findings and discussing their relevance to large LLMs.
In the appendices, we describe the numerical setup in detail and show additional numerical simulations, and they are written in a self-contained manner.

\section{Language model} \label{main_sec_language_model_001_001}

We construct a probabilistic language model based on the long-range Potts model, which we call the context-sensitive random language model~\footnote{The details of our model are given in Appendix~\ref{supp_sec_language_model_001_001}.}.
First of all, the generation process of a sentence begins with the start symbol, which is a nonterminal symbol.
Until all the nonterminal symbols (lexical elements that specify the production rules constituting a formal grammar) in the sentence are replaced by terminal symbols (outputs of the production rules), a nonterminal symbol is uniformly randomly chosen and one of the following three processes is applied to it:
\begin{subequations} \label{main_eq_rule_001_001}
	\begin{align}
		X                   & \to x                   & (\text{probability:} & ~qt), \label{main_eq_rule_002_001}        \\
		X                   & \to YZ                  & (\text{probability:} & ~q (1 - t)), \label{main_eq_rule_002_002} \\
		\alpha_- X \alpha_+ & \to \alpha_- Y \alpha_+ & (\text{probability:} & ~(1 - q)), \label{main_eq_rule_002_003}
	\end{align}
\end{subequations}
where uppercase and lowercase letters are nonterminal and terminal symbols, respectively, and $\alpha_-$ and $\alpha_+$ are strings.
To take the thermodynamic limit, we set $t = 0$; in other words, we do not consider Eq.~\eqref{main_eq_rule_002_001}~\footnote{This thermodynamic limit is the simplest one while there can be different definitions of thermodynamic limits, which are beyond our scope.}.
For Eq.~\eqref{main_eq_rule_002_002}, we consider two branching processes: (i) $X \to YZ$, where $Y$ and $Z$ are uniformly randomly chosen, and (ii) $X \to XX$.
(In the main text, we focus on (i); see Appendix~\ref{supp_sec_numerical_simulations_001_001} for (ii).)
Furthermore, when Eq.~\eqref{main_eq_rule_002_003} is chosen, the corresponding update is accepted with the following probability:
\begin{align}
	p & = \min (1, \mathrm{e}^{- \frac{\Delta E}{k_\mathrm{B} T}}), \label{main_eq_acceptance_rate_001_001}
\end{align}
where $\Delta E$ for $\sigma_0 \to \tilde{\sigma}_0$ is given by
\begin{align}
	\Delta E & \coloneqq J \vast( \sum_{\ell=1}^{\lfloor N r_- \rfloor} \frac{\delta_{\sigma_0, \sigma_{-\ell}} - \delta_{\tilde{\sigma}_0, \sigma_{-\ell}}}{\ell^{1 + s}} + \sum_{\ell=1}^{\lfloor N r_+ \rfloor} \frac{\delta_{\sigma_0, \sigma_{\ell}} - \delta_{\tilde{\sigma}_0, \sigma_{\ell}}}{{\ell}^{1 + s}} \vast). \label{main_eq_Delta_E_001_001}
\end{align}
Here, $J > 0$ is a hyperparameter that determines the strength of the interaction, and $\sigma_i \in \{ 0, 1, \dots, K - 1 \}$ is an integer corresponding to the symbol at the $i$-th character in the sentence.
Here, $r_+, r_- \in [0, 1]$ specify the range of interaction.
Furthermore, we consider the open boundary condition; so terms beyond the boundary are ignored.
Note that Eq.~\eqref{main_eq_acceptance_rate_001_001} is the acceptance rate of the Metropolis-Hastings algorithm~\cite{Landau_001, Binney_001, Newman_001}, and Eq.~\eqref{main_eq_Delta_E_001_001} is the energy gap of the one-dimensional long-range Potts model with respect to a one-spin flip.

\section{Methods} \label{main_sec_methods_001_001}

We summarize the physical quantities of interest used to discuss phase transitions~\footnote{The details of the physical quantities computed in this paper are given in Appendix~\ref{supp_sec_methods_001_001}.}.
First, we begin with the magnetization, which is the order parameter, and it reads
\begin{align}
	M & \coloneqq \| \bm{M} \|, \label{main_eq_def_magnetization_001_001}
\end{align}
where $\bm{M} \in \mathbb{R}^{K-1}$ is given by
\begin{align}
	\bm{M} & \coloneqq \frac{1}{N} \sum_{i = 1}^N \bm{e}_{\sigma_i}.
\end{align}
Here, $\{ \bm{e}_k \}_{k=1}^K$ satisfies $\| \bm{e}_k \| = 1$ for $k = 1, 2, \dots, K$, $\bm{e}_k \cdot \bm{e}_l = \mathrm{const}.$ for $k \ne l$, and $\sum_{k=1}^K \bm{e}_k = \bm{0}$~\footnote{See Appendix~\ref{supp_sec_magnetization_001_001} for the derivation of $\{ \bm{e}_k \}_{k=1}^K$.}.
Note that $\| \bm{M} \| \coloneqq \sqrt{\bm{M} \cdot \bm{M}}$.

We then define the magnetic susceptibility by
\begin{align}
	\chi & \coloneqq N (\langle M^2 \rangle - \langle M \rangle^2). \label{main_eq_def_specific_heat_001_001}
\end{align}
For later convenience, we also define
\begin{align}
	\tilde{\chi} & \coloneqq N \langle M^2 \rangle. \label{main_eq_def_specific_heat_001_002}
\end{align}

As a well-used method to discern phase transitions, the Binder parameter is often used, and it reads
\begin{align}
	U & \coloneqq - \frac{K - 1}{2} \bigg( \frac{\langle M^4 \rangle}{\langle M^2 \rangle^2} - \frac{K+1}{K-1} \bigg). \label{main_eq_def_Binder_parameter_001_001}
\end{align}
From the viewpoint of statistics, the Binder parameter, Eq.~\eqref{main_eq_def_Binder_parameter_001_001}, is the normalized kurtosis, which becomes zero for the normal distribution.
Thus, it can be used to distinguish the disordered phase, in which the \textit{generalized central limit theorem} empirically holds, and the ordered phase.

The correlation function is given by
\begin{align}
	\tilde{G} (i, j) & \coloneqq \langle \bm{e}_{\sigma_i} \cdot \bm{e}_{\sigma_j} \rangle. \label{main_eq_correlation_function_with_disconnected_diagram_Potts_001_002}
\end{align}
It is well known that in a critical regime, the correlation function exhibits a polynomial decay, while it shows an exponential decay outside a critical regime.

Finite-size scaling~\footnote{See Appendix~\ref{supp_sec_methods_002_011} for finite-size scaling.} is one of the most popular methods to discuss the existence of phase transitions.
By applying the scaling assumption to $\tilde{\chi}$, Eq.~\eqref{main_eq_def_specific_heat_001_002}, we obtain
\begin{align}
	\frac{\tilde{\chi} (T, N)}{N^\frac{\gamma}{\nu}} & = \tilde{f}_{\tilde{\chi}} (N^\frac{1}{\nu} t). \label{main_eq_finite_size_scaling_tilde_chi_001_001}
\end{align}
Note that the scaling assumption implies the existence of a universal function $\tilde{f}_A (\cdot)$ for any thermodynamic quantity $A$.
We will confirm Eq.~\eqref{main_eq_finite_size_scaling_tilde_chi_001_001} via numerical simulations.

\section{Numerical simulations} \label{main_sec_numerical_simulations_001_001}

We show numerical simulations restricting ourselves to the case of $K = 2$ and $X \to YZ$~\footnote{Numerical simulations for other setups are given in Appendix~\ref{supp_sec_numerical_simulations_001_001}.}.
We run the Monte Carlo simulations $80 \times 80$ times to compute expected values and standard errors.

\begin{figure}[t]
	\centering
	\includegraphics[scale=0.60]{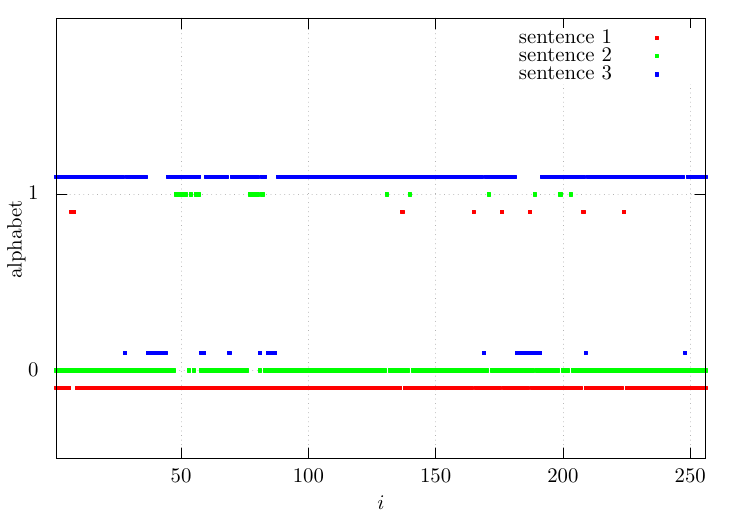}
	\includegraphics[scale=0.60]{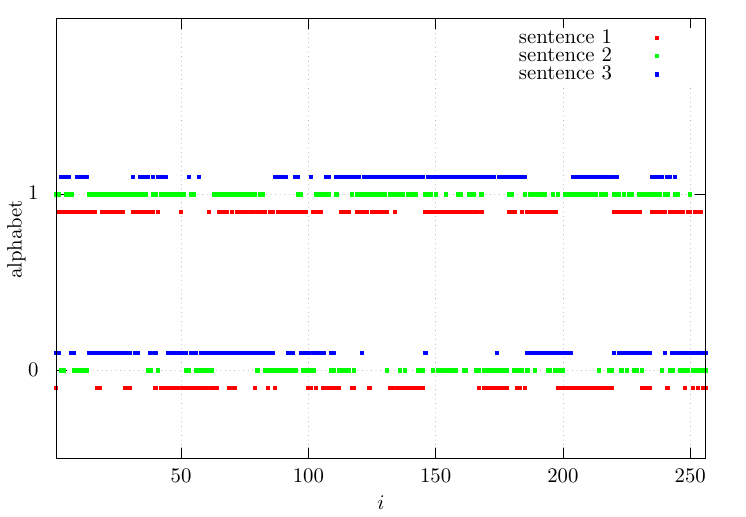}
	\caption{Typical configurations of symbols for $X \to YZ$. We plot (upper) states whose magnetization lies between $0.9000$ and $0.9200$ at $k_\mathrm{B} T = 0.760$, and (lower) states whose magnetization lies between $0.0000$ and $0.0200$ at $k_\mathrm{B} T = 1.160$. We set $K = 2$, $J = 1.0$, $q = 10^{-2.0}$, $t = 0$, $s = 0.9$, and $r_- = r_+ = 0.25$. We plot the first 256 symbols out of 4096. Note that the estimated critical temperature is $k_\mathrm{B} T_* = 0.960$.}
	\label{main_fig_configurations_K=2_q_10_-2.0_X_YZ_001_001}
\end{figure}

We first plot the first 256 symbols of typical configurations of symbols in Fig.~\ref{main_fig_configurations_K=2_q_10_-2.0_X_YZ_001_001}.
Here, ``typical'' configurations refer to those whose magnetization corresponds to the peak of the magnetization histograms~\footnote{The histogram of magnetization for this parameter set is shown in Fig.~\ref{supp_fig_histogram_mag_K=02_q=10^(-2.0)_X_YZ_001_001}.}.
As shown in the figure, typical configurations of letters in the alphabet depend on the temperature.
Hereafter, we investigate the statistical properties of these sentences.

\begin{figure}[t]
	\centering
	\includegraphics[scale=0.60]{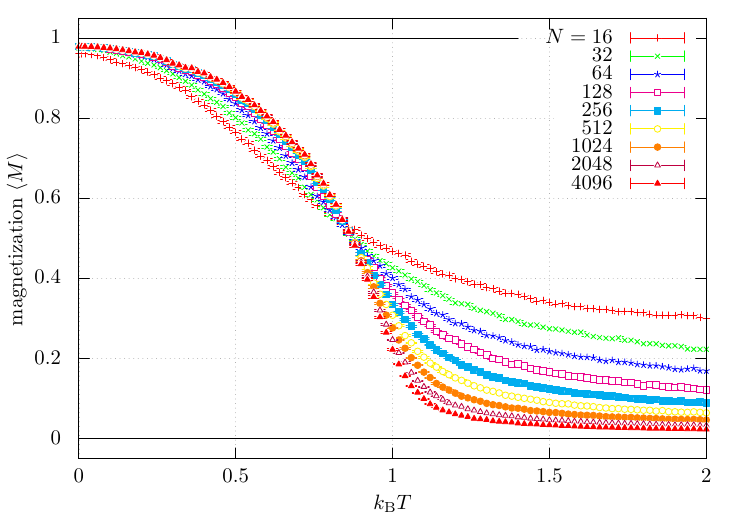}
	\includegraphics[scale=0.60]{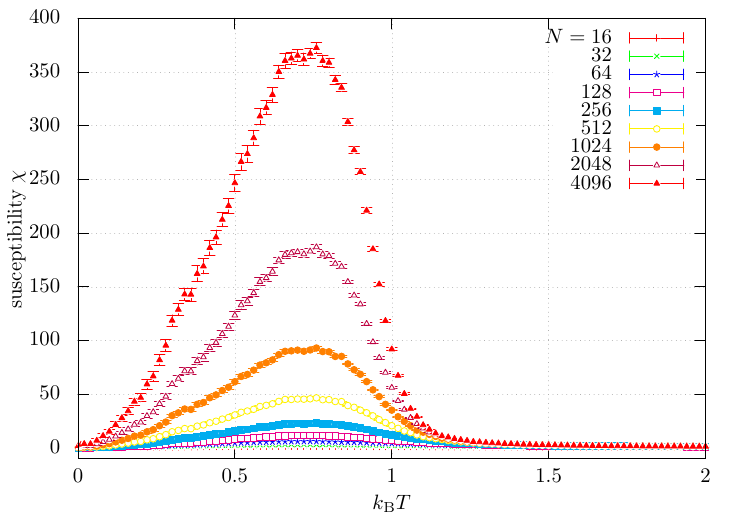}
	\includegraphics[scale=0.60]{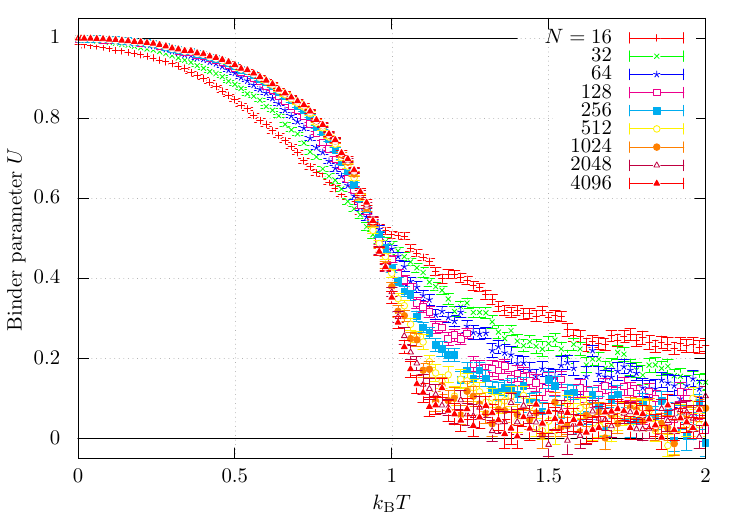}
	\caption{Temperature dependence of (upper) the magnetization, Eq.~\eqref{main_eq_def_magnetization_001_001}, (middle) the susceptibility, Eq.~\eqref{main_eq_def_specific_heat_001_001}, and (lower) the Binder parameter, Eq.~\eqref{main_eq_def_Binder_parameter_001_001}, for $X \to YZ$. We set $q = 10^{-2.0}$. We also set $K = 2$, $J = 1.0$, $t = 0$, $s = 0.9$, and $r_- = r_+ = 0.25$. The length of the generated sentence by the language model, $N$, was varied from $16$ to $4096$.}
	\label{main_fig_magnetization_K=2_q_10_-2.0_X_YZ_001_001}
\end{figure}

In Fig.~\ref{main_fig_magnetization_K=2_q_10_-2.0_X_YZ_001_001}, we plot the temperature dependence of the magnetization, Eq.~\eqref{main_eq_def_magnetization_001_001}, the susceptibility, Eq.~\eqref{main_eq_def_specific_heat_001_001}, and the Binder parameter, Eq.~\eqref{main_eq_def_Binder_parameter_001_001}.
First of all, for $k_\mathrm{B} T \lesssim 1$, the magnetization has a nonzero value ($\langle M \rangle > 0$), and the system seems to be ordered.
In the same regime, the susceptibility becomes not merely large but also divergent with respect to the system size, for $k_\mathrm{B} T \lesssim 1$, unlike in the second-order phase transition.
In other words, the system is critical for a parameter regime of nonzero measure, and this fact implies that the system is in the BKT phase.
The Binder parameter also has a crossing point when the system size is varied, which also supports the existence of a phase transition.

\begin{figure}[t]
	\centering
	\includegraphics[scale=0.60]{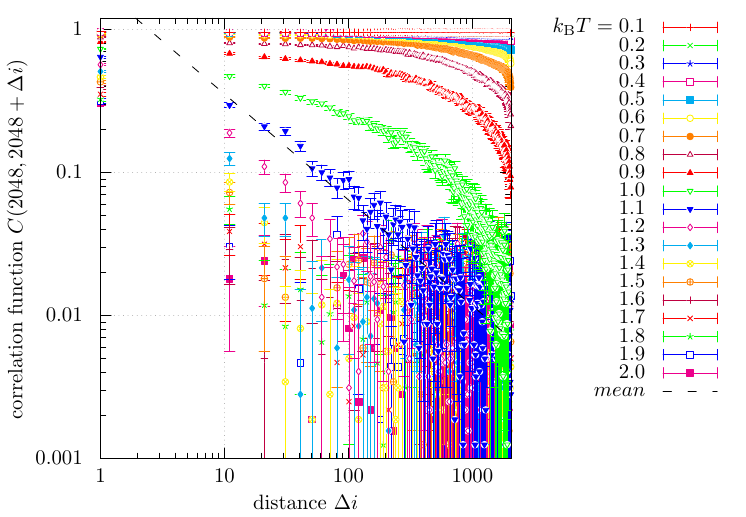}
	\includegraphics[scale=0.60]{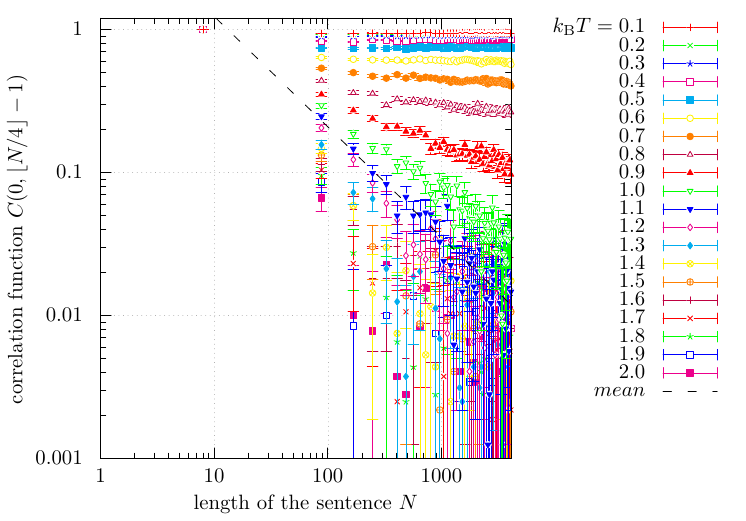}
	\caption{Correlation functions, Eq.~\eqref{main_eq_correlation_function_with_disconnected_diagram_Potts_001_002}, (upper) with $i=2048$ and $j=2048 + \Delta i$ and (lower) with $i = 0$ and $j = \lfloor N / 4 \rfloor - 1$ for $X \to YZ$. We set $K = 2$, $J = 1.0$, $q = 10^{-2.0}$, $t = 0$, $s = 0.9$, and $r_- = r_+ = 0.25$. Temperature $k_\mathrm{B} T$ was varied from $0.1$ to $2.0$. The dashed lines are fitted lines to the data for (upper) $\Delta i \in [10, 1000]$ and (lower) $N \in [100, 1000]$ at $k_\mathrm{B} T = 1.1$.}
	\label{main_fig_correlation_function_K=02_q=10^(-2.0)_X_YZ_001_001}
\end{figure}

In Fig.~\ref{main_fig_correlation_function_K=02_q=10^(-2.0)_X_YZ_001_001}, we plot the correlation functions for different conditions.
For a fixed system size, the correlation function looks straight for $k_\mathrm{B} T \sim 1.1$ [see Fig.~\ref{main_fig_correlation_function_K=02_q=10^(-2.0)_X_YZ_001_001}(upper)].
Then, we turn our attention to the system-size dependence of the correlation function.
As Fig.~\ref{main_fig_correlation_function_K=02_q=10^(-2.0)_X_YZ_001_001}(lower) depicts, the correlation functions become straight for $k_\mathrm{B} T \lesssim 1$.

\begin{figure}[t]
	\centering
	\includegraphics[scale=0.60]{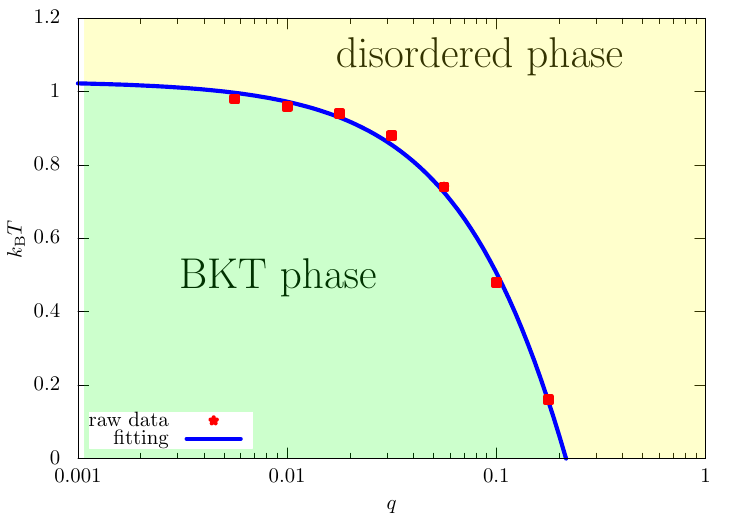}
	\includegraphics[scale=0.60]{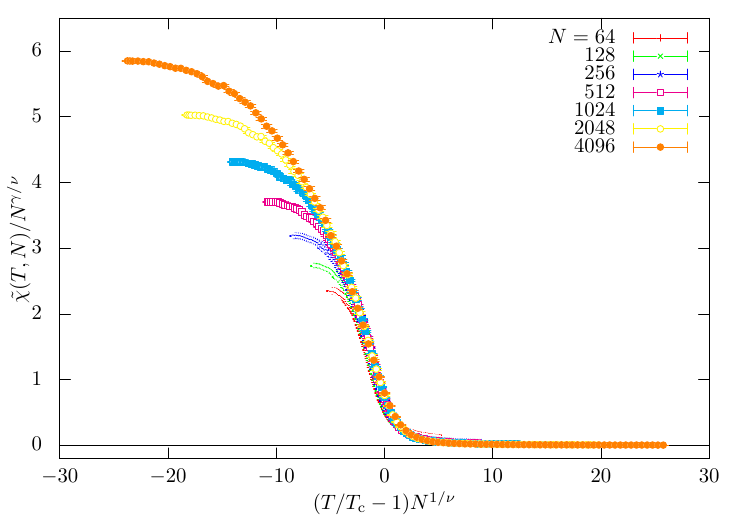}
	\includegraphics[scale=0.60]{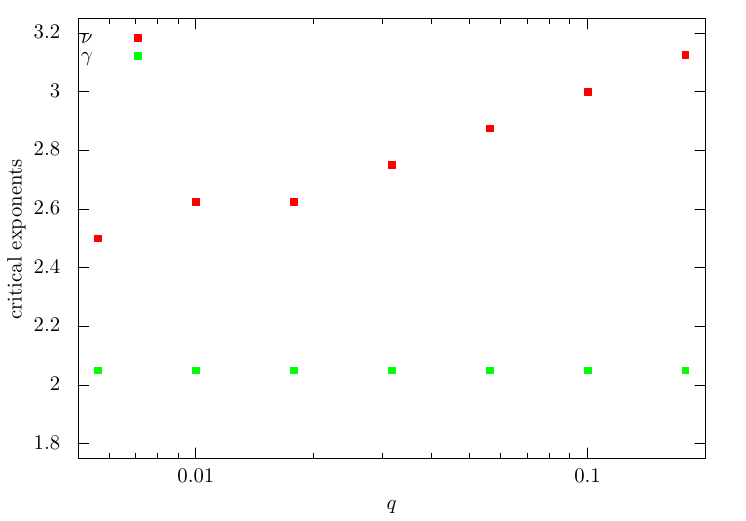}
	\caption{(upper) Phase diagram of the context-sensitive random language model, where the horizontal and vertical axes are the growth rate of a sentence $q$ and critical temperature $k_\mathrm{B} T$, respectively. We consider $X \to YZ$ and set $K = 2$, $J = 1.0$, $t = 0$, $s = 0.9$, and $r_- = r_+ = 0.25$. (middle) Finite-size scaling of $\tilde{\chi}$ at $q = 10^{-2.0}$. We set $T_\mathrm{c} = 0.960$, $\nu = 2.6250$, and $\gamma = 2.05$, where the values of $\nu$ and $\gamma$ are determined such that the scaling assumption holds. We varied $N = 64, 128, 256, 512, 1024, 2048, 4096$. (lower) $q$-dependence of the critical exponents $\nu$ and $\gamma$.}
	\label{main_fig_phase_diagram_001_001}
\end{figure}

In Fig.~\ref{main_fig_phase_diagram_001_001}(upper), we draw the phase diagram of the context-sensitive random language model with $K = 2$ and $X \to YZ$.
In this phase diagram, $T_\mathrm{c}$, below which the susceptibility is divergent with respect to the system size, is plotted by red dots, and the fitting curve is computed via least squares regression with a quadratic function.
In other words, the susceptibility is not divergent with respect to the system size above $T_\mathrm{c}$.
We conduct finite-size scaling on $\tilde{\chi}$ and plot the result in Fig.~\ref{main_fig_phase_diagram_001_001}(middle).
Finally, we show the $q$-dependence of the critical exponents $\nu$ and $\gamma$.
This figure shows that $\nu$ increases with $q$ but $\gamma$ stays unchanged.

\section{Discussions} \label{main_sec_discussions_001_001}

The results presented here demonstrate that a simple, context-sensitive generative language model---combining growth dynamics with long-range context-dependent interactions---can exhibit a BKT transition. 
This finding is notable for several reasons. 
First, classical discrete one-dimensional spin systems, even with long-range couplings, rarely display BKT behavior; such transitions typically require continuous symmetries or finely tuned interactions. 
In our model, the emergence of BKT-like criticality arises not from finely adjusted symmetry but from the inherent structure of the generative process itself: context sensitivity and expansion jointly modify the effective dimensionality of the system. 
This suggests that linguistic generative processes lie naturally outside the universality classes usually invoked for one-dimensional discrete models.

Second, the BKT transition observed here provides a principled explanation for why scaling phenomena in language appear robust across corpora, languages, and even across artificial generative systems such as LLMs. 
In BKT phases, scale-invariant behavior persists across a finite region rather than at a finely tuned critical point. This is consistent with the empirical observation that natural languages display power-law statistics without requiring external tuning, and that LLMs exhibit predictable scaling laws across multiple orders of magnitude. 
The present model shows that hierarchical growth, long-range dependencies, and context-aware constraints---all essential features of language---create a natural pathway to such extended criticality.

Third, the model highlights the role of grammar-like mechanisms in organizing long-range structure. 
While neural LLMs implement interactions through attention mechanisms and deep feedforward layers rather than explicit rewrite rules, both systems share the ability to propagate information across long ranges and adapt symbol generation to global context. 
The success of the simplified CSG-based model in generating BKT-like ordering suggests that long-range coherence in LLMs may arise from universal mechanisms rooted in context-sensitive generativity rather than purely neural architectural choices.

We next discuss in greater detail the numerical findings presented in the previous section, starting with the case of $K = 2$.
The identified phase transition, reminiscent of the Ising model, entails the spontaneous breaking of $\mathbb{Z}_2$ symmetry, presenting a unique form of phase transition distinct from those previously discussed in CFGs, as highlighted in prior works such as Refs.~\cite{De-Giuli_001, De-Giuli_002, Nakaishi_001} and similar studies.
Our investigation suggests that the phase transition phenomenon in CSGs is particularly interesting compared to that of CFGs, which has led to the ongoing debate.
Such a result implies that long-range interactions play a pivotal role in inducing a different type of phase transition in language models.

Furthermore, the critical nature of the discovered phase transition is crucial.
When arbitrarily constructing systems that undergo phase transitions, it is common for transitions to be discontinuous (first-order).
However, in such discontinuous transitions, the lack of nontrivial scaling laws makes them unsuitable for explaining phenomena like Zipf's law, which is observed in natural languages.

It is important to note that the rules, such as Eq.~\eqref{main_eq_rule_002_002}, involved in the studied statistical mechanics model are specific to language models and are not present in typical statistical mechanics models where the system does not grow over time.
Therefore, the observed phase transition phenomenon in this study is not merely a repetition of calculations for the one-dimensional long-range Potts model in the context of natural language processing.
The results of this study may offer new insights into the statistical mechanics of systems that spatially expand over time, beyond language-specific frameworks.

It is important to examine what occurs in the case of alphabets with $K > 2$ distinct letters, as this is a feature of any natural language.
Unlike the Potts model, where $K \geq 5$ exhibits a first-order phase transition, we observed the BKT transition in the context-sensitive random language model with $K = 10$.
However, its temperature dependence is heavily influenced by numerical conditions, and the unique processes specific to the context-sensitive random language model, Eq.~\eqref{main_eq_rule_001_001}, contribute to a much richer phase diagram compared to conventional statistical-mechanical models.

We have also computed the correlation functions and mutual information for several numerical setups and observed the scaling laws.
Due to the length-expanding nature of language models, there is some arbitrariness in the definition of correlation functions.
Depending on this definition, the correlation functions exhibit power-law decay.
This suggests the possibility of power laws occurring not only at the exact phase transition point but potentially throughout an entire phase.

The relative frequencies of letters in the alphabet are often studied in natural language processing.
For $K = 10$ and $q = 10^{-2.0}$, we plot the relative frequencies of letters in the alphabet for typical configurations [see Fig.~\ref{supp_fig_relative_frequencies_K=10_q=10^(-2.0)_X_YZ_001_001}].
Multiple symbols appear in sentences in the case of $X \to YZ$, while one or two symbols dominate sentences in the case of $X \to XX$.
In natural languages, Zipf's law is believed to hold; therefore, the case of $X \to YZ$ seems to be more natural, though the phenomenon observed in the case of $X \to XX$ looks quite similar to that of a first-order phase transition.
However, further investigation may be required to reach a definitive conclusion, as there are several different branching processes, including those that exist between $X \to YZ$ and $X \to XX$.

In finite-size scaling, we determined the critical exponents such that the scaling assumption holds.
It is important to note that the existence of such values is non-trivial, as there is no universal function in general.
In Ref.~\cite{Tomita_001}, the critical exponents of the one-dimensional long-range Ising model are reported, but our observation is that the scaling assumption does not hold for these quantities.
Furthermore, $\gamma$ and $\nu$ do not depend on the choice between $X \to YZ$ and $X \to XX$; however, they depend on $q$ and $K$.
The phase diagrams heavily depend on the choice between $X \to YZ$ and $X \to XX$, making the independence of $\gamma$ and $\nu$ from these choices an interesting feature.
The fact that $\gamma$ and $\nu$ depend on $K$ seems natural since the number of states or flavors often affects renormalization flows.
Contrary to the choice between $X \to YZ$ and $X \to XX$, $\gamma$ and $\nu$ depend on $q$, which is a newly introduced parameter that controls the growth rate of a sentence.

The reader may find that the above discussions seem to contradict common understanding regarding the BKT transition.
Typically, the BKT transition is studied within the context of the $XY$ model, where it is believed that $\mathbb{Z}_2$ spontaneous symmetry breaking does not occur~\cite{Berezinskii_001, Berezinskii_002, Kosterlitz_001}.
In contrast, the one-dimensional long-range Ising model is known to exhibit $\mathbb{Z}_2$ spontaneous symmetry breaking~\cite{Anderson_001, Yuval_001, Anderson_002, Hamann_001, Anderson_003} alongside the BKT transition~\cite{Kosterlitz_003, Kosterlitz_004}.
In this sense, our study sheds light on the \textit{unconventional} BKT transition and posits that it may capture the nature of language models.

Additionally, in deep learning models for natural language processing, attention mechanisms are considered crucial~\cite{Vaswani_001}.
These attention mechanisms facilitate long-range interactions, aligning with the results of our study, which suggest that long-range interactions play a role in stabilizing phase transitions in language models.
From the viewpoint of spin glass theory with the Potts model, the attention mechanism is investigated in Ref.~\cite{Rende_001}, but it is yet to be known how it affects phase transitions.
More recently, the possibility of a mirage of the emergent ability of LLMs was pointed out in Ref.~\cite{Schaeffer_001}; thus, a similar investigation of neural language models and LLMs is one of the most important future directions.

It is also worth noting that Refs.~\cite{Cui_001, Cagnetta_001, Cagnetta_002, Garnier-Brun_001, Sclocchi_001} discuss phase transitions in neural language and diffusion models, which are indeed intriguing.
However, the approach differs from ours; while our work focuses on identifying the minimal language model that exhibits phase transitions, Refs.~\cite{Cui_001, Cagnetta_001, Cagnetta_002, Garnier-Brun_001, Sclocchi_001} directly investigate neural language and diffusion models.

\section{Conclusions} \label{main_sec_conclusions_001_001}

In this study, we have constructed the context-sensitive random language model by combining the random language model developed by DeGiuli~\cite{De-Giuli_001, De-Giuli_002} with the one-dimensional long-range Potts model, and we numerically demonstrated the presence and absence of phase transitions depending on numerical conditions.
Moreover, we unveiled nontrivial phase transition phenomena specific to the language model rules.
In the future, it will be intriguing to explore whether more nontrivial phase transitions, as discussed in Refs.~\cite{De-Giuli_001, Nakaishi_001}, can be realized in CSGs or their probabilistic extensions, akin to the transitions observed in this study.
Additionally, by considering more realistic language models, we aim to elucidate the scaling laws observed in LLMs and the mechanisms behind emergent language capabilities.

Indeed, our study has shed light on the significance of long-range interactions in the phase transition of language models.
However, the precise relationship with mechanisms such as attention mechanisms remains ambiguous.
Future research will seek to elucidate the behavior of attention mechanisms in the long-range limit and establish connections between long-range interactions, attention mechanisms, phase transitions, scaling laws, and the emergence of language capabilities.

\begin{acknowledgments}
	H.M. thanks Koji Hukushima, Kai Nakaishi, Sho Yokoi, Ryo Ueda, Tatsuki Kuribayashi, Takashi Mori, Masahito Mochizuki, Yusuke Miyajima, and Masaharu Yoshioka for fruitful discussions.
	J.T. thanks Yoshihiko Nishikawa, K. Nakaishi, K. Hukushima, S. Yokoi, R. Ueda, and T. Kuribayashi for insightful discussions too.
	H.M. was supported by JSPS KAKENHI Grant Number JP25H01499.
\end{acknowledgments}

\section*{Author contributions}

H.M. conceived the original idea.
H.M. and J.T. designed the research.
Y.T. and H.M. performed numerical simulations.
All authors discussed the results.
H.M., J.T., and V.R. wrote the paper.

\section*{Competing interests}

The authors declare no competing financial interests.

\section*{Data availability}

We will provide all the numerical codes and plot files upon request.

\appendix
\renewcommand{\theequation}{\Alph{section}\arabic{equation}}
\renewcommand{\thefigure}{\Alph{section}\arabic{figure}}

\section{Overview of the appendices}

The appendices, which are written in a self-contained manner, provide the supplementary information and are organized as follows.
In Appendix~\ref{supp_sec_language_model_001_001}, we formulate the problem setting of this study.
Appendix~\ref{supp_sec_methods_001_001} describes some physical quantities computed in this paper.
Appendix~\ref{supp_sec_numerical_simulations_001_001} presents supporting numerical simulations.

\section{Language model} \label{supp_sec_language_model_001_001}

This section outlines the model under consideration in this study.
With regard to the model, particular emphasis is placed on explaining the interactions between symbols.
As for the interaction between symbols, we consider the one-dimensional Potts model with a long-range interaction and construct a new language model that falls under the class of CSGs, which we call the context-sensitive random language model.
Note that the Potts model is identical to the Ising model when the number of states that each spin takes is two.
Finally, we explain the context dependency of the constructed language model.

\subsection{Proposed model}

In this study, we consider a language model that significantly simplifies natural language, which we call the context-sensitive random language model~\cite{Sipser_001, De-Giuli_001, De-Giuli_002, Nakaishi_001, Lalegani_001, Stengele_001}.
In the language model, we begin with a randomly chosen symbol, which is a nonterminal symbol.
Then, a symbol in the sentence is first uniformly chosen at random and one of the following three generation rules is applied to the symbol:
\begin{subequations} \label{supp_eq_rule_001_001}
	\begin{align}
		X                   & \to x                   & (\text{probability:} & ~qt), \label{supp_eq_rule_002_001}        \\
		X                   & \to YZ                  & (\text{probability:} & ~q (1 - t)), \label{supp_eq_rule_002_002} \\
		\alpha_- X \alpha_+ & \to \alpha_- Y \alpha_+ & (\text{probability:} & ~(1 - q)). \label{supp_eq_rule_002_003}
	\end{align}
\end{subequations}
Here, uppercase letters $X$, $Y$, and $Z$ represent nonterminal symbols, taking values $A_1, A_2, \dots, A_K$.
Lowercase letters $x$, $y$, and $z$ represent terminal symbols, taking values $a_1, a_2, \dots, a_K$.
First, Eq.~\eqref{supp_eq_rule_002_001} represents rules where nonterminal symbols correspond to their respective terminal symbols.
Second, Eq.~\eqref{supp_eq_rule_002_002} represents rules where a nonterminal symbol is replaced by two nonterminal symbols.
Last, Eq.~\eqref{supp_eq_rule_002_003} introduces context-dependency to nonterminal symbols, rewriting them based on the context $\alpha_-$ and $\alpha_+$, which are strings before and after the symbol of interest, respectively.
Furthermore, $Y$ in Eq.~\eqref{supp_eq_rule_002_003}, called the proposed sample in the Monte Carlo (MC) method, is chosen from a probability distribution that satisfies the detailed balance condition.
In this paper, we run the Metropolis-Hastings algorithm to determine if this update is executed or not.
We describe the details later.

While there may be room for discussion regarding the extent of context-dependency in natural languages, we adopt a CSG with a focus on strong context access in modern neural language models.
It is worth noting that excluding Eq.~\eqref{supp_eq_rule_002_003} leads to a CFG, aligning with models discussed in previous studies~\cite{Sipser_001, De-Giuli_001, De-Giuli_002, Nakaishi_001, Lalegani_001, Stengele_001}.
The probabilities of adopting rules in Eqs.~\eqref{supp_eq_rule_002_001}, \eqref{supp_eq_rule_002_002}, and \eqref{supp_eq_rule_002_003} are denoted as $qt$, $q (1 - t)$, and $1 - q$, respectively.
The parameter $t \to 0$ represents the limit in which terminal symbols do not appear, and this limit enables us to discuss phase transitions easily because the thermodynamic limit $N \to \infty$ can be taken straightforwardly.
Then, in this paper, we focus on this limit for simplicity.

\subsection{One-dimensional long-range Potts model}

In Refs.~\cite{Dyson_001, Frohlich_001, Dyson_002, Dyson_003, Thouless_001, Chang_001, Martinez-Herrera_001, Tomita_001}, phase transitions in the following one-dimensional long-range Potts model~\footnote{When $K = 2$, this model reduces to the Ising model.} are discussed~\cite{Wu_001}:
	\begin{align}
		H & = - \frac{1}{2} \sum_{\substack{i, j = 1, 2, \dots, N: \\ i \ne j}} J_{i, j} \delta_{\sigma_i, \sigma_j}, \label{supp_eq_def_long_range_Potts_model_001_001}
	\end{align}
	where
	\begin{align}
		J_{i, j} & =
		\begin{cases}
			\frac{J}{|i - j|^{1 + s}} & (j - i \in [- N r_-, N r_+]), \\
			0                         & (\text{otherwise}).
		\end{cases}
	\end{align}%
Here $r_+, r_- \in [0, 1]$ are parameters that specify the range of interaction.
Note that Eq.~\eqref{supp_eq_def_long_range_Potts_model_001_001} is referred to as the ferromagnetic model because symbols are more likely to have the same state for $J \gg 0$.
Additionally, when $X_j = A_k, a_k$ for $k = 1, 2, \dots, K$, $\sigma_j$ is defined as
\begin{align}
	\sigma_j & = k. \label{supp_eq_def_sigma_001_001}
\end{align}
For pedagogical purposes, we describe the case of $K = 2$ in Appendix~\ref{supp_sec_model_K=2_001_001}.

\subsection{One-dimensional long-range Ising model} \label{supp_sec_model_K=2_001_001}

In the literature~\cite{Budrikis_001}, each $\sigma_i$ takes $\{1, -1 \}$ rather than $\{0, 1 \}$ for $K = 2$, i.e., the Ising model; then, we follow the notation in this subsection.
Therefore, Eq.~\eqref{supp_eq_def_long_range_Potts_model_001_001} with $K = 2$ reads
	\begin{align}
		H & = - \frac{1}{4} \sum_{\substack{i, j = 1, 2, \dots, N: \\ i \ne j}} J_{i, j} \sigma_i \sigma_j,
	\end{align}
and Eq.~\eqref{supp_eq_def_sigma_001_001} is transformed into
\begin{align}
	\sigma_j & \coloneqq
	\begin{cases}
		1  & (X_j = A_1, a_1), \\
		-1 & (X_j = A_2, a_2).
	\end{cases}
\end{align}
For Monte Carlo updates, the change of the energy of a system by a spin flip is crucially important; Eq.~\eqref{supp_eq_Delta_E_001_001} for $\sigma_0 \to \tilde{\sigma}_0 = - \sigma_0$ becomes
\begin{align}
	\Delta E & \coloneqq - \frac{J}{2} \vast( \sum_{\ell=1}^{\lfloor N r_- \rfloor} \frac{(\tilde{\sigma}_0 - \sigma_0) \sigma_{-\ell}}{\ell^{1 + s}} + \sum_{\ell=1}^{\lfloor N r_+ \rfloor} \frac{(\tilde{\sigma}_0 - \sigma_0) \sigma_{\ell}}{{\ell}^{1 + s}} \vast) \\
	         & = J \vast( \sum_{\ell=1}^{\lfloor N r_- \rfloor} \frac{\sigma_0 \sigma_{-\ell}}{\ell^{1 + s}} + \sum_{\ell=1}^{\lfloor N r_+ \rfloor} \frac{\sigma_0 \sigma_{\ell}}{{\ell}^{1 + s}} \vast).
\end{align}

Next, we review the studies on phase transitions of the one-dimensional long-range Ising model (the one-dimensional long-range Potts model with $K = 2$)~\eqref{supp_eq_def_long_range_Potts_model_001_001}.
For this model, phase transitions occur for $s \in (0, 1)$ while phase transitions do not occur for $s > 1$~\cite{Dyson_001, Dyson_002, Dyson_003, Tomita_001, Martinez-Herrera_001}.
In Ref.~\cite{Frohlich_001}, the existence of a phase transition for $s = 1$ was reported.
Furthermore, when $s = 1$, it is known that the BKT transition occurs~\cite{Kosterlitz_003, Kosterlitz_004}.

There are numerous studies on critical exponents for various $s$, which is called the decay exponent in this paper~\cite{Luijten_001}.
For $s \in (0, 0.5)$, it has been reported that the phase transition follows the mean-field approximation and exhibits a jump in the specific heat~\cite{Krech_001}.
Additionally, the critical exponent is noted to be independent of temperature in this range~\cite{Aizenman_001}.
For $s \in (0.5, 1.0)$, a sharp peak in the specific heat suggests a phase transition distinct from that predicted by the mean-field approximation~\cite{Krech_001}.
In this paper, we consider the decay exponent $s$ within this range for simplicity.
Furthermore, at $s = 1$, there is a jump in magnetization~\cite{Dyson_003, Aizenman_002, Thouless_001}.
It is also worth noting that an effective MC method was proposed in Ref.~\cite{Fukui_001}.

\subsection{Context-dependency}

Let us elaborate on Eq.~\eqref{supp_eq_rule_002_003}.
Let $N$ be the length of the entire string, and define $\alpha_- \coloneqq X_{- \lfloor N r_- \rfloor} X_{- (\lfloor N r_- \rfloor - 1)} \dots X_{-2} X_{-1}$ and $\alpha_+ \coloneqq X_1 X_2 \dots X_{\lfloor N r_+ \rfloor - 1} X_{\lfloor N r_+ \rfloor}$.
Here, $X$ represents some character in the alphabet, and the subscript values denote the relative positions with respect to $A_1, A_2, \dots, A_K$ in Eq.~\eqref{supp_eq_rule_002_003}.
The floor function $\lfloor x \rfloor$ indicates the greatest integer that is equal to or smaller than $x \in \mathbb{R}$, and $r_-$ and $r_+$ are parameters corresponding to the interaction range of the one-dimensional long-range Potts model.

Let $i = 0$ be the location of $X$.
Motivated by the Metropolis-Hastings algorithm, when Eq.~\eqref{supp_eq_rule_002_003} is chosen, symbols are changed with the following probability:
\begin{align}
	p & = \min (1, \mathrm{e}^{- \frac{\Delta E}{k_\mathrm{B} T}}), \label{supp_eq_probability_MC_rule_001_003}
\end{align}
where $\Delta E$ for $\sigma_0 \to \tilde{\sigma}_0$ is defined as
\begin{align}
	\Delta E & \coloneqq J \vast( \sum_{\ell=1}^{\lfloor N r_- \rfloor} \frac{\delta_{\sigma_0, \sigma_{-\ell}} - \delta_{\tilde{\sigma}_0, \sigma_{-\ell}}}{\ell^{1 + s}} + \sum_{\ell=1}^{\lfloor N r_+ \rfloor} \frac{\delta_{\sigma_0, \sigma_{\ell}} - \delta_{\tilde{\sigma}_0, \sigma_{\ell}}}{{\ell}^{1 + s}} \vast). \label{supp_eq_Delta_E_001_001}
\end{align}
Note that Eq.~\eqref{supp_eq_Delta_E_001_001} is computed from Eq.~\eqref{supp_eq_def_long_range_Potts_model_001_001}: $\Delta E = \tilde{H} - H$.

According to Eq.~\eqref{supp_eq_probability_MC_rule_001_003}, symbols are more likely to flip when $\Delta E$ is small.
Furthermore, from Eq.~\eqref{supp_eq_Delta_E_001_001}, $\Delta E$ decreases when the numerators of the first and second terms of the right-hand side of Eq.~\eqref{supp_eq_Delta_E_001_001} take negative values.
Considering Eq.~\eqref{supp_eq_def_sigma_001_001}, this means symbols are flipped when the surrounding symbols are of a different type.
The parameter $k_\mathrm{B} T$ controls the tendency of the same symbols to align and corresponds to the strength of the desire in statistical mechanics to align with the absolute temperature.

Equation~\eqref{supp_eq_Delta_E_001_001} is termed a long-range interaction due to the power-law decay ($1/\ell^{1+s}$) of interactions between symbols.
The decay exponent $s$ determines the strength of the power-law decay, and in this paper, we fix $s=0.9$.
In summary, the rule in Eq.~\eqref{supp_eq_rule_002_003} is statistically equivalent to the Metropolis-Hastings algorithm for flipping a single symbol in the one-dimensional long-range Potts model, as applied in MC simulations.

\section{Methods: physical quantities for discussing phase transitions and finite-size scaling} \label{supp_sec_methods_001_001}

Phase transitions are discerned through the singularity of order parameters, which represent the expectation value of a function of random variables.
In models like the Potts model, magnetization, denoting the ratio of the difference between the numbers of up spins and down spins to the sum of these numbers, is commonly employed as an order parameter.
Subsequently, we examine the second-order moment of the random variable, known as the susceptibility.
At critical points, the susceptibility typically diverges, making it crucial to observe its behavior to confirm phase transitions.
More interestingly, the susceptibility is always divergent in the BKT phase.
Last, the Binder parameter, akin to the normalized version of kurtosis, plays a crucial role in discussions of phase transitions.
According to a generalized version of the \textit{central limit theorem}, the Binder parameter is equal to one below the critical temperature and zero above it.
Furthermore, in the case of second-order phase transitions, the lines depicting the temperature dependence of the Binder parameter for fixed system sizes intersect at the critical temperature.

To explore the scaling law, which is known to apply at critical points, we also compute the correlation functions and mutual information.
It is important to note that due to the inherent growth property of language models, there is some arbitrariness in defining the correlation functions and mutual information.

At the end of this section, we review the finite-size scaling method, which is a powerful method based on the scaling assumption to discuss and quantify phase transitions.
In particular, critical exponents are used to classify phase transitions into universality classes.

\subsection{Magnetization} \label{supp_sec_magnetization_001_001}

Although this study does not deal with magnetic materials, an order parameter similar to that in the Potts model can be defined when considering two symbols.
We term this order parameter ``magnetization," and define it as follows:
\begin{align}
	M & \coloneqq \| \bm{M} \|, \label{supp_eq_def_magnetization_001_001}
\end{align}
where $\bm{M} \in \mathbb{R}^{K-1}$ is given by
\begin{align}
	\bm{M} & \coloneqq \frac{1}{N} \sum_{i = 1}^N \bm{e}_{\sigma_i}. \label{supp_eq_order_parameter_Potts_001_001}
\end{align}
Note that $\| \bm{M} \| \coloneqq \sqrt{\bm{M} \cdot \bm{M}}$.
The construction of $\{ \bm{e}_k \}_{k=1}^K$ in Eq.~\eqref{supp_eq_def_magnetization_001_001} is shown in Algorithm~\ref{supp_algo_basis_Binder_parameter_Potts_001_001}.
\begin{algorithm}[t]
	\DontPrintSemicolon
	\caption{Construction of $\{ \bm{e}_k \}_{k=1}^K$ in Eq.~\eqref{supp_eq_def_magnetization_001_001}.}
	\label{supp_algo_basis_Binder_parameter_Potts_001_001}
	set $\bm{e}_1 \coloneqq [1, 0, \dots, 0]^\intercal$ \;
	\For{$k = 2, 3, 4, \dots, K$}{
		initialize $\bm{e}_k \coloneqq [0, 0, \dots, 0]^\intercal$ \;
		\For{$l = 1, 2, 3, \dots, k - 1$}{
			$[\bm{e}_k]_l = - \frac{1}{K - l} [\bm{e}_l]_l$ \;
		}
		$[\bm{e}_k]_k = \sqrt{1 - \sum_{l = 1}^k ([\bm{e}_k]_l)^2}$ \;
	}
\end{algorithm}
Due to its construction in Algorithm~\ref{supp_algo_basis_Binder_parameter_Potts_001_001}, $\{ \bm{e}_k \}_{k=1}^K$ satisfy $\| \bm{e}_k \| = 1$ for $k = 1, 2, \dots, K$, $\bm{e}_k \cdot \bm{e}_l = \mathrm{const}.$ for $k \ne l$, and $\sum_{k=1}^K \bm{e}_k = \bm{0}$.
Then the $l$-th element of $\{ \bm{e}_k \}_{k=1}^K$ is computed as
\begin{align}
	[\bm{e}_k]_l & \coloneqq
	\begin{cases}
		\sqrt{\frac{K (K - k)}{(K - 1) (K - k + 1)}}   & (k = l), \\
		- \sqrt{\frac{K}{(K - 1) (K - l) (K - l + 1)}} & (k > l), \\
		0                                              & (k < l).
	\end{cases} \label{supp_eq_def_basis_Binder_parameter_Potts_001_001}
\end{align}
The detailed derivation of Eq.~\eqref{supp_eq_def_basis_Binder_parameter_Potts_001_001} is shown in Appendix~\ref{supp_sec_derivation_Binder_parameter_Potts_model_001_001}.
Note that $\langle \bm{M} \rangle = \bm{0}$ because of the construction of $\{ \bm{e}_k \}_{k=1}^K$.

Finally, we consider the meaning of $M$, as defined in Eq.~\eqref{supp_eq_def_magnetization_001_001}.
It signifies the degree of frequency bias among nonterminal symbols $A_1, A_2, \dots, A_K$.
In the subsequent analysis, by observing the variations of $M$, Eq.~\eqref{supp_eq_def_magnetization_001_001}, with respect to $k_\mathrm{B} T$, we aim to ascertain whether a critical point exists where the properties of the language, constructed based on the one-dimensional Potts model, undergo a qualitative and abrupt change.
At the critical point of the phase transition, the magnetization, as defined in Eq.~\eqref{supp_eq_def_magnetization_001_001}, becomes non-differentiable with respect to a control parameter, which is temperature in this paper, in the thermodynamic limit $N \to \infty$.

\subsection{Derivation of Eq.~\eqref{supp_eq_def_basis_Binder_parameter_Potts_001_001}} \label{supp_sec_derivation_Binder_parameter_Potts_model_001_001}

The derivation of Eq.~\eqref{supp_eq_def_basis_Binder_parameter_Potts_001_001} can be shown via mathematical induction.
That is, $[\bm{e}_{k + 1}]_{k + 1}$ is computed as follows:
\begin{align}
	 & [\bm{e}_{k + 1}]_{k + 1} \nonumber                                                                \\
	 & = \sqrt{1 - \sum_{l = 1}^k [\bm{e}_{k + 1}]_{l}^2}                                                \\
	 & = \sqrt{1 - \sum_{l = 1}^k \bigg(- \sqrt{\frac{K}{(K - (l - 1)) (K - 1) (K - l)}} \bigg)^2}       \\
	 & = \sqrt{1 - \sum_{l = 1}^k \frac{K}{(K - (l - 1)) (K - 1) (K - l)}}                               \\
	 & = \sqrt{1 - \frac{K}{K - 1} \sum_{l = 1}^k \frac{1}{(K - (l - 1)) (K - l)}}                       \\
	 & = \sqrt{1 - \frac{K}{K - 1} \sum_{l = 1}^k \bigg( \frac{1}{K - l} - \frac{1}{K - (l - 1)} \bigg)} \\
	 & = \sqrt{1 - \frac{K}{K - 1} \bigg( \frac{1}{K - k} - \frac{1}{K} \bigg)}                          \\
	 & = \sqrt{1 - \frac{k}{(K - 1) (K - k)}}                                                            \\
	 & = \sqrt{\frac{K (K - k - 1)}{(K - 1) (K - k)}}. \label{supp_eq_e_k+1_k+1_001_001}
\end{align}
Using the value of $[\bm{e}_{k + 1}]_{k + 1}$, Eq.~\eqref{supp_eq_e_k+1_k+1_001_001}, and the two conditions of $\{ \bm{e}_k \}_k$, we can compute $[\bm{e}_{k + 1}]_{l}$ for $k + 1 > l$, and the other elements are zero by definition.
Thus, Eq.~\eqref{supp_eq_def_basis_Binder_parameter_Potts_001_001} holds for $k + 1$.

\subsection{Susceptibility}

Next, we define the susceptibility as follows:
\begin{align}
	\chi & \coloneqq N (\langle M^2 \rangle - \langle M \rangle^2). \label{supp_eq_def_specific_heat_001_001}
\end{align}
The susceptibility, Eq.~\eqref{supp_eq_def_specific_heat_001_001}, is known to diverge just above the critical point and is well-studied due to its ease of experimental measurement, e.g., in the $q$-state clock model~\cite{Kumano_001, Miyajima_001}.
We also note that the following form is another candidate for defining the susceptibility:
\begin{align}
	\tilde{\chi} & \coloneqq N (\langle \bm{M} \cdot \bm{M} \rangle - \langle \bm{M} \rangle \cdot \langle \bm{M} \rangle) \\
	             & = N \langle \bm{M} \cdot \bm{M} \rangle                                                                 \\
	             & = N \langle M^2 \rangle. \label{supp_eq_def_specific_heat_001_002}
\end{align}
However, Eq.~\eqref{supp_eq_def_specific_heat_001_002} is known to be valid only for the disordered phase.

\subsection{Binder parameter}

The non-differentiability of the magnetization, as expressed in Eq.~\eqref{supp_eq_def_magnetization_001_001}, and the divergence of susceptibility, as defined in Eq.~\eqref{supp_eq_def_specific_heat_001_001}, are fundamental characteristics of phase transitions.
However, in numerical experiments, encountering challenges in taking the thermodynamic limit $N \to \infty$ due to finite-system sizes is common.
Consequently, analysis using the Binder parameter is often preferred~\cite{Sandvik_001, Landau_001}.
The Binder parameter is defined as follows:
\begin{align}
	U & \coloneqq - \frac{K - 1}{2} \bigg( \frac{\langle (\bm{M} \cdot \bm{M})^2 \rangle}{\langle \bm{M} \cdot \bm{M} \rangle^2} - \frac{K^2-1}{(K-1)^2} \bigg) \\
	  & = - \frac{K - 1}{2} \bigg( \frac{\langle M^4 \rangle}{\langle M^2 \rangle^2} - \frac{K+1}{K-1} \bigg). \label{supp_eq_def_Binder_parameter_001_001}
\end{align}
It is noteworthy that when $\bm{M}$ is sampled from a Gaussian distribution with $K$ degrees of freedom, $\langle M^4 \rangle / \langle M^2 \rangle^2$ evaluates to $K^2-1 / (K-1)^2$.
Consequently, Eq.~\eqref{supp_eq_def_Binder_parameter_001_001} equals zero for the disordered state and one when all the spin variables assume the same value.
It is known that at the critical point, the temperature dependence of the Binder parameter, as defined in Eq.~\eqref{supp_eq_def_Binder_parameter_001_001}, for different system sizes intersects.
Moreover, it is also recognized that it diverges negatively at non-critical phase transition points, that is, the first-order phase transition.
Hence, the method of investigating the system size dependence of the Binder parameter, as defined in Eq.~\eqref{supp_eq_def_Binder_parameter_001_001}, has become standard for identifying phase transition points in numerical computations.
From a statistical perspective, the Binder parameter is analogous to kurtosis, as it quantifies the tailedness of probability distributions.

\subsection{Correlation function} \label{supp_sec_correlation_function_001_001}

By using $\{ \bm{e}_k \}_{k=1}^K$ in Eq.~\eqref{supp_eq_def_basis_Binder_parameter_Potts_001_001}, we define the following correlation function:
\begin{align}
	G (i, j) & \coloneqq \langle (\bm{e}_{\sigma_i} - \langle \bm{e}_{\sigma_i} \rangle) \cdot (\bm{e}_{\sigma_j} - \langle \bm{e}_{\sigma_j} \rangle) \rangle                                                 \\
	         & = \langle \bm{e}_{\sigma_i} \cdot \bm{e}_{\sigma_j} \rangle - \langle \bm{e}_{\sigma_i} \rangle \cdot \langle \bm{e}_{\sigma_j} \rangle. \label{supp_eq_def_correlation_function_Potts_001_001}
\end{align}
In Eq.~\eqref{supp_eq_def_correlation_function_Potts_001_001}, the following quantity is an essential part:
\begin{align}
	\tilde{G} (i, j) & \coloneqq \langle \bm{e}_{\sigma_i} \cdot \bm{e}_{\sigma_j} \rangle. \label{supp_eq_correlation_function_with_disconnected_diagram_Potts_001_001}
\end{align}
Similar to Eq.~\eqref{supp_eq_def_specific_heat_001_002}, we compute Eq.~\eqref{supp_eq_correlation_function_with_disconnected_diagram_Potts_001_001} instead of Eq.~\eqref{supp_eq_def_correlation_function_Potts_001_001}.
After some algebra, we obtain the following correlation function from Eq.~\eqref{supp_eq_correlation_function_with_disconnected_diagram_Potts_001_001}~\cite{Fujimoto_001}:
\begin{align}
	\tilde{G} (i, j) & = \frac{K \langle \delta_{\sigma_i, \sigma_j} \rangle - 1}{K - 1}.
\end{align}
The correlation function is considered to be important for the research on phase transitions since it decays polynomially at critical points, at which second-order phase transitions take place, and decays exponentially otherwise.

\subsection{Mutual information}

Mutual information is a measure that quantifies the correlation between two random variables~\cite{Lin_001, Nakaishi_002}:
\begin{align}
	I (i, j) & = \sum_{\sigma_i, \sigma_j = 1}^{K} p (\sigma_i, \sigma_j) \ln \frac{p (\sigma_i, \sigma_j)}{p (\sigma_i) p (\sigma_j)}. \label{supp_eq_mutual_information_Potts_001_001}
\end{align}
In Ref.~\cite{Lin_001}, the scaling law of mutual information for CFGs was reported.

\subsection{Histogram of the magnetization}

For $\Delta M$ denoting the discretization interval of the magnetization and $n$ an integer, the histogram of the magnetization, Eq.~\eqref{supp_eq_def_magnetization_001_001}, is defined as
\begin{align}
	\mathrm{hist} (M) & \coloneqq \# \{ M_i | M_i \in [n \Delta M, (n + 1) \Delta M) \}, \label{supp_eq_histogram_magnetization_001_001}
\end{align}
where $M_i$ is the magnetization, Eq.~\eqref{supp_eq_def_magnetization_001_001}, of the $i$-th MC sample for $i = 1, 2, \dots, N_\mathrm{MC}$ and $N_\mathrm{MC}$ is the number of MC samples.

In second-order phase transitions, the histogram of the magnetization, as defined in Eq.~\eqref{supp_eq_histogram_magnetization_001_001}, transitions from a distribution with a peak at $M = 0$ for high $k_\mathrm{B} T$ above the critical temperature to a uniform distribution at the critical point.
Subsequently, it transitions to a distribution with a peak at $M = 1$ for low $k_\mathrm{B} T$ below the critical temperature.
However, in first-order phase transitions, the histogram of the magnetization, as defined in Eq.~\eqref{supp_eq_histogram_magnetization_001_001}, exhibits two peaks at the transition temperature.
This phenomenon is attributed to hysteresis.

\subsection{Finite-size scaling} \label{supp_sec_methods_002_011}

In thermodynamics, we are interested in physical quantities in the thermodynamic limit $N \to \infty$; however, it is obviously impossible to take this limit in numerical simulations.
Finite-size scaling is one of the numerical methods based on the scaling assumption to infer the behavior of physical quantities in the thermodynamic limit from numerical results of finite-size systems by extrapolating physical quantities of several system sizes.

Let us consider the system size and temperature dependence of a certain physical quantity $A$: $A (N, T)$.
In general, we can consider any parameter instead of temperature $T$ or even multiple parameters at the same time.
At $T \sim T_\mathrm{c}$, $A (N, T)$ is empirically known to behave in the thermodynamic limit as follows:
\begin{align}
	A (N, T) & = t^{- x_A}, \label{supp_eq_functional_form_A_L_T_001_001}
\end{align}
where $x_A$ is the critical exponent of $A$ and
\begin{align}
	t & \coloneqq \bigg| \frac{T_\mathrm{c} - T}{T_\mathrm{c}} \bigg|.
\end{align}
Similarly, the correlation length $\xi_\infty$ diverges as
\begin{align}
	\xi_\infty & \propto t^{- \nu}.
\end{align}
where $\nu$ is the critical exponent of the correlation length.
The scaling assumption states that there exists a function $f_A (\cdot)$ for $A (N, T)$ such that
\begin{align}
	A (N, T) & = N^\frac{x_A}{\nu} f_A \bigg( \frac{N}{\xi_\infty} \bigg). \label{supp_eq_scaling_assumption_A_L_T_001_001}
\end{align}

Let us take $\tilde{\chi}$, Eq.~\eqref{supp_eq_def_specific_heat_001_002}, as an example.
Equation~\eqref{supp_eq_functional_form_A_L_T_001_001} for it reads
\begin{align}
	\tilde{\chi} (T) & \propto t^{- \gamma},
\end{align}
where $\gamma$ is the critical exponent for $\tilde{\chi}$, Eq.~\eqref{supp_eq_def_specific_heat_001_002}.
The scaling assumption, Eq.~\eqref{supp_eq_scaling_assumption_A_L_T_001_001}, for $\tilde{\chi}$, Eq.~\eqref{supp_eq_def_specific_heat_001_002}, becomes
\begin{align}
	\frac{\tilde{\chi} (T, N)}{N^\frac{\gamma}{\nu}} & = f_{\tilde{\chi}} (N t^\nu)                                                                            \\
	                                                                                     & = \tilde{f}_{\tilde{\chi}} (N^\frac{1}{\nu} t). \label{supp_eq_scaling_assumption_tile_chi_L_T_001_001}
\end{align}
Note that $\tilde{f}_{\tilde{\chi}} (\cdot)$ is also a universal function for $\tilde{\chi}$, Eq.~\eqref{supp_eq_def_specific_heat_001_002}.
We verify Eq.~\eqref{supp_eq_scaling_assumption_tile_chi_L_T_001_001} in the main text.

\section{Numerical simulations} \label{supp_sec_numerical_simulations_001_001}

We conduct Monte Carlo simulations $80 \times 80$ times and present the numerical results of the physical quantities defined in Appendix~\ref{supp_sec_methods_001_001} for the proposed model, as described in Eq.~\eqref{supp_eq_rule_001_001}.
Additionally, considering the rule $X \to YZ$, in which symbols are added with equal probability, from Eq.~\eqref{supp_eq_rule_002_002}, we also examine the case $X \to XX$, in which symbols are duplicated, because we aim to investigate how the rule, in which a sentence grows, unique to language models affects physical quantities in the thermodynamic limit.
Here, we investigate the cases of $K = 2$ and $K = 10$, setting $J = 1.0$.
This is because the Potts model is known to exhibit a second-order phase transition for $K \le 4$ and a first-order phase transition for $K \ge 5$, suggesting that the proposed model is also expected to show qualitatively different behavior for $K = 2$ and $K = 10$.

First, we show the magnetization, the susceptibility, and the Binder parameter.
Second, we plot the correlation functions and mutual information.
Third, we depict the histogram of the magnetization.

Next, we show the system-size dependence of the magnetization, the two susceptibilities, and the Binder parameter.
We then draw the phase diagrams for several numerical conditions.
We see that the phase diagram drastically changes depending on $X \to YZ$ and $X \to XX$.
We then plot the results of the finite-size scaling method.
We see that the critical exponents are independent of the choice of $X \to YZ$ or $X \to XX$ but vary depending on $q$.
Finally, we plot the $q$-dependence of the critical exponents $\nu$ and $\gamma$.

For $K = 10$, we show the relative frequencies of letters in the alphabet.
In the literature on natural language processing, the relative frequencies of letters in the alphabet are known to often satisfy Zipf's law.

\subsection{Case of $K=2$ and $X \to YZ$}

Let us consider the case of $K = 2$ and $X \to YZ$, in which symbols are added with equal probability.
Since there are two nonterminal symbols and they increase randomly, there are eight processes to consider:
\begin{align}
	A \to AA, \ A \to AB, \ A \to BA, \ A \to BB, \nonumber \\
	B \to AA, \ B \to AB, \ B \to BA, \ B \to BB.
\end{align}
That is, it is assumed that these processes occur with equal probability for $A$ and $B$.

\subsubsection{Magnetization, susceptibilities, and Binder parameter}

\begin{figure}[t]
	\centering
	\includegraphics[scale=0.60]{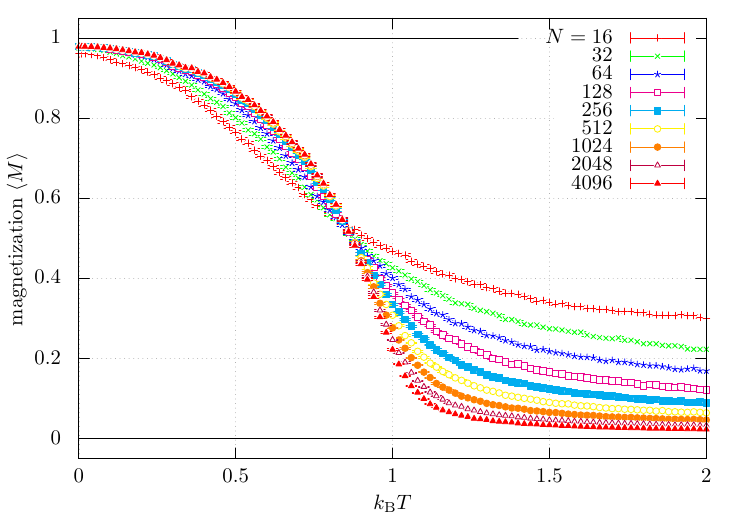}
	\includegraphics[scale=0.60]{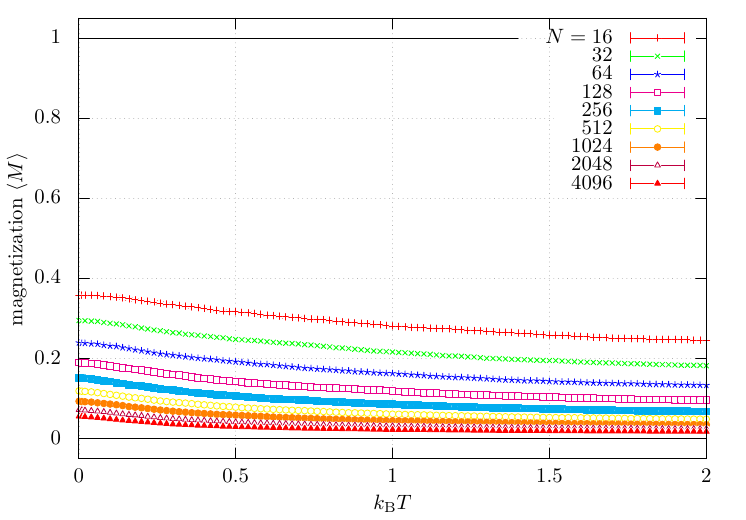}
	\caption{Temperature dependence of the magnetization, Eq.~\eqref{supp_eq_def_magnetization_001_001}, for $X \to YZ$. We set (upper) $q = 10^{-2.0}$ and (lower) $q = 10^{-0.5}$. We also set $K = 2$, $J = 1.0$, $t = 0$, $s = 0.9$, and $r_- = r_+ = 0.25$. The length of the generated sentence by the language model, $N$, was varied from $16$ to $4096$. The upper panel is identical to Fig.~\ref{main_fig_magnetization_K=2_q_10_-2.0_X_YZ_001_001}(upper).}
	\label{supp_fig_magnetization_K=2_X_YZ_001_001}
\end{figure}

In Fig.~\ref{supp_fig_magnetization_K=2_X_YZ_001_001}, the temperature dependence of the magnetization, Eq.~\eqref{supp_eq_def_magnetization_001_001}, for $q = 10^{-2.0}, 10^{-0.5}$ is shown.
In the case of $q = 10^{-2.0}$, as the system size $N$ increases below the transition point $T \lesssim 1$, the magnetization converges to a nonzero value, while above the transition point, it converges to zero.
This behavior is consistent with the critical $\mathbb{Z}_2$ symmetry breaking observed alongside the intersection of the Binder parameters.
However, in the case of $q = 10^{-0.5}$, $\langle M \rangle$ increases toward $k_\mathrm{B} T = 0$, but it seems to hover around zero for any temperature.
Therefore, Fig.~\ref{supp_fig_magnetization_K=2_X_YZ_001_001} implies that the critical behaviors of the context-sensitive random language model, as described in Eq.~\eqref{supp_eq_rule_001_001}, strongly depend on the value of $q$.

\begin{figure}[t]
	\centering
	\includegraphics[scale=0.60]{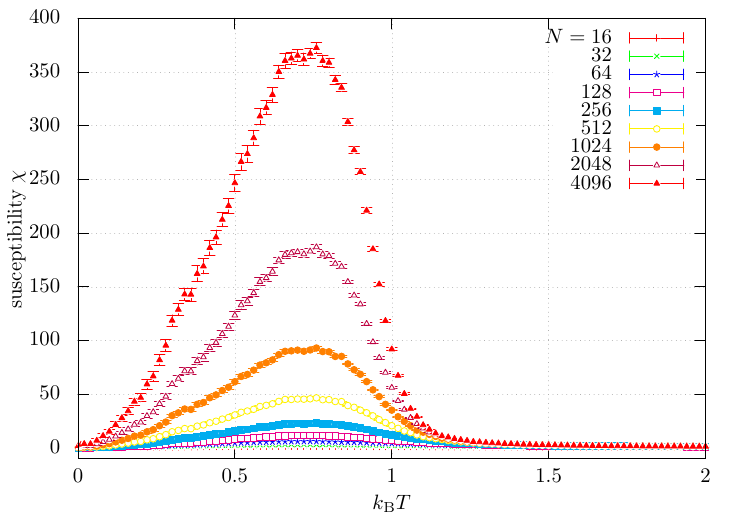}
	\includegraphics[scale=0.60]{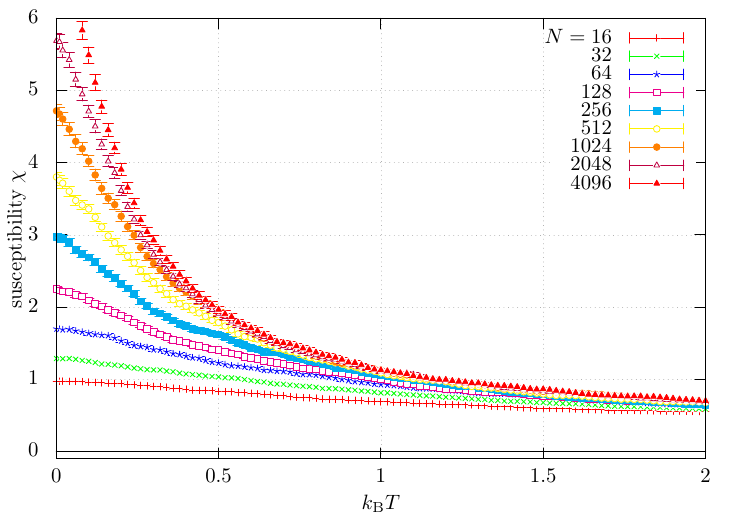}
	\caption{Temperature dependence of the susceptibility, Eq.~\eqref{supp_eq_def_specific_heat_001_001}, for $X \to YZ$. We set (upper) $q = 10^{-2.0}$ and (lower) $q = 10^{-0.5}$. We also set $K = 2$, $J = 1.0$, $t = 0$, $s = 0.9$, and $r_- = r_+ = 0.25$. The length of the generated sentence by the language model, $N$, was varied from $16$ to $4096$. The upper panel is identical to Fig.~\ref{main_fig_magnetization_K=2_q_10_-2.0_X_YZ_001_001}(middle).}
	\label{supp_fig_susceptibility_K=2_X_YZ_001_001}
\end{figure}

Next, we show the susceptibility, Eq.~\eqref{supp_eq_def_specific_heat_001_001}, in Fig.~\ref{supp_fig_susceptibility_K=2_X_YZ_001_001}.
Figure~\ref{supp_fig_susceptibility_K=2_X_YZ_001_001} seems to be consistent with Fig.~\ref{supp_fig_magnetization_K=2_X_YZ_001_001} since it illustrates a diverging behavior of the susceptibility, typical for a second-order phase transition, at a specific temperature in the case of $q = 10^{-2.0}$.
However, it does not show a cusp at its peak, which is a typical signal of the second-order phase transition.
However, it monotonically increases toward zero temperature without exhibiting any peaks for $q = 10^{-0.5}$.
This result also implies that there is no phase transition for $q = 10^{-0.5}$.

\begin{figure}[t]
	\centering
	\includegraphics[scale=0.60]{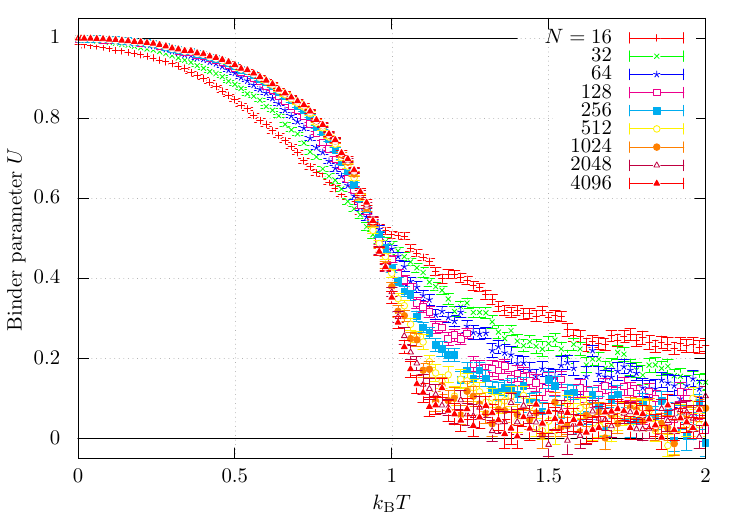}
	\includegraphics[scale=0.60]{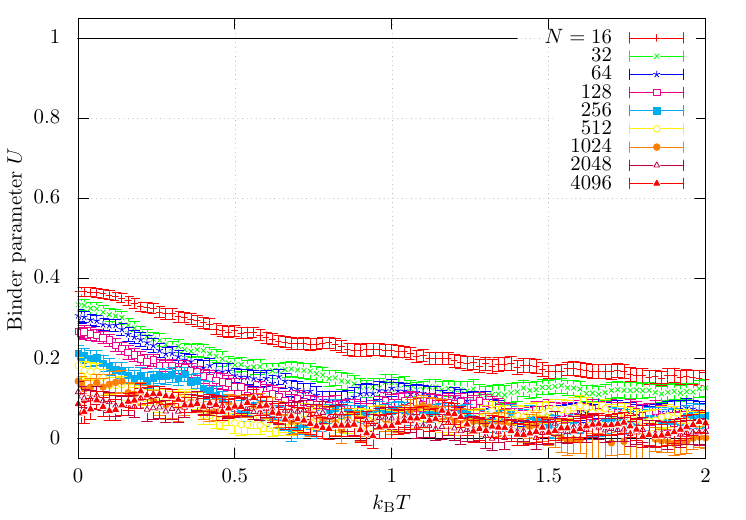}
	\caption{Temperature dependence of the Binder parameter, Eq.~\eqref{supp_eq_def_Binder_parameter_001_001}, for $X \to YZ$. We set (upper) $q = 10^{-2.0}$ and (lower) $q = 10^{-0.5}$. We also set $K = 2$, $J = 1.0$, $t = 0$, $s = 0.9$, and $r_- = r_+ = 0.25$. The length of the generated sentence by the language model, $N$, was varied from $16$ to $4096$. The upper panel is identical to Fig.~\ref{main_fig_magnetization_K=2_q_10_-2.0_X_YZ_001_001}(lower).}
	\label{supp_fig_Binder_parameter_K=2_X_YZ_001_001}
\end{figure}

In Fig.~\ref{supp_fig_Binder_parameter_K=2_X_YZ_001_001}, the Binder parameter, Eq.~\eqref{supp_eq_def_Binder_parameter_001_001}, is shown.
In Fig.~\ref{supp_fig_Binder_parameter_K=2_X_YZ_001_001}(upper), a clear intersection of the Binder parameter is visible, indicating the presence of a critical point around $k_\mathrm{B} T \sim 1.0$, while there is no crossing of the Binder parameter in Fig.~\ref{supp_fig_Binder_parameter_K=2_X_YZ_001_001}(lower).

\subsubsection{Correlation functions, mutual information, and histogram of magnetization for $q = 10^{-2.0}$}

\begin{figure}[t]
	\centering
	\includegraphics[scale=0.60]{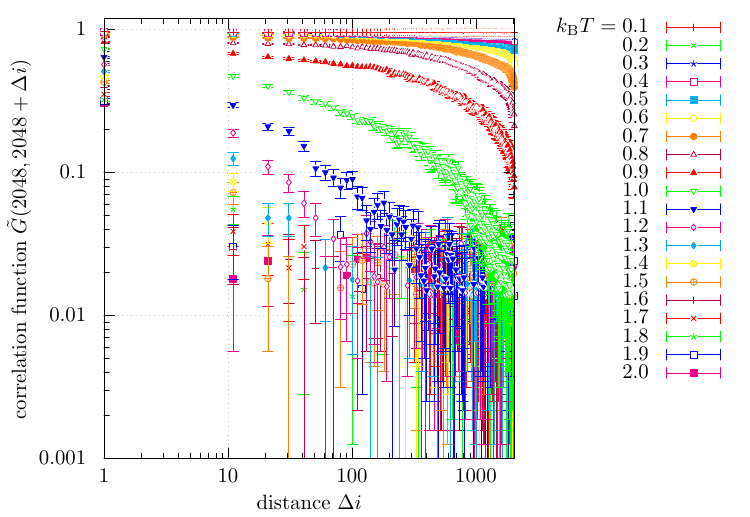}
 	\includegraphics[scale=0.60]{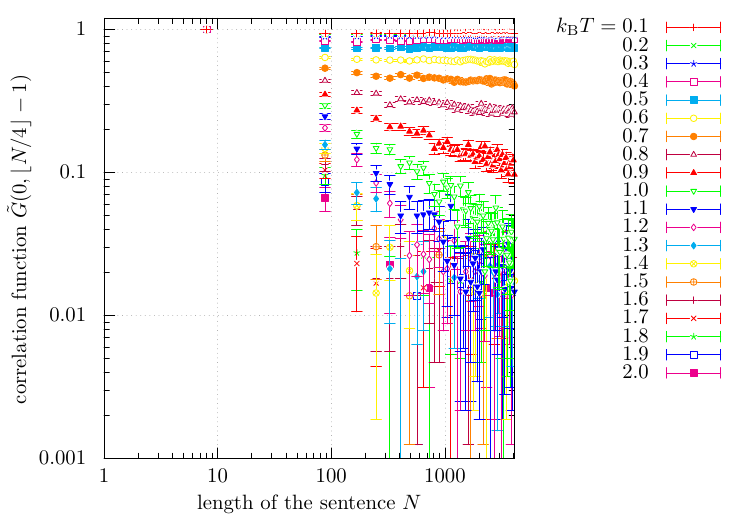}
	\caption{Correlation functions, Eq.~\eqref{supp_eq_correlation_function_with_disconnected_diagram_Potts_001_001}, (upper) with $i=2048$ and $j=2048 + \Delta i$ and (lower) with $i = 0$ and $j = \lfloor N / 4 \rfloor - 1$ for $X \to YZ$. We set $K = 2$, $J = 1.0$, $q = 10^{-2.0}$, $t = 0$, $s = 0.9$, and $r_- = r_+ = 0.25$. The temperature $k_\mathrm{B} T$ was varied from $0.1$ to $2.0$. This figure is identical to Fig.~\ref{main_fig_correlation_function_K=02_q=10^(-2.0)_X_YZ_001_001}.}
	\label{supp_fig_correlation_function_K=02_q=10^(-2.0)_X_YZ_001_001}
\end{figure}

In Fig.~\ref{supp_fig_correlation_function_K=02_q=10^(-2.0)_X_YZ_001_001}, we display two types of correlation functions.
The first one, defined by Eq.~\eqref{supp_eq_correlation_function_with_disconnected_diagram_Potts_001_001} with $i=2048$ and $j=2048 + \Delta i$, quantifies the correlation between symbols $i$ and $j$ within a given temporal sentence.
It is equivalent to the conventional correlation function studied in research on spin systems.
The second type, defined by Eq.~\eqref{supp_eq_correlation_function_with_disconnected_diagram_Potts_001_001} with $i = 0$ and $j = \lfloor N / 4 \rfloor - 1$, measures how symbols correlate with each other throughout the generation process.
This type of correlation function is especially pertinent to models, in which the length of a generated sentence $N$ grows, and cannot be defined for conventional spin models.
Figure~\ref{supp_fig_correlation_function_K=02_q=10^(-2.0)_X_YZ_001_001}(upper) illustrates that for high $k_\mathrm{B} T$, the correlation function decays rapidly with $\Delta i$, while for small $k_\mathrm{B} T$, there is still correlation even for large $\Delta i$.
More interestingly, the correlation function appears to satisfy the scaling behavior around $k_\mathrm{B} T \sim 1.1$.
In Fig.~\ref{supp_fig_correlation_function_K=02_q=10^(-2.0)_X_YZ_001_001}(lower), the scaling law seems to hold for $T \in [0.7, 1.1]$.

\begin{figure}[t]
	\centering
	\includegraphics[scale=0.60]{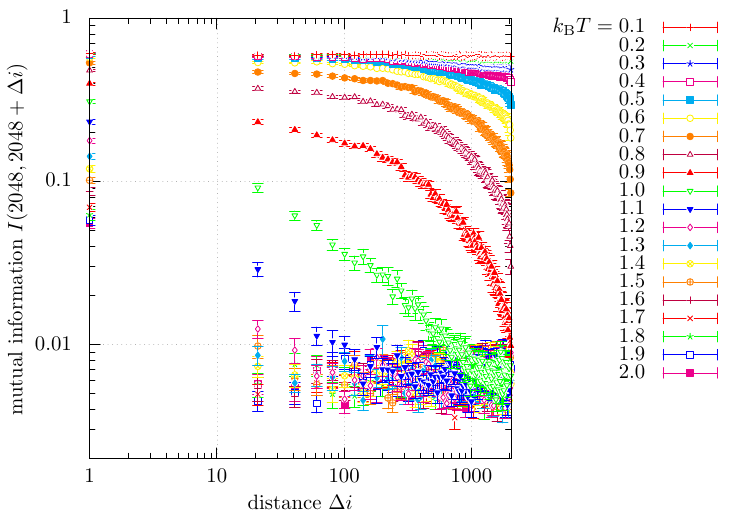}
	\caption{Mutual information, Eq.~\eqref{supp_eq_mutual_information_Potts_001_001}, with $i=2048$ and $j=2048 + \Delta i$ for $X \to YZ$. We set $K = 2$, $J = 1.0$, $q = 10^{-2.0}$, $t = 0$, $s = 0.9$, and $r_- = r_+ = 0.25$. The temperature $k_\mathrm{B} T$ was varied from $0.1$ to $2.0$.}
	\label{supp_fig_mutual_information_K=02_q=10^(-2.0)_X_YZ_001_001}
\end{figure}

In Fig.~\ref{supp_fig_mutual_information_K=02_q=10^(-2.0)_X_YZ_001_001}, we plot mutual information, Eq.~\eqref{supp_eq_mutual_information_Potts_001_001}.
Similar to Fig.~\ref{supp_fig_correlation_function_K=02_q=10^(-2.0)_X_YZ_001_001}(upper), the scaling law seems to hold for $k_\mathrm{B} T \sim 1.0$.

\begin{figure}[t]
	\centering
	\includegraphics[scale=0.60]{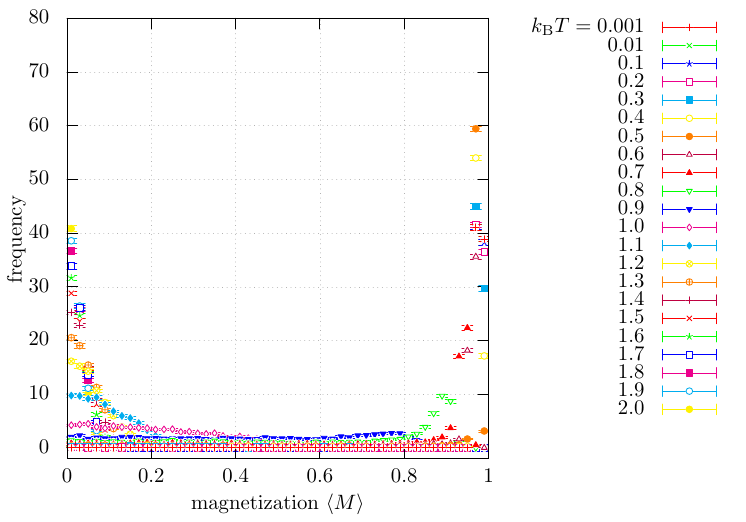}
	\caption{Histogram of the magnetization, Eq.~\eqref{supp_eq_def_magnetization_001_001}, for $X \to YZ$. We set $K = 2$, $J = 1.0$, $q = 10^{-2.0}$, $t = 0$, $s = 0.9$, and $r_- = r_+ = 0.25$. The temperature $k_\mathrm{B} T$ was varied from $0.1$ to $2.0$.}
	\label{supp_fig_histogram_mag_K=02_q=10^(-2.0)_X_YZ_001_001}
\end{figure}

Finally, we show the histogram of the magnetization, Eq.~\eqref{supp_eq_histogram_magnetization_001_001}, in Fig.~\ref{supp_fig_histogram_mag_K=02_q=10^(-2.0)_X_YZ_001_001}.
The histogram of the magnetization, Eq.~\eqref{supp_eq_histogram_magnetization_001_001}, in Fig.~\ref{supp_fig_histogram_mag_K=02_q=10^(-2.0)_X_YZ_001_001} is almost uniform at $k_\mathrm{B} T \sim 0.9$, and this behavior is typical for a second-order phase transition.
For smaller $k_\mathrm{B} T \sim 0.7$, it has a sharp peak around 1, and for larger $k_\mathrm{B} T \sim 1.4$, it has a sharp peak around zero.
Up to now, we can numerically conclude that this phase transition is not a first-order phase transition.
As discussed later, we will conclude that this is a BKT transition, which is similar to a second-order phase transition.

\subsubsection{Correlation functions, mutual information, and histogram of magnetization for $q = 10^{-0.5}$}

\begin{figure}[t]
	\centering
	\includegraphics[scale=0.60]{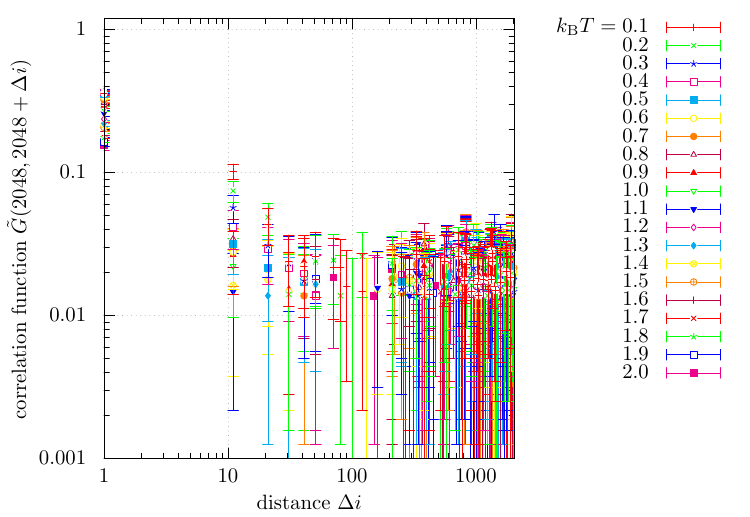}
	\includegraphics[scale=0.60]{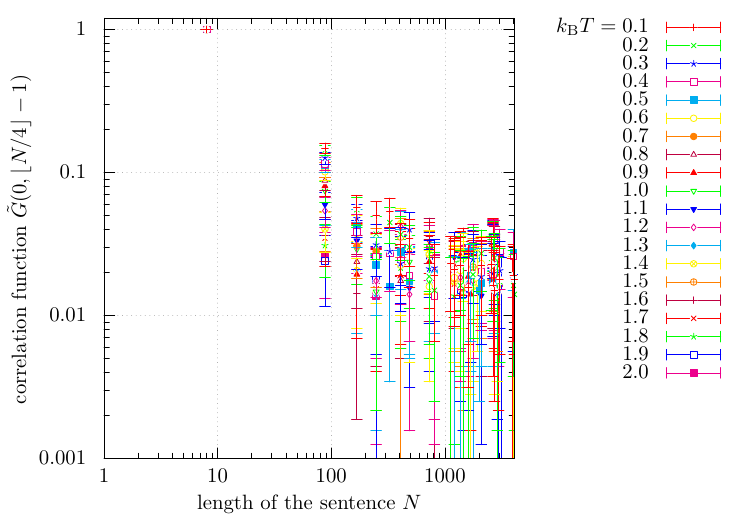}
	\caption{Correlation functions, Eq.~\eqref{supp_eq_correlation_function_with_disconnected_diagram_Potts_001_001}, (upper) with $i=2048$ and $j=2048 + \Delta i$ and (lower) with $i = 0$ and $j = \lfloor N / 4 \rfloor - 1$ for $X \to YZ$. We set $K = 2$, $J = 1.0$, $q = 10^{-0.5}$, $t = 0$, $s = 0.9$, and $r_- = r_+ = 0.25$. The temperature $k_\mathrm{B} T$ was varied from $0.1$ to $2.0$.}
	\label{supp_fig_correlation_function_K=02_q=10^(-0.5)_X_YZ_001_001}
\end{figure}

\begin{figure}[t]
	\centering
	\includegraphics[scale=0.60]{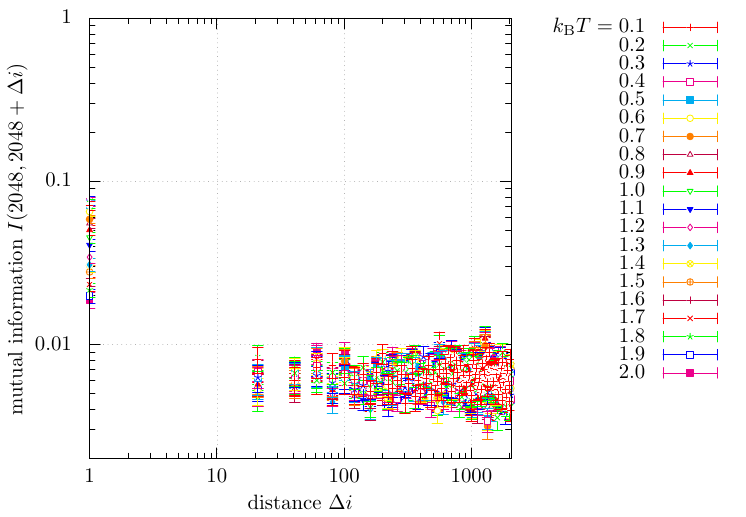}
	\caption{Mutual information, Eq.~\eqref{supp_eq_mutual_information_Potts_001_001}, with $i=2048$ and $j=2048 + \Delta i$ for $X \to YZ$. We set $K = 2$, $J = 1.0$, $q = 10^{-0.5}$, $t = 0$, $s = 0.9$, and $r_- = r_+ = 0.25$. The temperature $k_\mathrm{B} T$ was varied from $0.1$ to $2.0$.}
	\label{supp_fig_mutual_information_K=02_q=10^(-0.5)_X_YZ_001_001}
\end{figure}

\begin{figure}[t]
	\centering
	\includegraphics[scale=0.60]{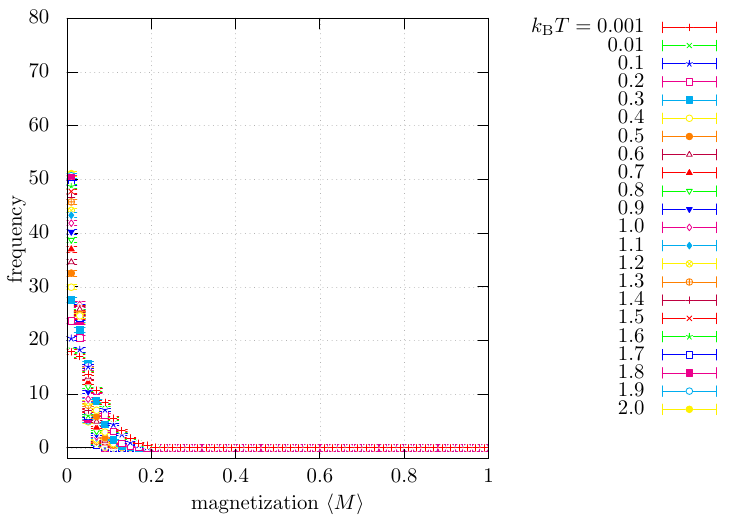}
	\caption{Histogram of the magnetization, Eq.~\eqref{supp_eq_def_magnetization_001_001}, for $X \to YZ$. We set $K = 2$, $J = 1.0$, $q = 10^{-0.5}$, $t = 0$, $s = 0.9$, and $r_- = r_+ = 0.25$. The temperature $k_\mathrm{B} T$ was varied from $0.1$ to $2.0$.}
	\label{supp_fig_histogram_mag_K=02_q=10^(-0.5)_X_YZ_001_001}
\end{figure}

In Figs.~\ref{supp_fig_correlation_function_K=02_q=10^(-0.5)_X_YZ_001_001}, \ref{supp_fig_mutual_information_K=02_q=10^(-0.5)_X_YZ_001_001}, and \ref{supp_fig_histogram_mag_K=02_q=10^(-0.5)_X_YZ_001_001}, we present the correlation functions, Eq.~\eqref{supp_eq_correlation_function_with_disconnected_diagram_Potts_001_001}, mutual information, Eq.~\eqref{supp_eq_mutual_information_Potts_001_001}, and the magnetization, Eq.~\eqref{supp_eq_histogram_magnetization_001_001}, respectively.
These figures are all consistent with the estimate that the system is in the trivial ``disordered'' state.

\subsubsection{System-size dependence of the magnetization, the susceptibilities, and the Binder parameter}

\begin{figure}[t]
	\centering
	\includegraphics[scale=0.60]{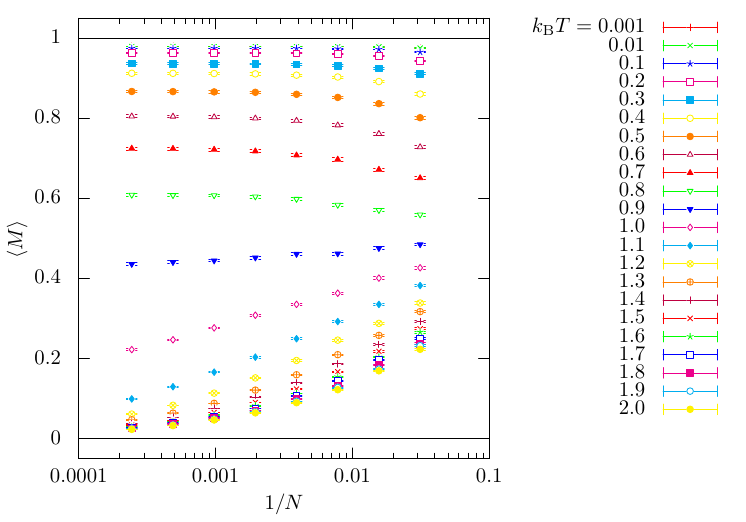}
	\caption{System-size dependence of the magnetization, Eq.~\eqref{supp_eq_def_magnetization_001_001}, at $q = 10^{-2.0}$. We consider $X \to YZ$ and set $K = 2$, $J = 1.0$, $t = 0$, $s = 0.9$, and $r_- = r_+ = 0.25$. We varied $N = 32, 64, 128, 256, 512, 1024, 2048, 4096$.}
	\label{supp_fig_system-size-dependence_magnetization_001_001}
\end{figure}

\begin{figure}[t]
	\centering
	\includegraphics[scale=0.60]{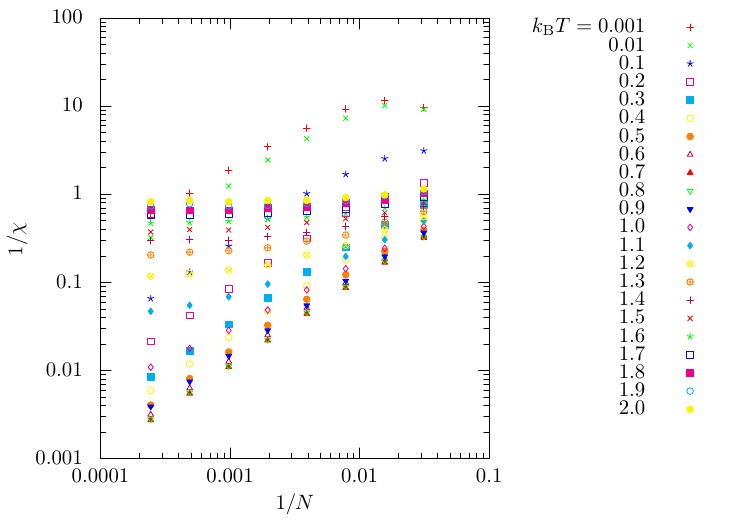}
	\includegraphics[scale=0.60]{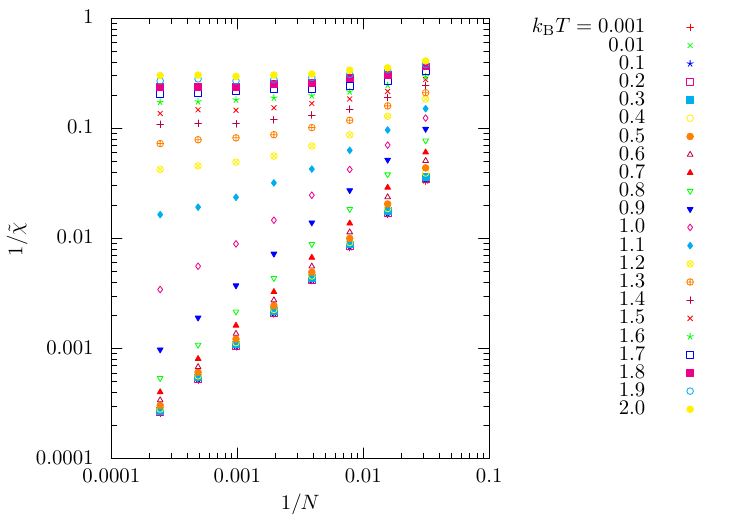}
	\caption{System-size dependence of the susceptibilities, (upper) Eq.~\eqref{supp_eq_def_specific_heat_001_001} and (lower) Eq.~\eqref{supp_eq_def_specific_heat_001_002} at $q = 10^{-2.0}$. We consider $X \to YZ$ and set $K = 2$, $J = 1.0$, $t = 0$, $s = 0.9$, and $r_- = r_+ = 0.25$. We varied $N = 32, 64, 128, 256, 512, 1024, 2048, 4096$.}
	\label{supp_fig_system-size-dependence_susceptibility_001_001}
\end{figure}

\begin{figure}[t]
	\centering
	\includegraphics[scale=0.60]{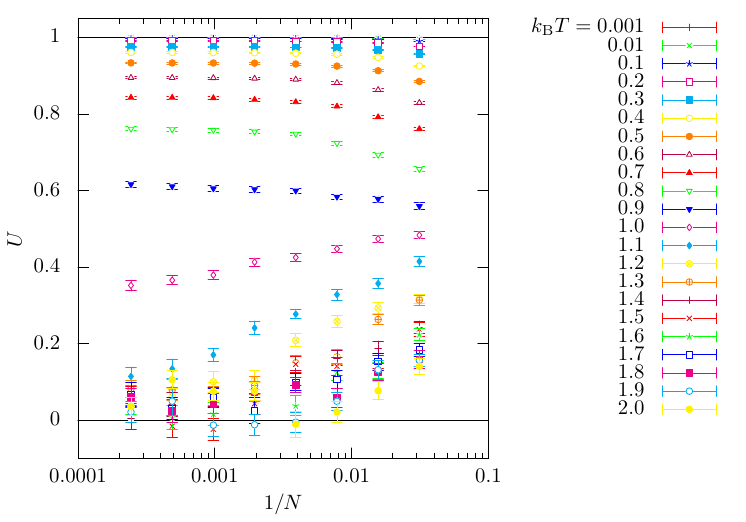}
	\caption{System-size dependence of the Binder parameter, Eq.~\eqref{supp_eq_def_Binder_parameter_001_001}, at $q = 10^{-2.0}$. We consider $X \to YZ$ and set $K = 2$, $J = 1.0$, $t = 0$, $s = 0.9$, and $r_- = r_+ = 0.25$. We varied $N = 32, 64, 128, 256, 512, 1024, 2048, 4096$.}
	\label{supp_fig_system-size-dependence_Binder_001_001}
\end{figure}

In Figs.~\ref{supp_fig_system-size-dependence_magnetization_001_001}, \ref{supp_fig_system-size-dependence_susceptibility_001_001}, and \ref{supp_fig_system-size-dependence_Binder_001_001}, we show the system-size dependence of the magnetization, Eq.~\eqref{supp_eq_def_magnetization_001_001}, the two susceptibilities, Eqs.~\eqref{supp_eq_def_specific_heat_001_001} and \eqref{supp_eq_def_specific_heat_001_002}, and the Binder parameter, Eq.~\eqref{supp_eq_def_Binder_parameter_001_001}, at $q = 10^{-2.0}$, respectively.
The critical temperature was determined by assessing whether the line of the inverse of the susceptibilities in Fig.~\ref{supp_fig_system-size-dependence_susceptibility_001_001} is linear or not.
This estimate is consistent with the temperature at which the Binder parameter takes a non-zero value in the thermodynamic limit, as shown in Fig.~\ref{supp_fig_system-size-dependence_Binder_001_001}.

\subsubsection{Phase diagram, finite-size scaling, and $q$-dependence of critical exponents}

\begin{figure}[t]
	\centering
	\includegraphics[scale=0.60]{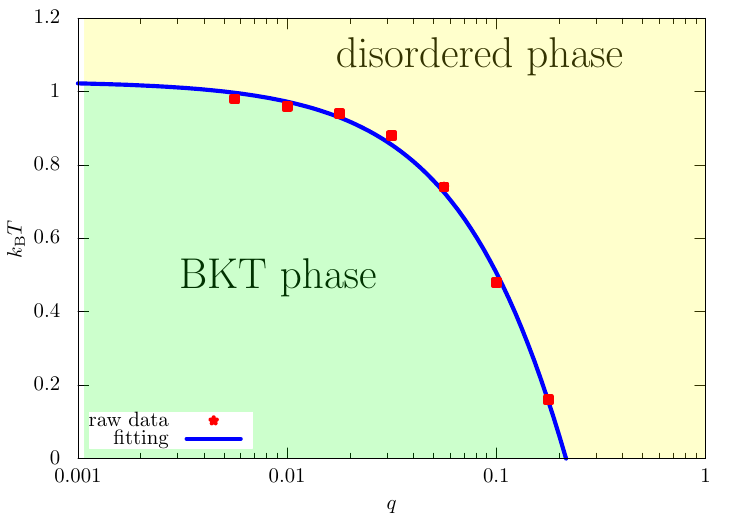}
	\caption{Phase diagram of the context-sensitive random language model where the horizontal and vertical axes are the growth rate of a sentence $q$ and temperature $k_\mathrm{B} T$, respectively. We consider $X \to YZ$ and set $K = 2$, $J = 1.0$, $t = 0$, $s = 0.9$, and $r_- = r_+ = 0.25$. This figure is identical to Fig.~\ref{main_fig_phase_diagram_001_001}(upper).}
	\label{supp_fig_phase_diagram_001_001}
\end{figure}

In Fig.~\ref{supp_fig_phase_diagram_001_001}, we plot the phase diagram of our language model with $K = 2$ and $X \to YZ$.
We determined the boundary between the two phases by fitting a quadratic function via least squares regression.
This figure shows that the BKT phase exists for $k_\mathrm{B} T \lesssim 0.20$.

\begin{figure}[t]
	\centering
	\includegraphics[scale=0.60]{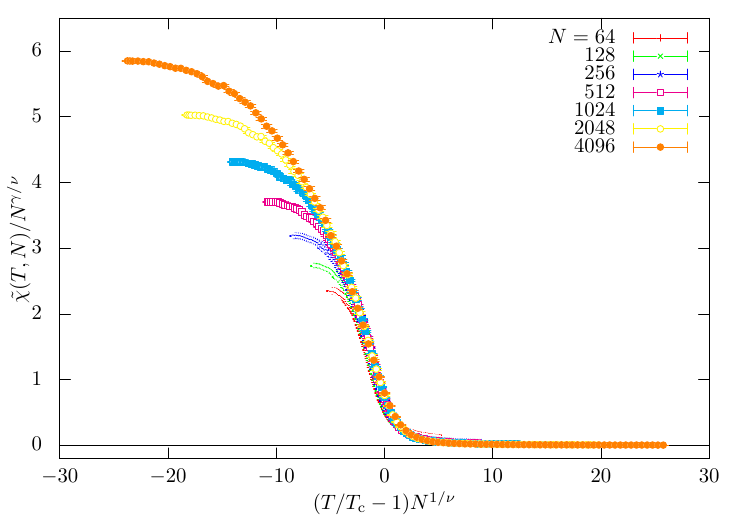}
	\includegraphics[scale=0.60]{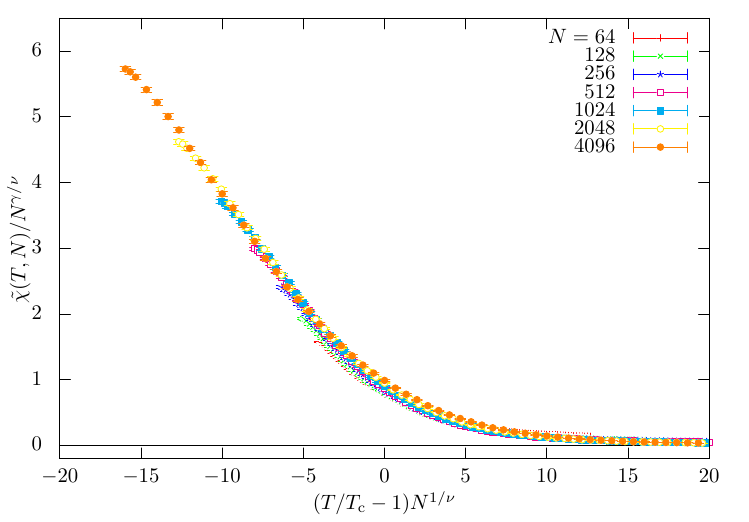}
	\caption{Finite-size scaling of $\tilde{\chi}$ at (upper) $q = 10^{-2.0}$ and (lower) $q = 10^{-1.0}$. We set $T_\mathrm{c} = 0.960$, $\nu = 2.6250$, and $\gamma = 2.05$ for $q = 10^{-2.0}$ and $T_\mathrm{c} = 0.48$, $\nu = 3.0$, and $\gamma = 2.05$ for $q = 10^{-1.0}$, where the values of $\nu$ and $\gamma$ are determined such that the scaling assumption holds. We consider $X \to YZ$ and set $K = 2$, $J = 1.0$, $t = 0$, $s = 0.9$, and $r_- = r_+ = 0.25$. We varied $N = 64, 128, 256, 512, 1024, 2048, 4096$.  The upper panel is identical to Fig.~\ref{main_fig_phase_diagram_001_001}(middle).}
	\label{supp_fig_finite-size-scaling_001_001}
\end{figure}

In Fig.~\ref{supp_fig_finite-size-scaling_001_001}, we perform finite-size scaling on $\tilde{\chi}$ at $q = 10^{-2.0}$.
We set $T_\mathrm{c} = 0.960$, $\nu = 2.6250$, and $\gamma = 2.05$, where we determine the values of $\nu$ and $\gamma$ such that the scaling assumption holds; this figure shows that the scaling assumption holds.

\begin{figure}[t]
	\centering
	\includegraphics[scale=0.60]{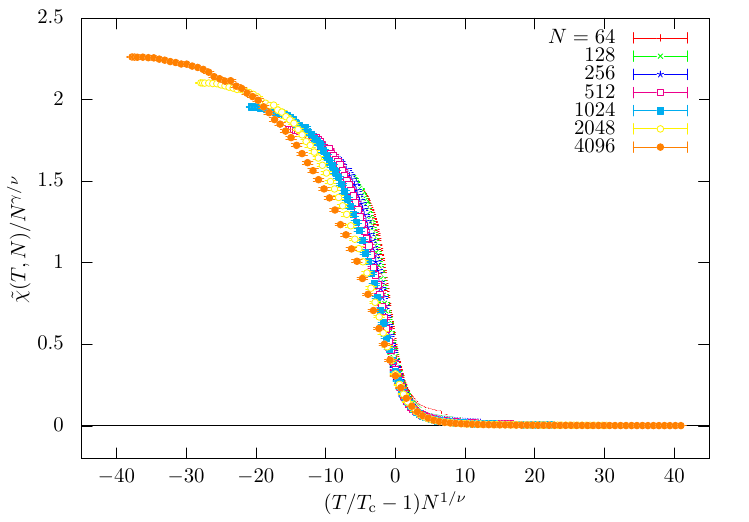}
	\includegraphics[scale=0.60]{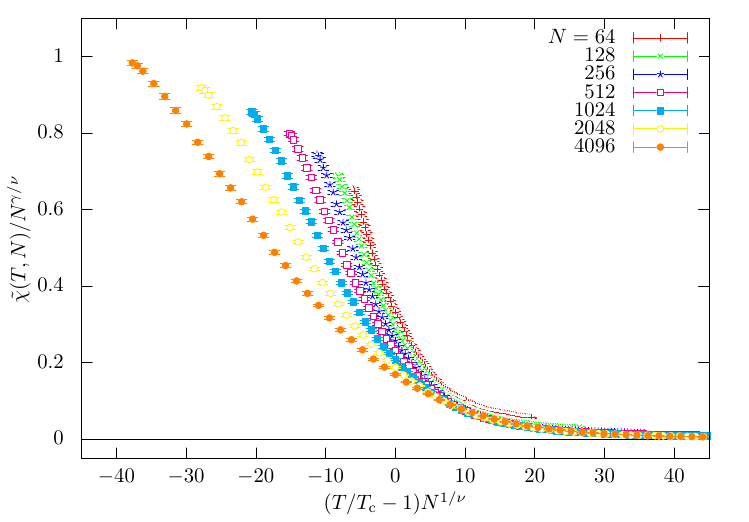}
	\caption{Finite-size scaling of $\tilde{\chi}$ at (upper) $q = 10^{-2.0}$ and (lower) $q = 10^{-1.0}$. We set $T_\mathrm{c} = 0.960$ for $q = 10^{-2.0}$ and $T_\mathrm{c} = 0.48$ for $q = 10^{-1.0}$. We also set $\nu = 2.29$, and $\gamma = 2.05$, where the values of $\nu$ and $\gamma$ are taken from Ref.~\cite{Tomita_001}. We consider $X \to YZ$ and set $K = 2$, $J = 1.0$, $t = 0$, $s = 0.9$, and $r_- = r_+ = 0.25$. We varied $N = 64, 128, 256, 512, 1024, 2048, 4096$.}
	\label{supp_fig_finite-size-scaling_001_002}
\end{figure}

In Fig.~\ref{supp_fig_finite-size-scaling_001_002}, we set $T_\mathrm{c} = 0.960$, $\nu = 2.29$, and $\gamma = 2.05$, where $\nu$ and $\gamma$ are taken from Ref.~\cite{Tomita_001}; the critical exponents are different from those of the one-dimensional long-range Ising model.
More specifically, the critical exponents depend on the value of $q$.

\begin{figure}[t]
	\centering
	\includegraphics[scale=0.60]{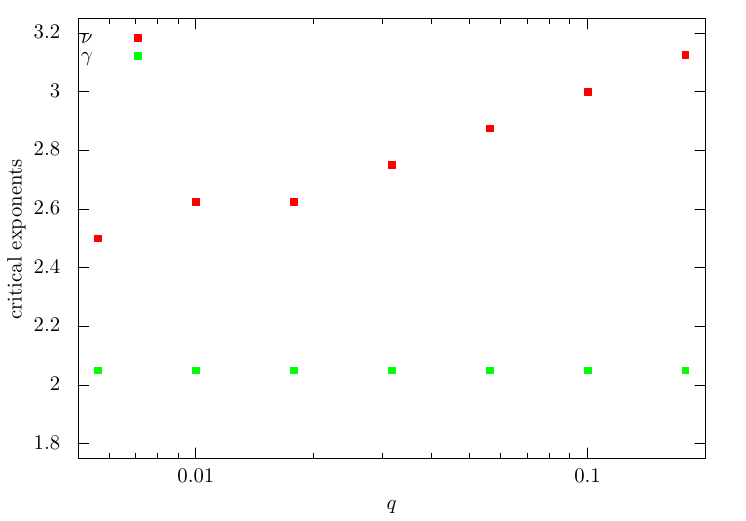}
	\caption{$q$-dependence of the critical exponents $\nu$ and $\gamma$. We consider $X \to YZ$ and set $K = 2$, $J = 1.0$, $t = 0$, $s = 0.9$, and $r_- = r_+ = 0.25$.  This figure is identical to Fig.~\ref{main_fig_phase_diagram_001_001}(lower).}
	\label{supp_fig_q_dependence_critical_exponents_001_001}
\end{figure}

In Fig.~\ref{supp_fig_q_dependence_critical_exponents_001_001}, we plot the $q$-dependence of the critical exponents $\nu$ and $\gamma$.
By observing these robust critical phenomena over a wide range of the parameter space, we conclude that the BKT transition is occurring in the proposed model, Eq.~\eqref{supp_eq_rule_001_001}, rather than a second-order phase transition.

\subsection{Case of $K=2$ and $X \to XX$}

We consider the case of $K = 2$ and $X \to XX$, in which symbols are duplicated.
Since there are two nonterminal symbols, there are two processes:
\begin{align}
	A \to AA, \quad B \to BB.
\end{align}
Furthermore, we assume that these two processes occur with equal probability.
Note that this model can be regarded as a generalization of the P\'olya urn model since Eq.~\eqref{supp_eq_rule_001_001} reduces to it in the limit of $t \to 0$ and $q \to 1$.

\subsubsection{Magnetization, susceptibilities, and Binder parameter}

\begin{figure}[t]
	\centering
	\includegraphics[scale=0.60]{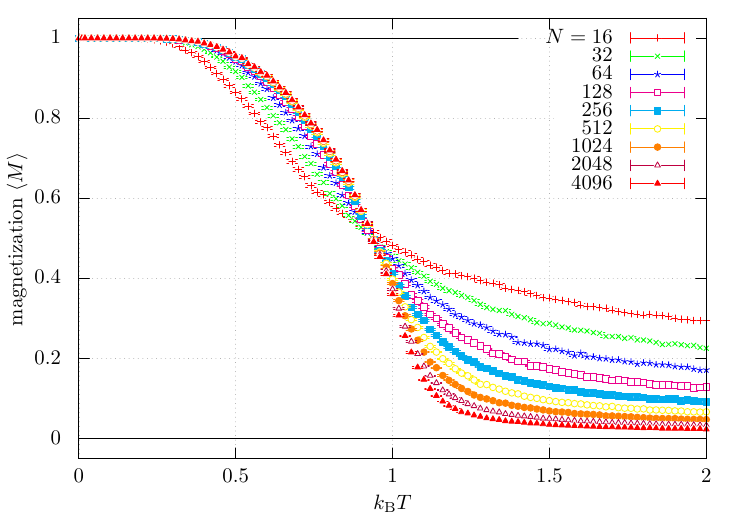}
	\includegraphics[scale=0.60]{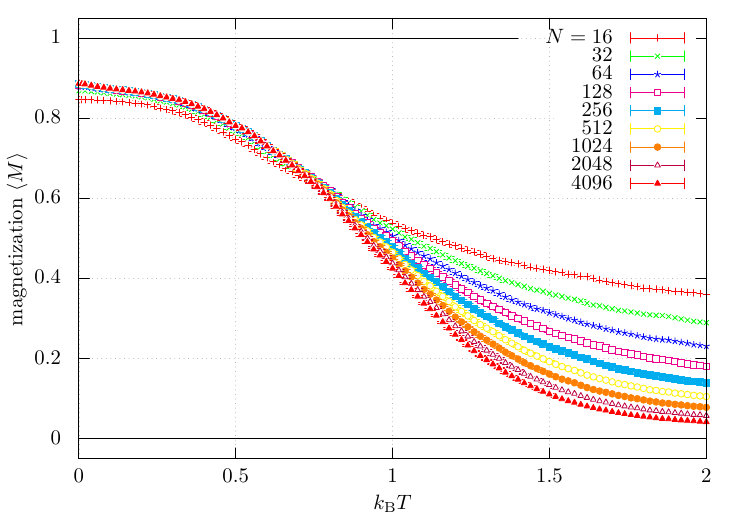}
	\caption{Temperature dependence of the magnetization, Eq.~\eqref{supp_eq_def_magnetization_001_001}, for $X \to XX$. We set (upper) $q = 10^{-2.0}$ and (lower) $q = 10^{-0.5}$. We also set $K = 2$, $J = 1.0$, $q = 10^{-0.5}$, $t = 0$, $s = 0.9$, and $r_- = r_+ = 0.25$. The length of the generated sentence by the language model, $N$, was varied from $16$ to $4096$.}
	\label{supp_fig_magnetization_K=2_X_XX_001_001}
\end{figure}

In Fig.~\ref{supp_fig_magnetization_K=2_X_XX_001_001}, the magnetization, Eq.~\eqref{supp_eq_def_magnetization_001_001}, for $q = 10^{-2.0}$ and $q = 10^{-0.5}$ is shown.
For $q = 10^{-2.0}$, the behavior of the magnetization, Eq.~\eqref{supp_eq_def_magnetization_001_001}, is similar to the case of $X \to YZ$ shown in Fig.~\ref{supp_fig_magnetization_K=2_X_YZ_001_001}.
However, the behavior of the magnetization, Eq.~\eqref{supp_eq_def_magnetization_001_001}, is quite different for $X \to XX$ and $X \to YZ$.
That is, in Fig.~\ref{supp_fig_magnetization_K=2_X_XX_001_001}(lower), the magnetization seems to have a nontrivial value for small $k_\mathrm{B} T$.

\begin{figure}[t]
	\centering
	\includegraphics[scale=0.60]{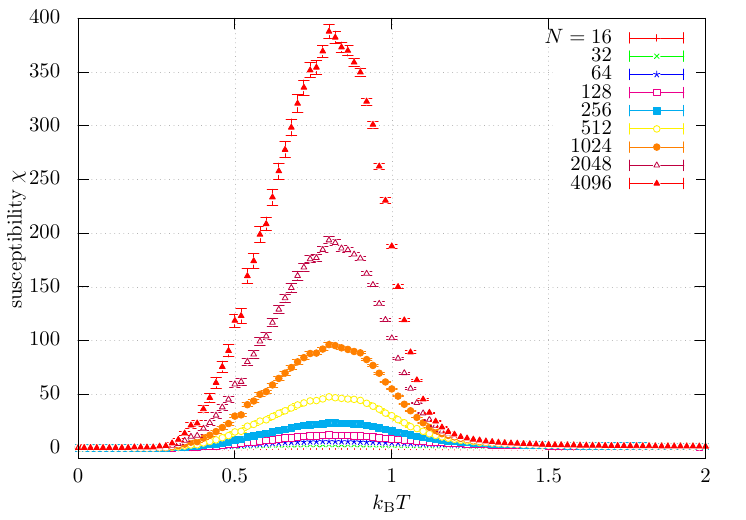}
	\includegraphics[scale=0.60]{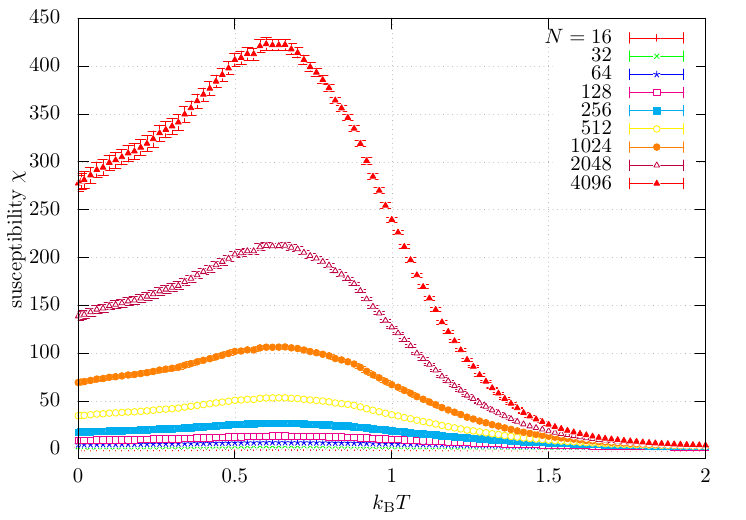}
	\caption{Temperature dependence of the susceptibility, Eq.~\eqref{supp_eq_def_specific_heat_001_001}, for $X \to XX$. We set (upper) $q = 10^{-2.0}$ and (lower) $q = 10^{-0.5}$. We also set $K = 2$, $J = 1.0$, $q = 10^{-0.5}$, $t = 0$, $s = 0.9$, and $r_- = r_+ = 0.25$. The length of the generated sentence by the language model, $N$, was varied from $16$ to $4096$.}
	\label{supp_fig_susceptibility_K=2_X_XX_001_001}
\end{figure}

In Fig.~\ref{supp_fig_susceptibility_K=2_X_XX_001_001}, we show the susceptibility, Eq.~\eqref{supp_eq_def_specific_heat_001_001}.
As expected, Fig.~\ref{supp_fig_susceptibility_K=2_X_XX_001_001} shows diverging behavior for finite temperatures in both cases.

\begin{figure}[t]
	\centering
	\includegraphics[scale=0.60]{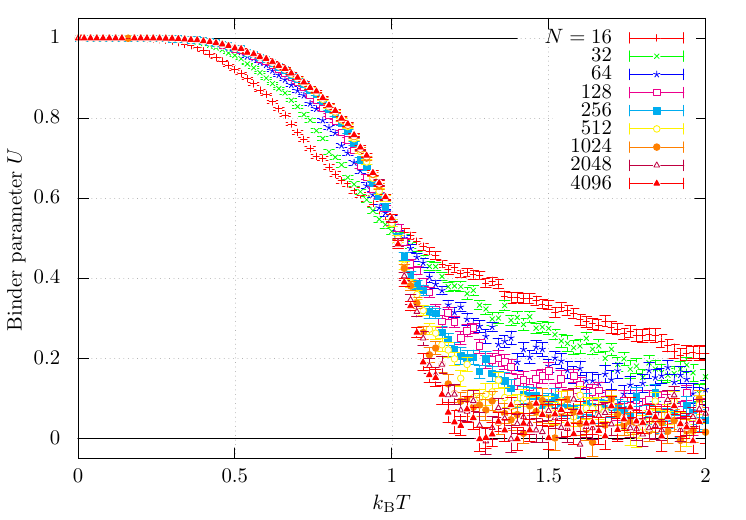}
	\includegraphics[scale=0.60]{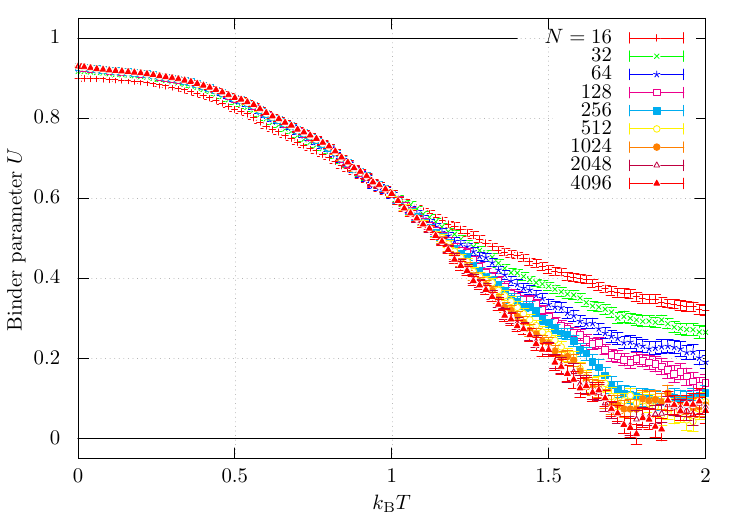}
	\caption{Temperature dependence of the Binder parameter, Eq.~\eqref{supp_eq_def_Binder_parameter_001_001}, for $X \to XX$. We set (upper) $q = 10^{-2.0}$ and (lower) $q = 10^{-0.5}$. We also set $K = 2$, $J = 1.0$, $q = 10^{-0.5}$, $t = 0$, $s = 0.9$, and $r_- = r_+ = 0.25$. The length of the generated sentence by the language model, $N$, was varied from $16$ to $4096$.}
	\label{supp_fig_Binder_parameter_K=2_X_XX_001_001}
\end{figure}

In Fig.~\ref{supp_fig_Binder_parameter_K=2_X_XX_001_001}, we plot the Binder parameter.
In Fig.~\ref{supp_fig_Binder_parameter_K=2_X_XX_001_001}, we can observe the crossings of the Binder parameter.
This fact implies that, in the case of $X \to XX$, there exists second-order phase transitions or BKT transitions for a wide range of $q$.

\subsubsection{Correlation functions, mutual information, and histogram of magnetization for $q = 10^{-2.0}$}

\begin{figure}[t]
	\centering
	\includegraphics[scale=0.60]{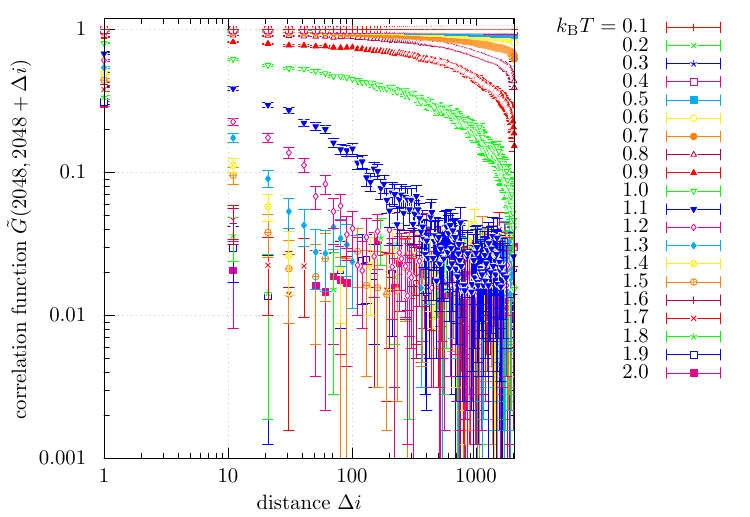}
	\includegraphics[scale=0.60]{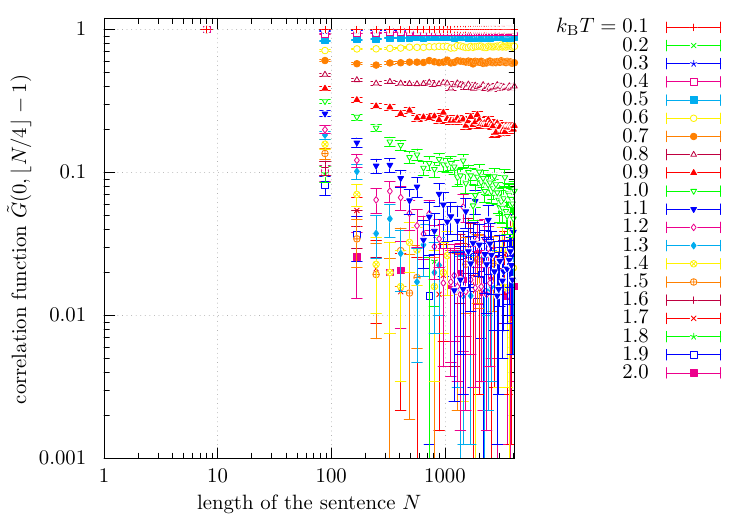}
	\caption{Correlation functions, Eq.~\eqref{supp_eq_correlation_function_with_disconnected_diagram_Potts_001_001}, (upper) with $i=2048$ and $j=2048 + \Delta i$ and (lower) with $i = 0$ and $j = \lfloor N / 4 \rfloor - 1$ for $X \to XX$. We set $K = 2$, $J = 1.0$, $q = 10^{-2.0}$, $t = 0$, $s = 0.9$, and $r_- = r_+ = 0.25$. The temperature $k_\mathrm{B} T$ was varied from $0.1$ to $2.0$.}
	\label{supp_fig_correlation_function_K=02_q=10^(-2.0)_X_XX_001_001}
\end{figure}

\begin{figure}[t]
	\centering
	\includegraphics[scale=0.60]{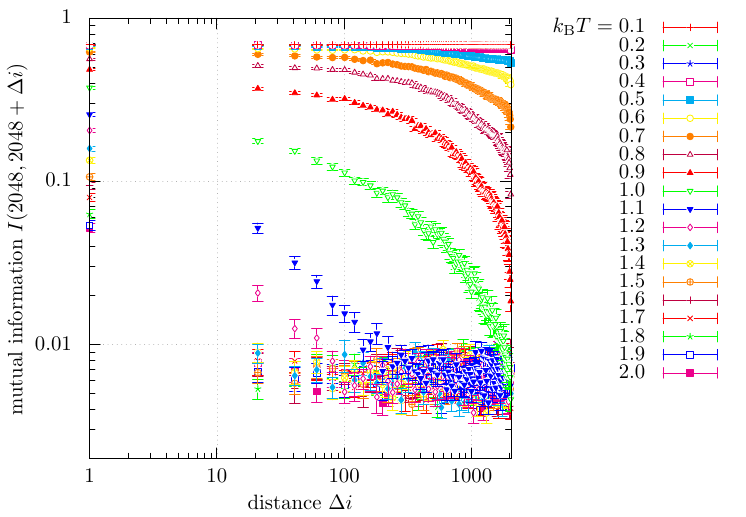}
	\caption{Mutual information, Eq.~\eqref{supp_eq_mutual_information_Potts_001_001}, with $i=2048$ and $j=2048 + \Delta i$ for $X \to XX$. We set $K = 2$, $J = 1.0$, $q = 10^{-2.0}$, $t = 0$, $s = 0.9$, and $r_- = r_+ = 0.25$. The temperature $k_\mathrm{B} T$ was varied from $0.1$ to $2.0$.}
	\label{supp_fig_mutual_information_K=02_q=10^(-2.0)_X_XX_001_001}
\end{figure}

\begin{figure}[t]
	\centering
	\includegraphics[scale=0.60]{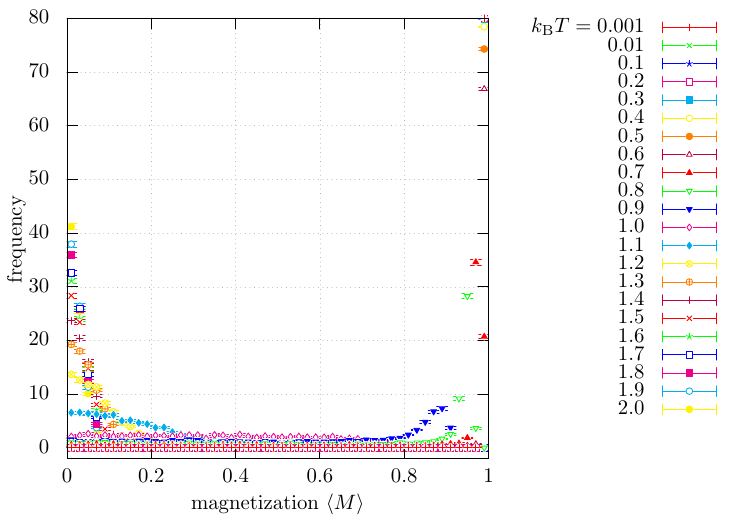}
	\caption{Histogram of the magnetization, Eq.~\eqref{supp_eq_def_magnetization_001_001}, for $X \to XX$. We set $K = 2$, $J = 1.0$, $q = 10^{-2.0}$, $t = 0$, $s = 0.9$, and $r_- = r_+ = 0.25$. The temperature $k_\mathrm{B} T$ was varied from $0.1$ to $2.0$.}
	\label{supp_fig_histogram_mag_K=02_q=10^(-2.0)_X_XX_001_001}
\end{figure}

In Figs.~\ref{supp_fig_correlation_function_K=02_q=10^(-2.0)_X_XX_001_001}, \ref{supp_fig_mutual_information_K=02_q=10^(-2.0)_X_XX_001_001}, and \ref{supp_fig_histogram_mag_K=02_q=10^(-2.0)_X_XX_001_001}, we show the correlation functions, Eq.~\eqref{supp_eq_correlation_function_with_disconnected_diagram_Potts_001_001}, mutual information, Eq.~\eqref{supp_eq_mutual_information_Potts_001_001}, and the histogram of the magnetization, Eq.~\eqref{supp_eq_histogram_magnetization_001_001}, respectively.
These figures are consistent with either second-order phase transitions or BKT transitions.

\subsubsection{Correlation functions, mutual information, and histogram of magnetization for $q = 10^{-0.5}$}

\begin{figure}[t]
	\centering
	\includegraphics[scale=0.60]{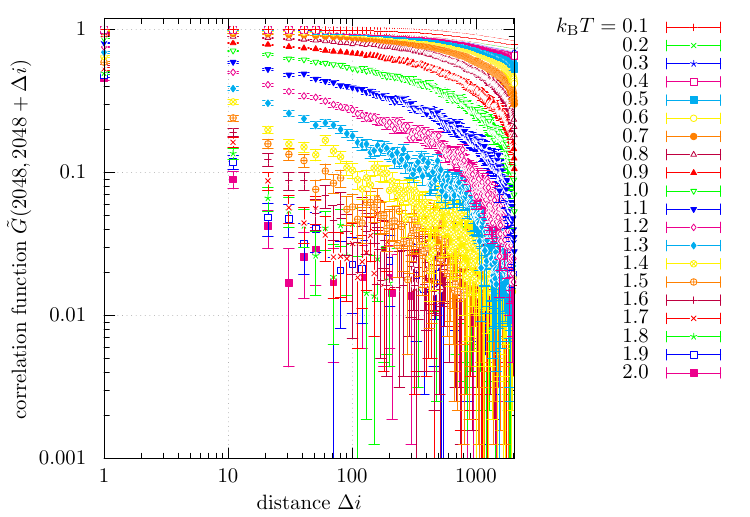}
	\includegraphics[scale=0.60]{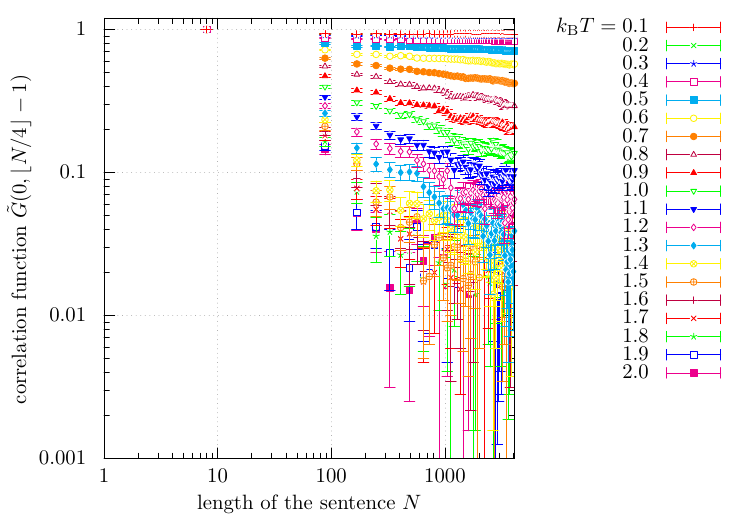}
	\caption{Correlation functions, Eq.~\eqref{supp_eq_correlation_function_with_disconnected_diagram_Potts_001_001}, (upper) with $i=2048$ and $j=2048 + \Delta i$ and (lower) with $i = 0$ and $j = \lfloor N / 4 \rfloor - 1$ for $X \to XX$. We set $K = 2$, $J = 1.0$, $q = 10^{-0.5}$, $t = 0$, $s = 0.9$, and $r_- = r_+ = 0.25$. The temperature $k_\mathrm{B} T$ was varied from $0.1$ to $2.0$.}
	\label{supp_fig_correlation_function_K=02_q=10^(-0.5)_X_XX_001_001}
\end{figure}

\begin{figure}[t]
	\centering
	\includegraphics[scale=0.60]{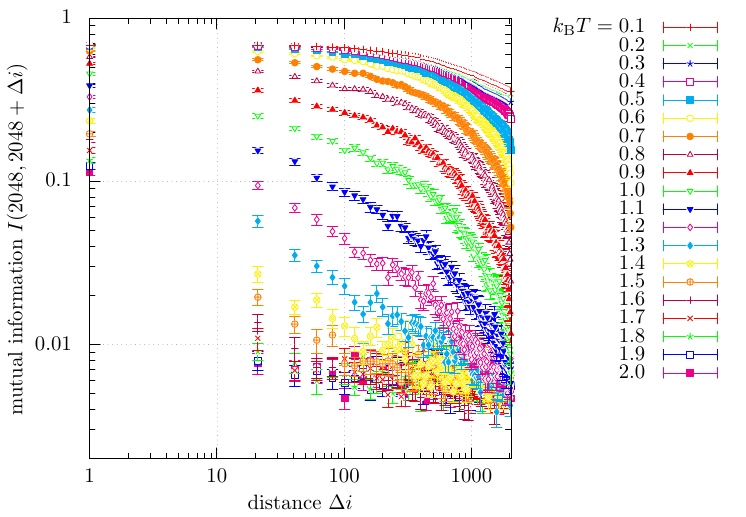}
	\caption{Mutual information, Eq.~\eqref{supp_eq_mutual_information_Potts_001_001}, with $i=2048$ and $j=2048 + \Delta i$ for $X \to XX$. We set $K = 2$, $J = 1.0$, $q = 10^{-0.5}$, $t = 0$, $s = 0.9$, and $r_- = r_+ = 0.25$. The temperature $k_\mathrm{B} T$ was varied from $0.1$ to $2.0$.}
	\label{supp_fig_mutual_information_K=02_q=10^(-0.5)_X_XX_001_001}
\end{figure}

\begin{figure}[t]
	\centering
	\includegraphics[scale=0.60]{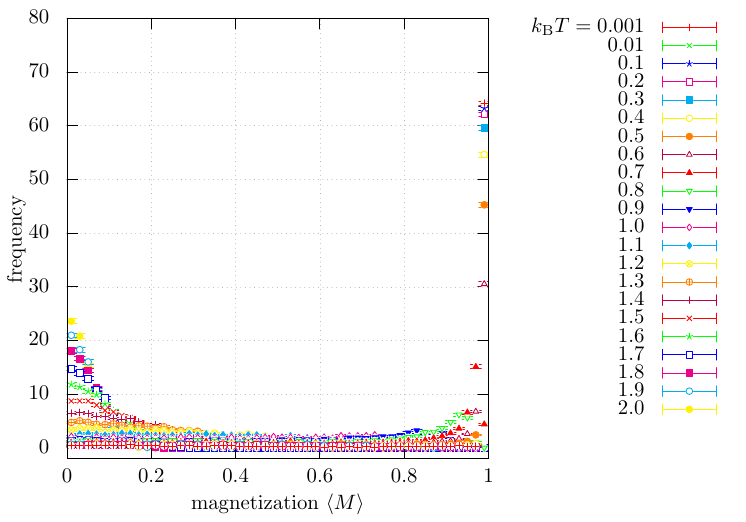}
	\caption{Histogram of the magnetization, Eq.~\eqref{supp_eq_def_magnetization_001_001}, for $X \to XX$. We set $K = 2$, $J = 1.0$, $q = 10^{-0.5}$, $t = 0$, $s = 0.9$, and $r_- = r_+ = 0.25$. The temperature $k_\mathrm{B} T$ was varied from $0.1$ to $2.0$.}
	\label{supp_fig_histogram_mag_K=02_q=10^(-0.5)_X_XX_001_001}
\end{figure}

In Figs.~\ref{supp_fig_correlation_function_K=02_q=10^(-0.5)_X_XX_001_001}, \ref{supp_fig_mutual_information_K=02_q=10^(-0.5)_X_XX_001_001}, and \ref{supp_fig_histogram_mag_K=02_q=10^(-0.5)_X_XX_001_001}, we show the correlation functions, Eq.~\eqref{supp_eq_correlation_function_with_disconnected_diagram_Potts_001_001}, mutual information, Eq.~\eqref{supp_eq_mutual_information_Potts_001_001}, and the histogram of the magnetization, Eq.~\eqref{supp_eq_histogram_magnetization_001_001}, respectively.
Similar to the case of $q = 10^{-2.0}$, these figures are also consistent with either second-order phase transitions or BKT transitions.

\subsubsection{System-size dependence of the magnetization, the susceptibilities, and the Binder parameter}

\begin{figure}[t]
	\centering
	\includegraphics[scale=0.60]{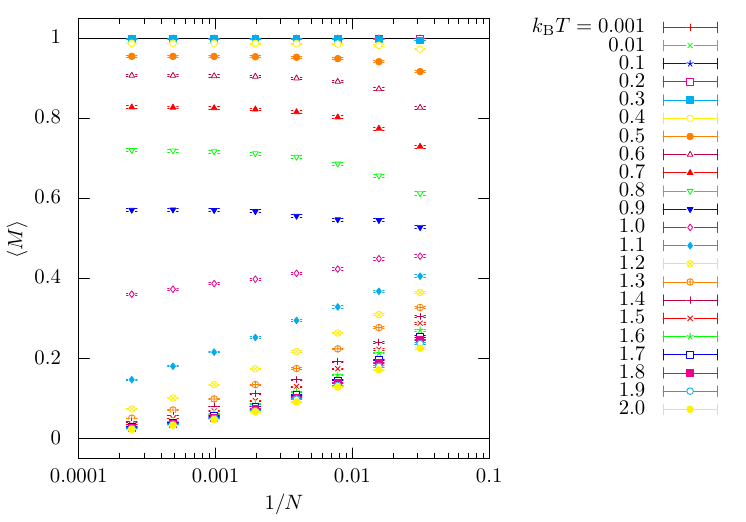}
	\caption{System-size dependence of the magnetization, Eq.~\eqref{supp_eq_def_magnetization_001_001}, at $q = 10^{-2.0}$. We consider $X \to XX$ and set $K = 2$, $J = 1.0$, $t = 0$, $s = 0.9$, and $r_- = r_+ = 0.25$. We varied $N = 32, 64, 128, 256, 512, 1024, 2048, 4096$.}
	\label{supp_fig_system-size-dependence_magnetization_001_002}
\end{figure}

\begin{figure}[t]
	\centering
	\includegraphics[scale=0.60]{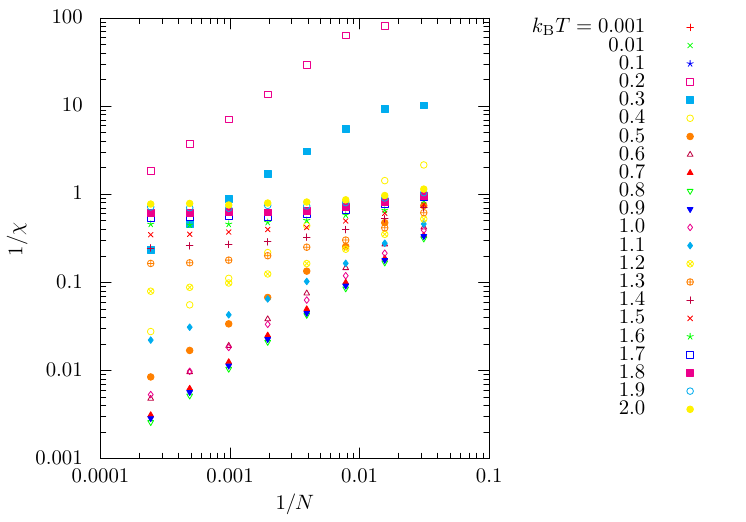}
	\includegraphics[scale=0.60]{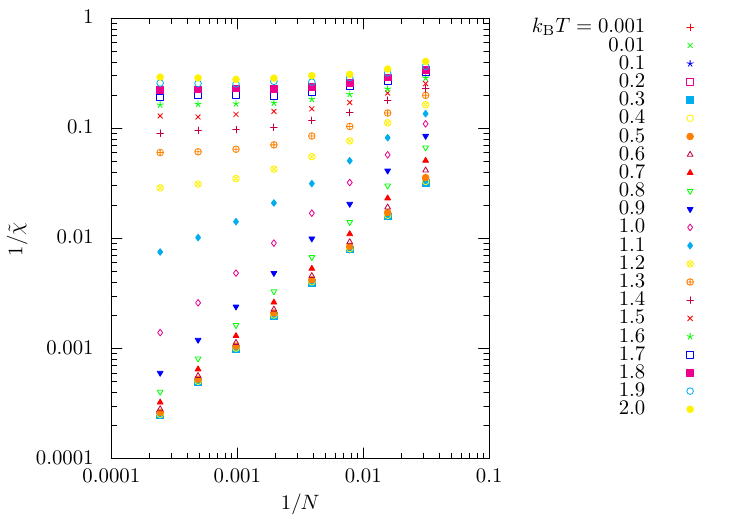}
	\caption{System-size dependence of the susceptibilities, (upper) Eq.~\eqref{supp_eq_def_specific_heat_001_001} and (lower) Eq.~\eqref{supp_eq_def_specific_heat_001_002} at $q = 10^{-2.0}$. We consider $X \to XX$ and set $K = 2$, $J = 1.0$, $t = 0$, $s = 0.9$, and $r_- = r_+ = 0.25$. We varied $N = 32, 64, 128, 256, 512, 1024, 2048, 4096$.}
	\label{supp_fig_system-size-dependence_susceptibility_001_002}
\end{figure}

\begin{figure}[t]
	\centering
	\includegraphics[scale=0.60]{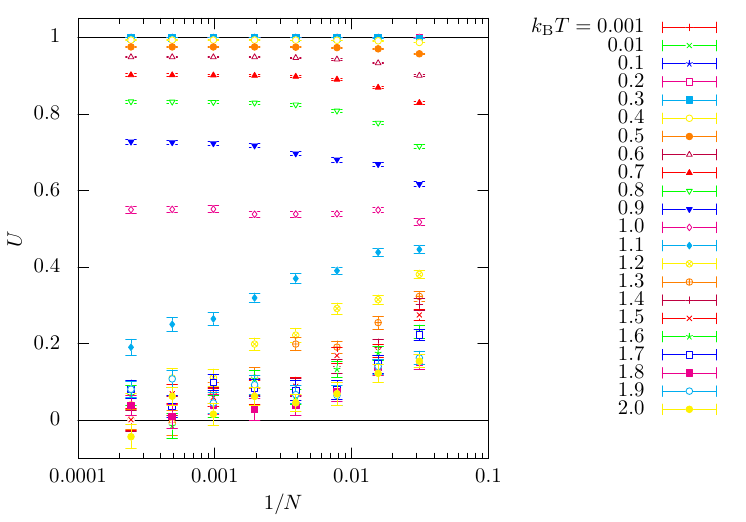}
	\caption{System-size dependence of the Binder parameter, Eq.~\eqref{supp_eq_def_Binder_parameter_001_001}, at $q = 10^{-2.0}$. We consider $X \to XX$ and set $K = 2$, $J = 1.0$, $t = 0$, $s = 0.9$, and $r_- = r_+ = 0.25$. We varied $N = 32, 64, 128, 256, 512, 1024, 2048, 4096$.}
	\label{supp_fig_system-size-dependence_Binder_001_002}
\end{figure}

In Figs.~\ref{supp_fig_system-size-dependence_magnetization_001_002}, \ref{supp_fig_system-size-dependence_susceptibility_001_002}, and \ref{supp_fig_system-size-dependence_Binder_001_002}, we show the system-size dependence of the magnetization, Eq.~\eqref{supp_eq_def_magnetization_001_001}, the two susceptibilities, Eqs.~\eqref{supp_eq_def_specific_heat_001_001} and \eqref{supp_eq_def_specific_heat_001_002}, and the Binder parameter, Eq.~\eqref{supp_eq_def_Binder_parameter_001_001}, at $q = 10^{-2.0}$, respectively.
The critical temperature was again determined by assessing whether the line is linear or not.
This estimate is consistent with the temperature at which the Binder parameter takes a non-zero value in the thermodynamic limit, as shown in Fig.~\ref{supp_fig_system-size-dependence_Binder_001_002}.

\subsubsection{Phase diagram, finite-size scaling, and $q$-dependence of critical exponents}

\begin{figure}[t]
	\centering
	\includegraphics[scale=0.60]{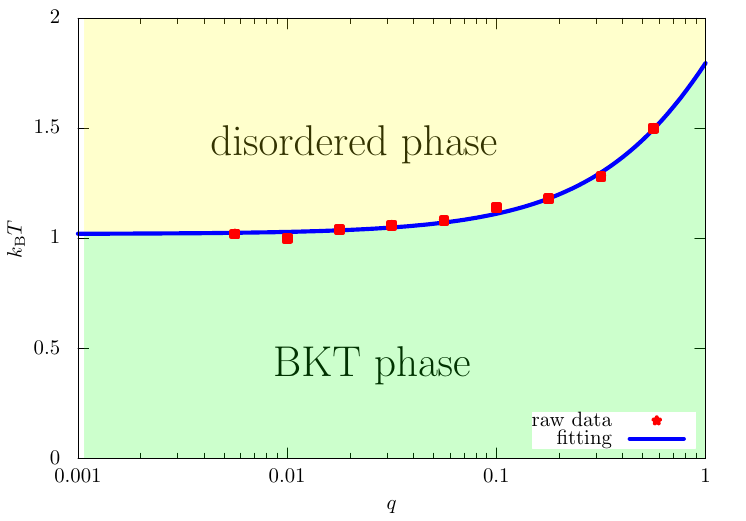}
	\caption{Phase diagram of the context-sensitive random language model, where the horizontal and vertical axes are the growth rate of a sentence $q$ and temperature $k_\mathrm{B} T$, respectively. We consider $X \to XX$ and set $K = 2$, $J = 1.0$, $t = 0$, $s = 0.9$, and $r_- = r_+ = 0.25$.}
	\label{supp_fig_phase_diagram_002_001}
\end{figure}

In Fig.~\ref{supp_fig_phase_diagram_002_001}, we plot the phase diagram of our language model with $K = 2$ and $X \to XX$.
We determined the boundary between the two phases by fitting a quadratic function via least squares regression.
In the case of $X \to XX$, the regime of the BKT phase is drastically expanded, especially for large $q$ compared with the case of $X \to YZ$.

\begin{figure}[t]
	\centering
	\includegraphics[scale=0.60]{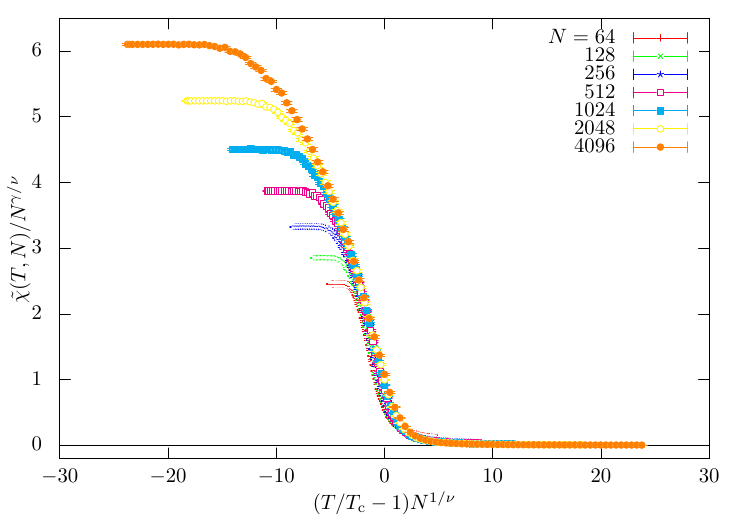}
	\includegraphics[scale=0.60]{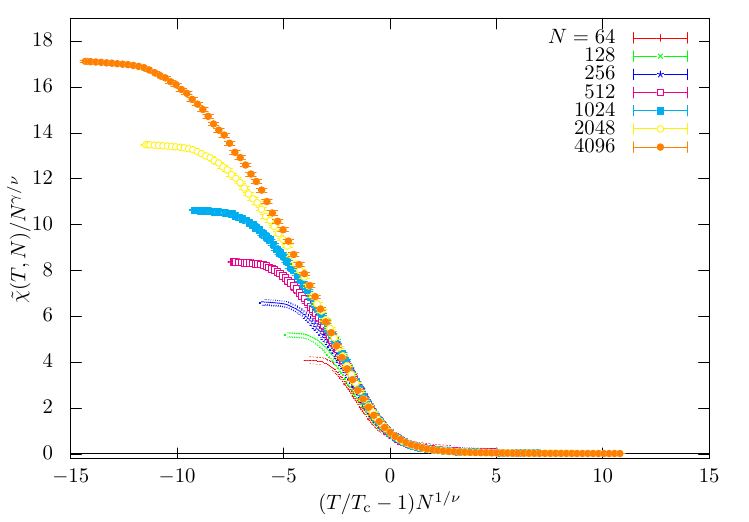}
	\caption{Finite-size scaling of $\tilde{\chi}$ at (upper) $q = 10^{-2.0}$ and (lower) $q = 10^{-1.0}$. We set $T_\mathrm{c} = 1.0$, $\nu = 2.6250$, and $\gamma = 2.05$ for $q = 10^{-2.0}$ and $T_\mathrm{c} = 1.14$, $\nu = 3.1250$, and $\gamma = 2.05$ for $q = 10^{-1.0}$, where the values of $\nu$ and $\gamma$ are determined such that the scaling assumption holds. We consider $X \to XX$ and set $K = 2$, $J = 1.0$, $t = 0$, $s = 0.9$, and $r_- = r_+ = 0.25$. We varied $N = 64, 128, 256, 512, 1024, 2048, 4096$.}
	\label{supp_fig_finite-size-scaling_002_001}
\end{figure}

\begin{figure}[t]
	\centering
	\includegraphics[scale=0.60]{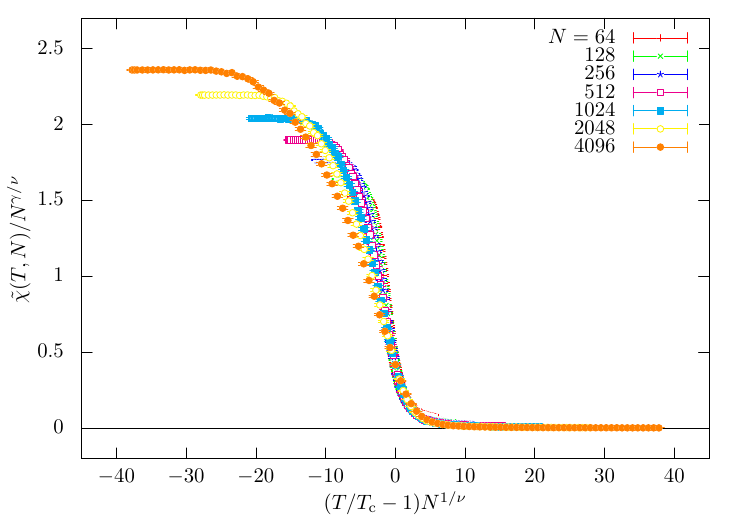}
	\includegraphics[scale=0.60]{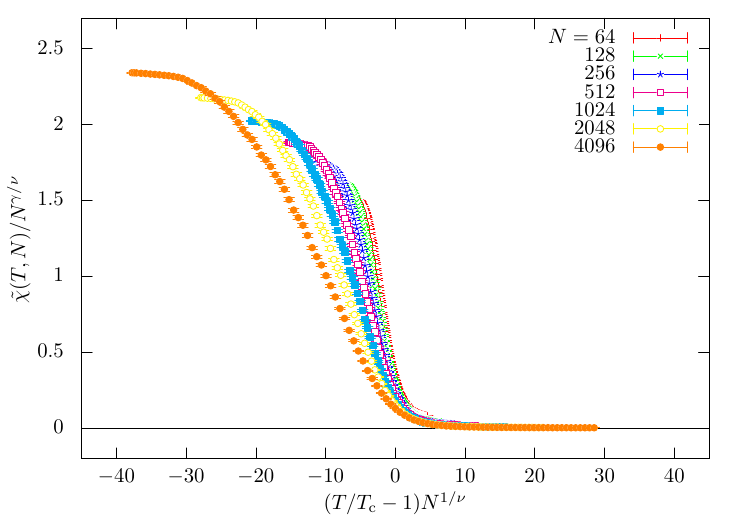}
	\caption{Finite-size scaling of $\tilde{\chi}$ at (upper) $q = 10^{-2.0}$ and (lower) $q = 10^{-1.0}$. We set $T_\mathrm{c} = 1.0$ for $q = 10^{-2.0}$ and $T_\mathrm{c} = 1.14$ for $q = 10^{-1.0}$. We also set $\nu = 2.29$, and $\gamma = 2.05$, where the values of $\nu$ and $\gamma$ are taken from Ref.~\cite{Tomita_001}. We consider $X \to XX$ and set $K = 2$, $J = 1.0$, $t = 0$, $s = 0.9$, and $r_- = r_+ = 0.25$. We varied $N = 64, 128, 256, 512, 1024, 2048, 4096$.}
	\label{supp_fig_finite-size-scaling_002_002}
\end{figure}

In Fig.~\ref{supp_fig_finite-size-scaling_002_001}, we perform finite-size scaling on $\tilde{\chi}$ at $q = 10^{-2.0}$.
We set $T_\mathrm{c} = 1.0$, $\nu = 2.6250$, and $\gamma = 2.05$, where we determine the values of $\nu$ and $\gamma$ such that the scaling assumption holds.
This figure shows that the scaling assumption holds.
In Fig.~\ref{supp_fig_finite-size-scaling_002_002}, we set $T_\mathrm{c} = 1.0$, $\nu = 2.29$, and $\gamma = 2.05$, where $\nu$ and $\gamma$ are taken from Ref.~\cite{Tomita_001}.
Similar to the case of $X \to YZ$, the critical exponents depend on the value of $q$.

\begin{figure}[t]
	\centering
	\includegraphics[scale=0.60]{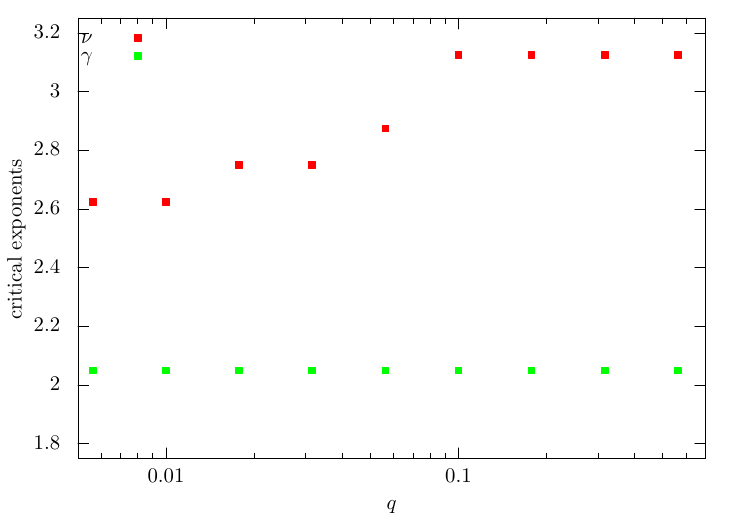}
	\caption{$q$-dependence of the critical exponents $\nu$ and $\gamma$. We consider $X \to XX$ and set $K = 2$, $J = 1.0$, $t = 0$, $s = 0.9$, and $r_- = r_+ = 0.25$.}
	\label{supp_fig_q_dependence_critical_exponents_002_001}
\end{figure}

In Fig.~\ref{supp_fig_q_dependence_critical_exponents_002_001}, we plot the $q$-dependence of the critical exponents $\nu$ and $\gamma$.
As already discussed above, we conclude that the BKT transition occurs since critical phenomena are observed over a wide range of the parameter space.

\subsection{Case of $K=10$ and $X \to YZ$}

Here we consider the case of $K=10$ and $X \to YZ$, where $X$ is replaced by two distinct symbols $Y$ and $Z$.

\subsubsection{Magnetization, susceptibilities, and Binder parameter}

\begin{figure}[t]
	\centering
	\includegraphics[scale=0.60]{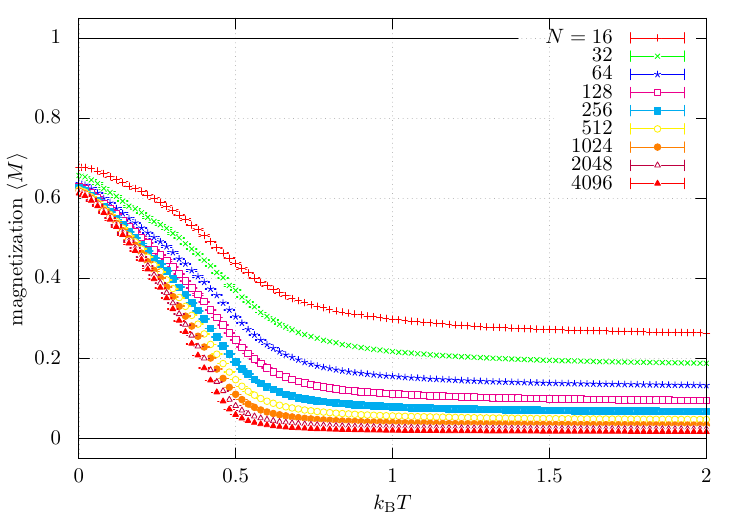}
	\includegraphics[scale=0.60]{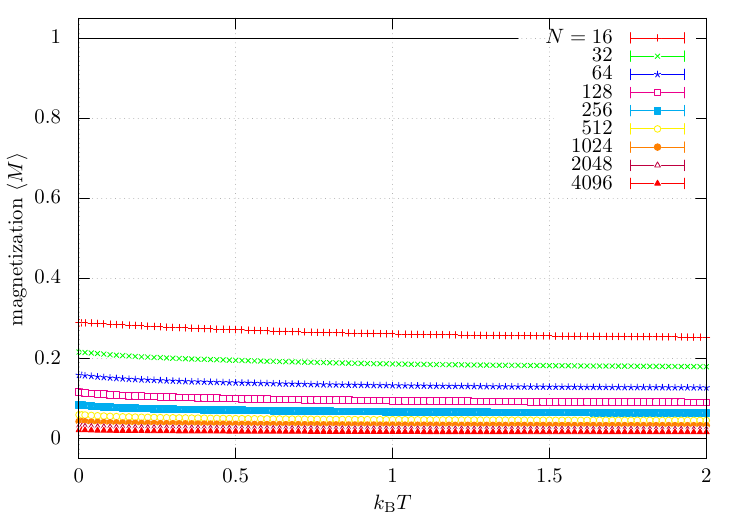}
	\caption{Temperature dependence of the magnetization, Eq.~\eqref{supp_eq_def_magnetization_001_001}, for $X \to YZ$. We set (upper) $q = 10^{-2.0}$ and (lower) $q = 10^{-0.5}$. We also set $K = 10$, $J = 1.0$, $t = 0$, $s = 0.9$, and $r_- = r_+ = 0.25$. The length of the generated sentence by the language model, $N$, was varied from $16$ to $4096$.}
	\label{supp_fig_magnetization_K=10_X_YZ_001_001}
\end{figure}

In Fig.~\ref{supp_fig_magnetization_K=10_X_YZ_001_001}, we plot the magnetization, Eq.~\eqref{supp_eq_def_magnetization_001_001}.
Figure~\ref{supp_fig_magnetization_K=10_X_YZ_001_001} shows that the magnetization grows for small $k_\mathrm{B} T$ in the case of $q = 10^{-2.0}$, but it remains in a trivial state for both $q = 10^{-2.0}$ and $q = 10^{-0.5}$.

\begin{figure}[t]
	\centering
	\includegraphics[scale=0.60]{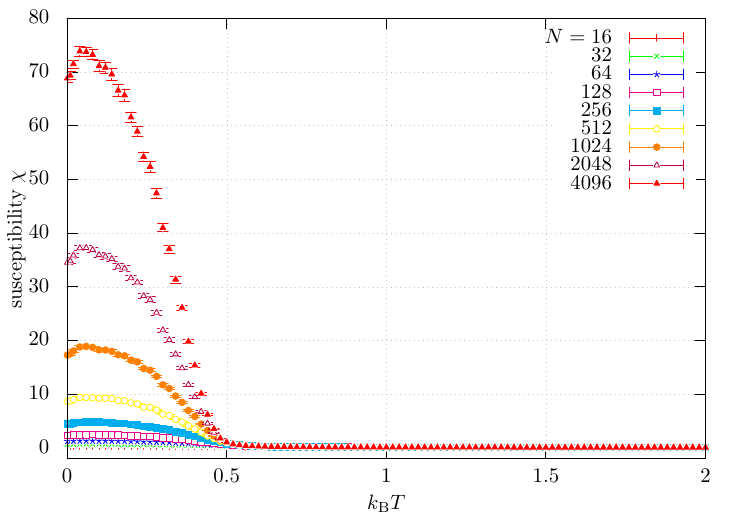}
	\includegraphics[scale=0.60]{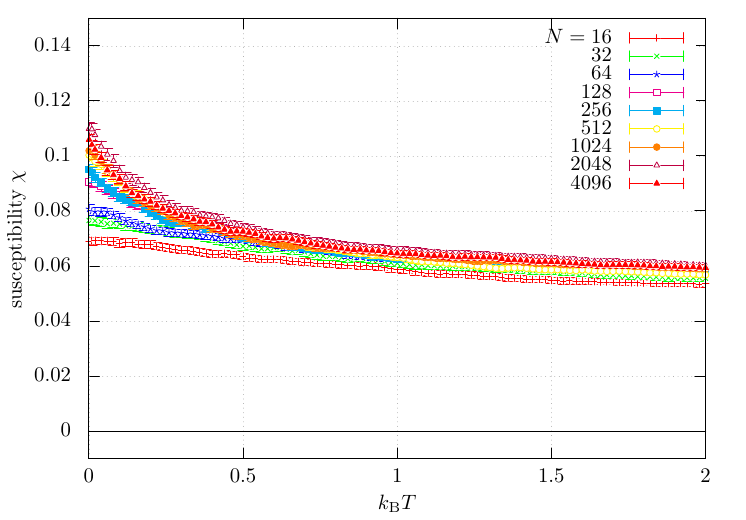}
	\caption{Temperature dependence of the susceptibility, Eq.~\eqref{supp_eq_def_specific_heat_001_001}, for $X \to YZ$. We set (upper) $q = 10^{-2.0}$ and (lower) $q = 10^{-0.5}$. We also set $K = 10$, $J = 1.0$, $t = 0$, $s = 0.9$, and $r_- = r_+ = 0.25$. The length of the generated sentence by the language model, $N$, was varied from $16$ to $4096$.}
	\label{supp_fig_susceptibility_K=10_X_YZ_001_001}
\end{figure}

In Fig.~\ref{supp_fig_susceptibility_K=10_X_YZ_001_001}, we show the susceptibility, Eq.~\eqref{supp_eq_def_specific_heat_001_001}.
This figure is consistent with Figure~\ref{supp_fig_magnetization_K=10_X_YZ_001_001} because there is no peak for finite temperature while the susceptibility grows or even appears to diverge for $q = 10^{-2.0}$.

\begin{figure}[t]
	\centering
	\includegraphics[scale=0.60]{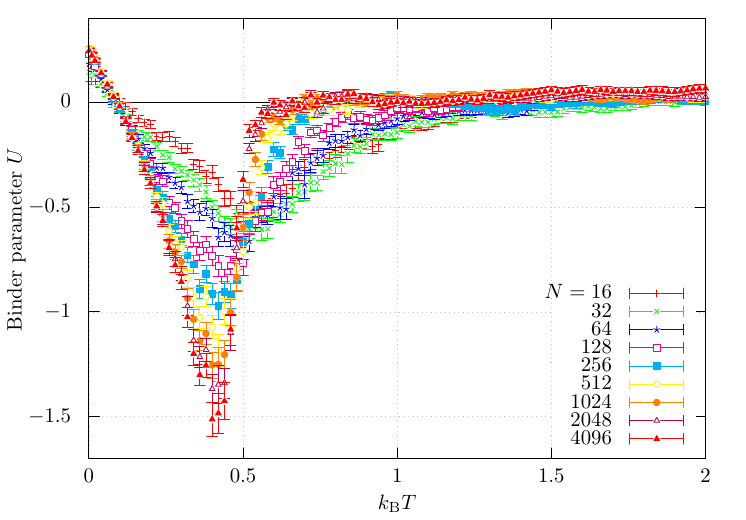}
	\includegraphics[scale=0.60]{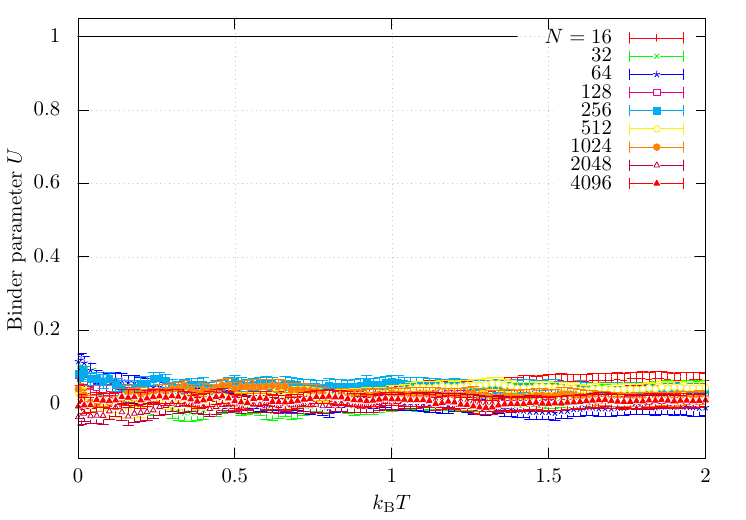}
	\caption{Temperature dependence of the Binder parameter, Eq.~\eqref{supp_eq_def_Binder_parameter_001_001}, for $X \to YZ$. We set (upper) $q = 10^{-2.0}$ and (lower) $q = 10^{-0.5}$. We also set $K = 10$, $J = 1.0$, $t = 0$, $s = 0.9$, and $r_- = r_+ = 0.25$. The length of the generated sentence by the language model, $N$, was varied from $16$ to $4096$.}
	\label{supp_fig_Binder_parameter_K=10_X_YZ_001_001}
\end{figure}

In Fig.~\ref{supp_fig_Binder_parameter_K=10_X_YZ_001_001}, we show the Binder parameter.
As expected, the Binder parameter does not go to 1 for small $k_\mathrm{B} T$, but we can see the dip for $q = 10^{-2.0}$.
This fact suggests that the phase transition occurring here is akin to the first-order phase transition at $T \to 0$, rather than the second-order phase transition.

\subsubsection{Correlation functions, mutual information, and histogram of magnetization for $q = 10^{-2.0}$}

\begin{figure}[t]
	\centering
	\includegraphics[scale=0.60]{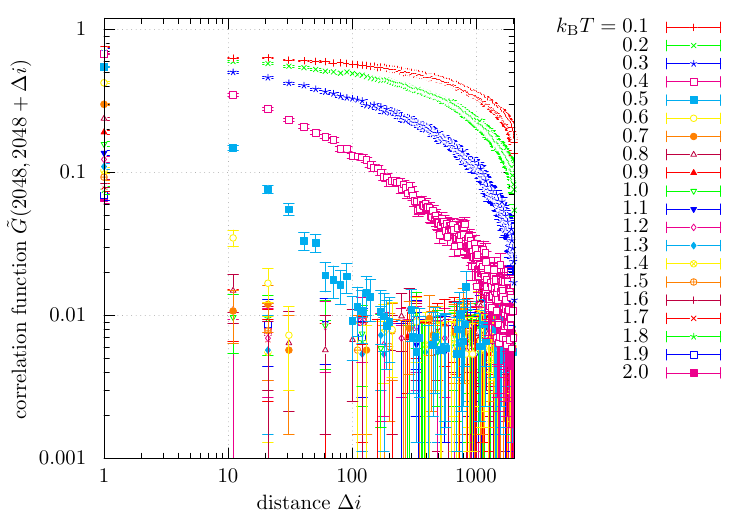}
	\includegraphics[scale=0.60]{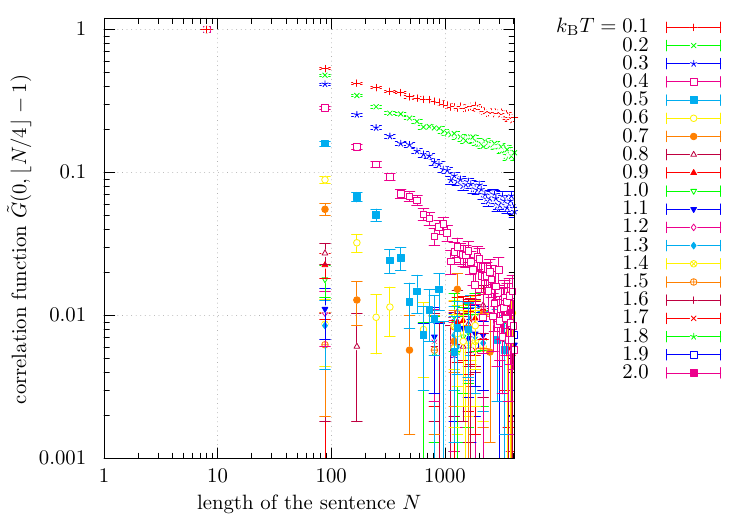}
	\caption{Correlation functions, Eq.~\eqref{supp_eq_correlation_function_with_disconnected_diagram_Potts_001_001}, (upper) with $i=2048$ and $j=2048 + \Delta i$ and (lower) with $i = 0$ and $j = \lfloor N / 4 \rfloor - 1$ for $X \to YZ$. We set $K = 10$, $J = 1.0$, $q = 10^{-2.0}$, $t = 0$, $s = 0.9$, and $r_- = r_+ = 0.25$. The temperature $k_\mathrm{B} T$ was varied from $0.1$ to $2.0$.}
	\label{supp_fig_correlation_function_K=10_q=10^(-2.0)_X_YZ_001_001}
\end{figure}

In Fig.~\ref{supp_fig_correlation_function_K=10_q=10^(-2.0)_X_YZ_001_001}, we show the correlation functions, Eq.~\eqref{supp_eq_correlation_function_with_disconnected_diagram_Potts_001_001}.
The behavior of the correlation function with $i=2048$ and $j=2048 + \Delta i$ shown in Fig.~\ref{supp_fig_correlation_function_K=10_q=10^(-2.0)_X_YZ_001_001}(upper) looks similar to that of the first-order phase transition, since it changes from concave functions to convex functions as the temperature is lowered without becoming a straight line in the log-log plot.
However, the correlation function with $i = 0$ and $j = \lfloor N / 4 \rfloor - 1$ shown in Fig.~\ref{supp_fig_correlation_function_K=10_q=10^(-2.0)_X_YZ_001_001}(lower) shows straight lines in the log-log plot for a wide range of $k_\mathrm{B} T$.

\begin{figure}[t]
	\centering
	\includegraphics[scale=0.60]{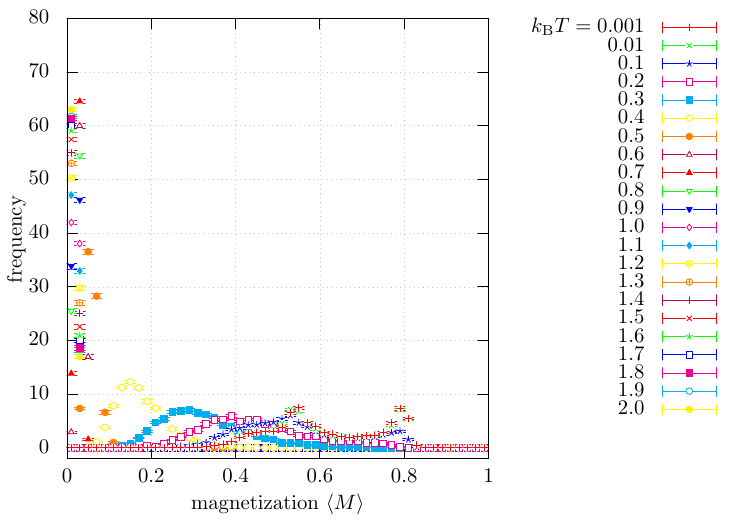}
	\caption{Histogram of the magnetization, Eq.~\eqref{supp_eq_def_magnetization_001_001}, for $X \to YZ$. We set $K = 10$, $J = 1.0$, $q = 10^{-2.0}$, $t = 0$, $s = 0.9$, and $r_- = r_+ = 0.25$. The temperature $k_\mathrm{B} T$ was varied from $0.1$ to $2.0$.}
	\label{supp_fig_histogram_mag_K=10_q=10^(-2.0)_X_YZ_001_001}
\end{figure}

In Fig.~\ref{supp_fig_histogram_mag_K=10_q=10^(-2.0)_X_YZ_001_001}, the histogram of the magnetization, Eq.~\eqref{supp_eq_histogram_magnetization_001_001}, is shown.
The novel point of Fig.~\ref{supp_fig_histogram_mag_K=10_q=10^(-2.0)_X_YZ_001_001} is that the histogram has two peaks toward $k_\mathrm{B} T = 0$, and this also suggests that there is a phase transition similar to a first-order phase transition at $k_\mathrm{B} T = 0$.

\subsubsection{Correlation functions, mutual information, and histogram of magnetization for $q = 10^{-0.5}$}

\begin{figure}[t]
	\centering
	\includegraphics[scale=0.60]{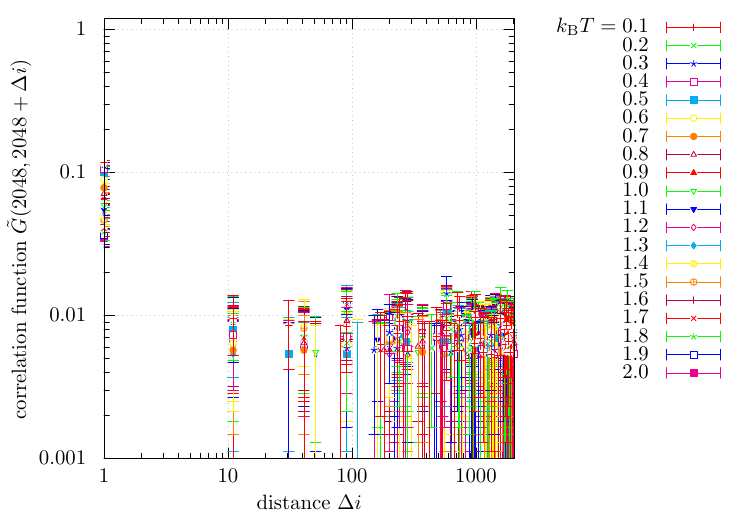}
	\includegraphics[scale=0.60]{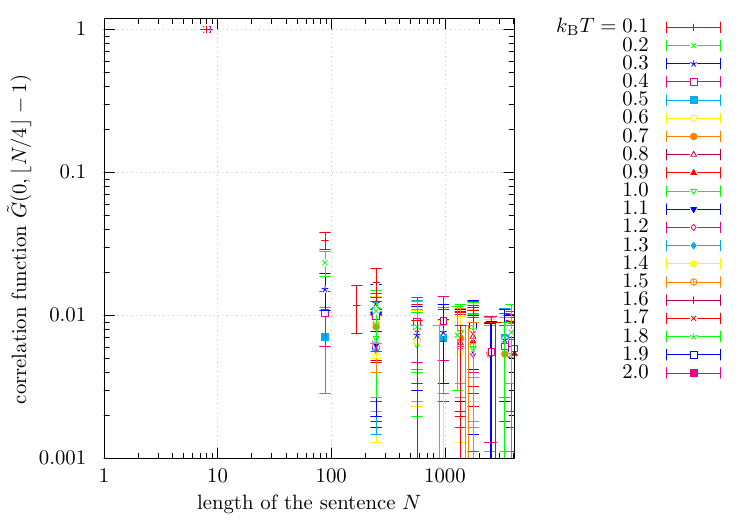}
	\caption{Correlation functions, Eq.~\eqref{supp_eq_correlation_function_with_disconnected_diagram_Potts_001_001}, (upper) with $i=2048$ and $j=2048 + \Delta i$ and (lower) with $i = 0$ and $j = \lfloor N / 4 \rfloor - 1$ for $X \to YZ$. We set $K = 10$, $J = 1.0$, $q = 10^{-0.5}$, $t = 0$, $s = 0.9$, and $r_- = r_+ = 0.25$. The temperature $k_\mathrm{B} T$ was varied from $0.1$ to $2.0$.}
	\label{supp_fig_correlation_function_K=10_q=10^(-0.5)_X_YZ_001_001}
\end{figure}

In Fig.~\ref{supp_fig_correlation_function_K=10_q=10^(-0.5)_X_YZ_001_001}, the correlation function, Eq.~\eqref{supp_eq_correlation_function_with_disconnected_diagram_Potts_001_001}, is shown.

\begin{figure}[t]
	\centering
	\includegraphics[scale=0.60]{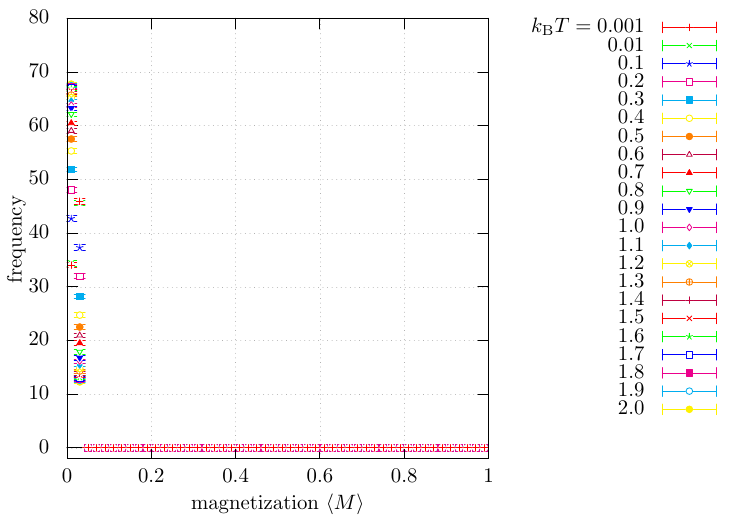}
	\caption{Histogram of the magnetization, Eq.~\eqref{supp_eq_def_magnetization_001_001}, for $X \to YZ$. We set $K = 10$, $J = 1.0$, $q = 10^{-0.5}$, $t = 0$, $s = 0.9$, and $r_- = r_+ = 0.25$. The temperature $k_\mathrm{B} T$ was varied from $0.1$ to $2.0$.}
	\label{supp_fig_histogram_mag_K=10_q=10^(-0.5)_X_YZ_001_001}
\end{figure}

In Fig.~\ref{supp_fig_histogram_mag_K=10_q=10^(-0.5)_X_YZ_001_001}, we show the histogram of the magnetization, Eq.~\eqref{supp_eq_histogram_magnetization_001_001}.
The above figures consistently imply that the state is always trivial without any singularities.

\subsubsection{System-size dependence of the magnetization, the susceptibilities, and the Binder parameter}

\begin{figure}[t]
	\centering
	\includegraphics[scale=0.60]{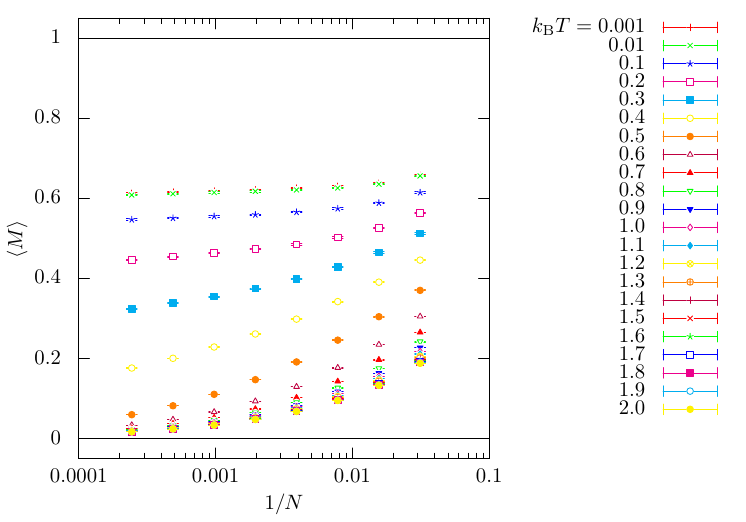}
	\caption{System-size dependence of the magnetization, Eq.~\eqref{supp_eq_def_magnetization_001_001}, at $q = 10^{-2.0}$. We consider $X \to YZ$ and set $K = 10$, $J = 1.0$, $t = 0$, $s = 0.9$, and $r_- = r_+ = 0.25$. We varied $N = 32, 64, 128, 256, 512, 1024, 2048, 4096$.}
	\label{supp_fig_system-size-dependence_magnetization_002_001}
\end{figure}

\begin{figure}[t]
	\centering
	\includegraphics[scale=0.60]{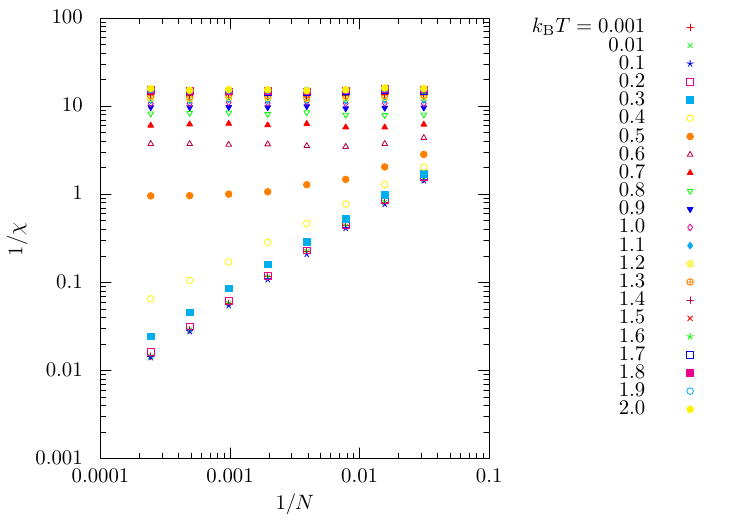}
	\includegraphics[scale=0.60]{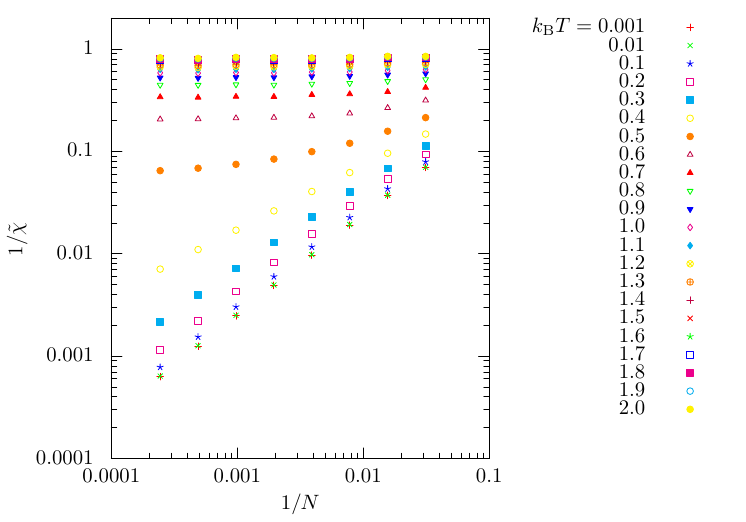}
	\caption{System-size dependence of the susceptibilities, (upper) Eq.~\eqref{supp_eq_def_specific_heat_001_001} and (lower) Eq.~\eqref{supp_eq_def_specific_heat_001_002} at $q = 10^{-2.0}$. We consider $X \to YZ$ and set $K = 10$, $J = 1.0$, $t = 0$, $s = 0.9$, and $r_- = r_+ = 0.25$. We varied $N = 32, 64, 128, 256, 512, 1024, 2048, 4096$.}
	\label{supp_fig_system-size-dependence_susceptibility_002_001}
\end{figure}

\begin{figure}[t]
	\centering
	\includegraphics[scale=0.60]{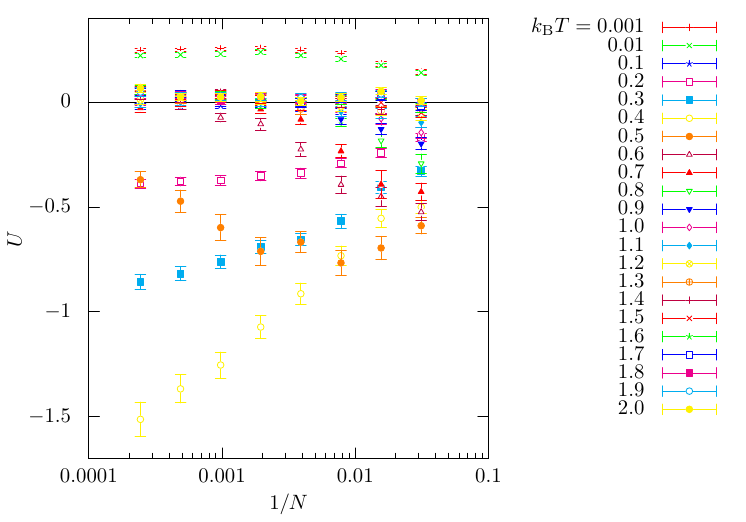}
	\caption{System-size dependence of the Binder parameter, Eq.~\eqref{supp_eq_def_Binder_parameter_001_001}, at $q = 10^{-2.0}$. We consider $X \to YZ$ and set $K = 10$, $J = 1.0$, $t = 0$, $s = 0.9$, and $r_- = r_+ = 0.25$. We varied $N = 32, 64, 128, 256, 512, 1024, 2048, 4096$.}
	\label{supp_fig_system-size-dependence_Binder_002_001}
\end{figure}

In Figs.~\ref{supp_fig_system-size-dependence_magnetization_002_001}, \ref{supp_fig_system-size-dependence_susceptibility_002_001}, and \ref{supp_fig_system-size-dependence_Binder_002_001}, we show the system-size dependence of the magnetization, Eq.~\eqref{supp_eq_def_magnetization_001_001}, the two susceptibilities, Eqs.~\eqref{supp_eq_def_specific_heat_001_001} and \eqref{supp_eq_def_specific_heat_001_002}, and the Binder parameter, Eq.~\eqref{supp_eq_def_Binder_parameter_001_001}, at $q = 10^{-2.0}$, respectively.
Similarly, the critical temperature was determined by assessing whether the line is linear or not.
This estimate is consistent with the temperature at which the Binder parameter takes a non-zero value in the thermodynamic limit, as shown in Fig.~\ref{supp_fig_system-size-dependence_Binder_002_001}.
Figure~\ref{supp_fig_system-size-dependence_Binder_002_001} suggests that phase transitions take place twice, but we do not discuss the details in this paper.

\subsubsection{Phase diagram, finite-size scaling, and $q$-dependence of critical exponents}

\begin{figure}[t]
	\centering
	\includegraphics[scale=0.60]{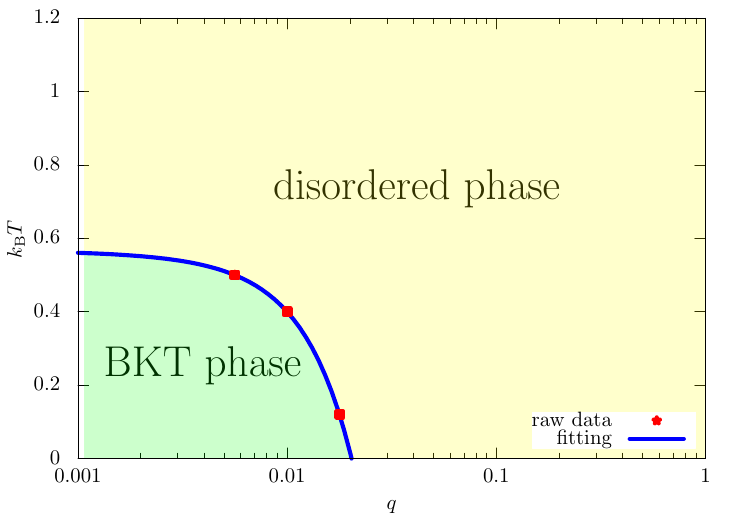}
	\caption{Phase diagram of the context-sensitive random language model, where the horizontal and vertical axes are the growth rate of a sentence $q$ and temperature $k_\mathrm{B} T$, respectively. We consider $X \to YZ$ and set $K = 10$, $J = 1.0$, $t = 0$, $s = 0.9$, and $r_- = r_+ = 0.25$.}
	\label{supp_fig_phase_diagram_003_001}
\end{figure}

In Fig.~\ref{supp_fig_phase_diagram_003_001}, we plot the phase diagram of our language model with $K = 10$ and $X \to YZ$.
We determined the boundary between the two phases by fitting a quadratic function via least squares regression.
Compared to the case of $K = 2$, the regime of the BKT phase is reduced.

\begin{figure}[t]
	\centering
	\includegraphics[scale=0.60]{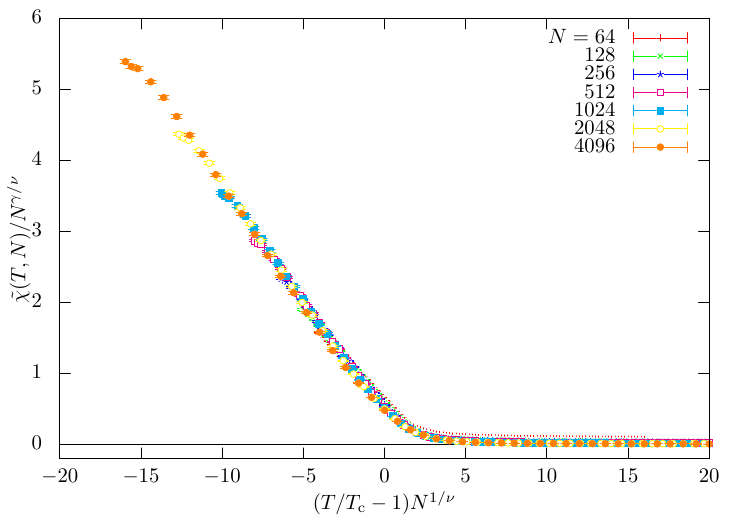}
	\caption{Finite-size scaling of $\tilde{\chi}$ at $q = 10^{-2.0}$. We set $T_\mathrm{c} = 0.40$, $\nu = 3.0$, and $\gamma = 2.05$, where the values of $\nu$ and $\gamma$ are determined such that the scaling assumption is satisfied. We consider $X \to YZ$ and set $K = 10$, $J = 1.0$, $t = 0$, $s = 0.9$, and $r_- = r_+ = 0.25$. We varied $N = 64, 128, 256, 512, 1024, 2048, 4096$.}
	\label{supp_fig_finite-size-scaling_003_001}
\end{figure}

In Fig.~\ref{supp_fig_finite-size-scaling_003_001}, we carry out finite-size scaling for $\tilde{\chi}$ at $q = 10^{-2.0}$.
We set $T_\mathrm{c} = 0.40$, $\nu = 3.0$, and $\gamma = 2.05$, where $\nu$ and $\gamma$ are determined such that the scaling assumption holds.

\begin{figure}[t]
	\centering
	\includegraphics[scale=0.60]{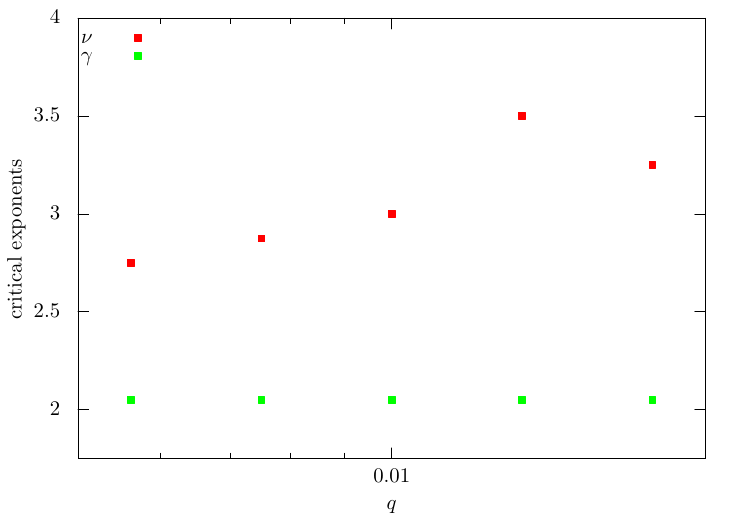}
	\caption{$q$-dependence of the critical exponents $\nu$ and $\gamma$. We consider $X \to YZ$ and set $K = 10$, $J = 1.0$, $t = 0$, $s = 0.9$, and $r_- = r_+ = 0.25$.}
	\label{supp_fig_q_dependence_critical_exponents_003_001}
\end{figure}

In Fig.~\ref{supp_fig_q_dependence_critical_exponents_003_001}, we plot the $q$-dependence of the critical exponents $\nu$ and $\gamma$.
We have observed the robust critical behavior, concluding that the BKT transition occurs.

\subsubsection{Configurations and relative frequencies for $q = 10^{-2.0}$}

\begin{figure}[t]
	\centering
	\includegraphics[scale=0.60]{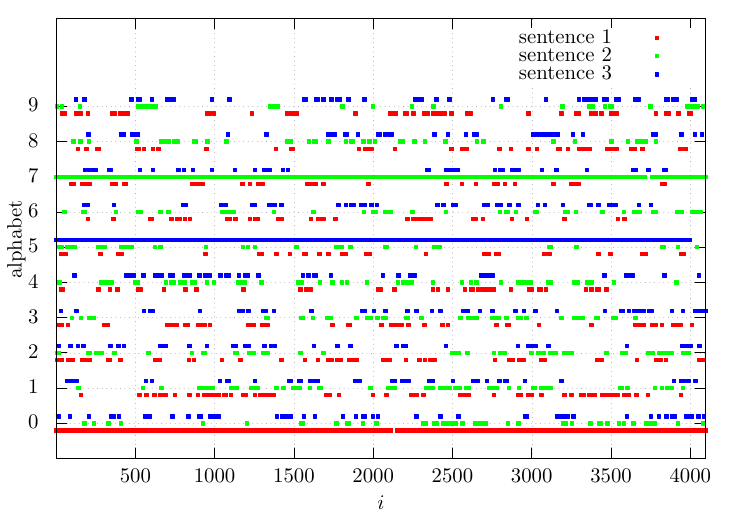}
	\includegraphics[scale=0.60]{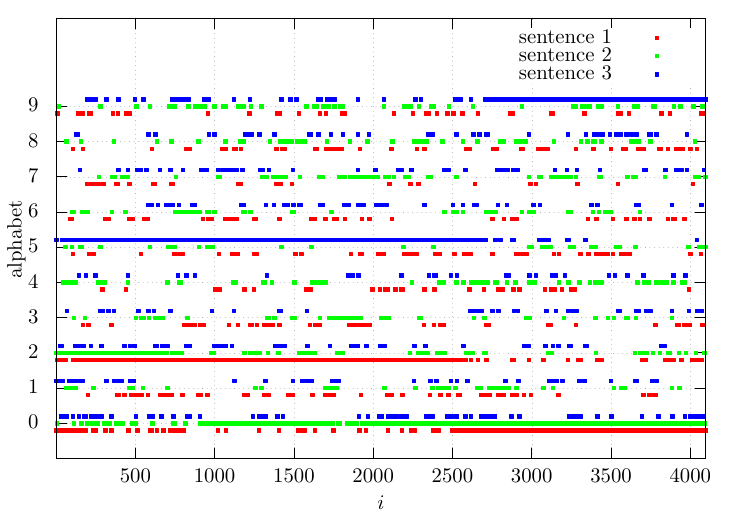}
	\caption{Typical configurations of symbols for $X \to YZ$. (upper) We considered states whose magnetization lies between $0.5400$ and $0.5600$ at $k_\mathrm{B} T = 0.0010$, and (lower) states whose magnetization lies between $0.7800$ and $0.8000$ at $k_\mathrm{B} T = 0.0010$. We set $K = 10$, $J = 1.0$, $q = 10^{-2.0}$, $t = 0$, $s = 0.9$, and $r_- = r_+ = 0.25$.}
	\label{supp_fig_configurations_K=10_q=10^(-2.0)_X_YZ_001_001}
\end{figure}

As shown in Fig.~\ref{supp_fig_histogram_mag_K=10_q=10^(-2.0)_X_YZ_001_001}, the histogram of magnetization has peaks at $\langle M \rangle = 0.550, 0.790$.
Then, in Fig.~\ref{supp_fig_configurations_K=10_q=10^(-2.0)_X_YZ_001_001}, we show typical configurations whose magnetization values are $0.550$ and $0.790$.
In both cases, symbols are scattered, but their distributions have different features.
Around $\langle M \rangle = 0.790$, one character dominates, whereas near $\langle M \rangle = 0.550$, two major characters are comparatively dominant.

\begin{figure}[t]
	\centering
	\includegraphics[scale=0.60]{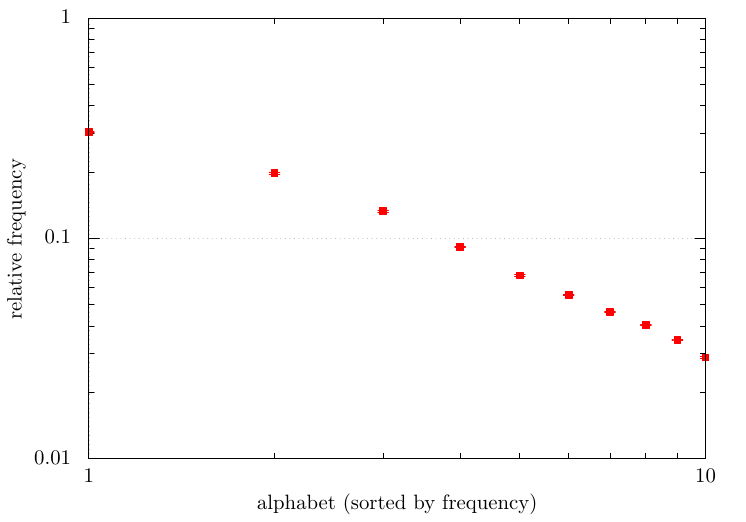}
	\includegraphics[scale=0.60]{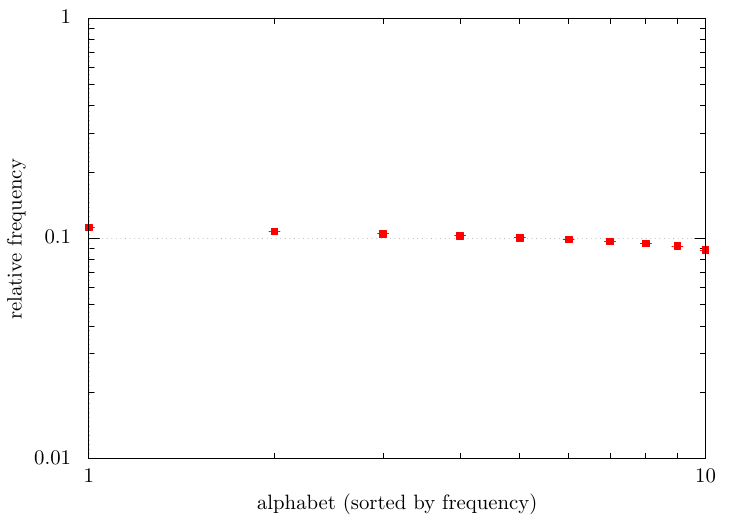}
	\caption{Relative frequencies of letters in the alphabet for $X \to YZ$. (upper) We considered states whose magnetization lies between $0.2800$ and $0.3000$ at $k_\mathrm{B} T = 0.3000$, and (lower) states whose magnetization lies between $0.0200$ and $0.0400$ at $k_\mathrm{B} T = 0.1000$. We set $K = 10$, $J = 1.0$, $q = 10^{-2.0}$, $t = 0$, $s = 0.9$, and $r_- = r_+ = 0.25$. Symbols on the horizontal axis are sorted by frequency and incremented by unity. The scales of the vertical and horizontal axes are logarithmic.}
	\label{supp_fig_relative_frequencies_K=10_q=10^(-2.0)_X_YZ_001_001}
\end{figure}

\begin{figure}[t]
	\centering
	\includegraphics[scale=0.60]{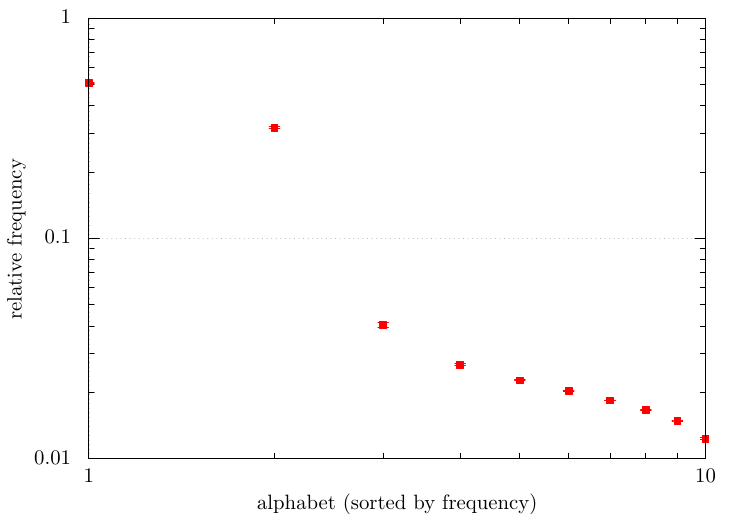}
	\includegraphics[scale=0.60]{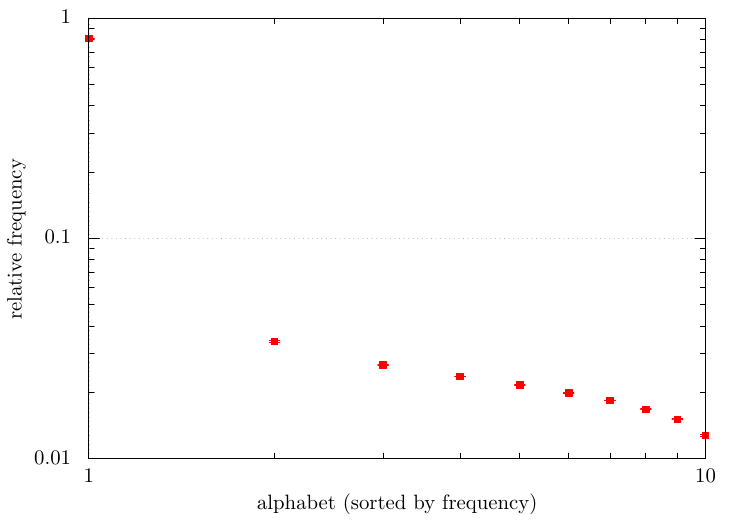}
	\caption{Relative frequencies of letters in the alphabet for $X \to YZ$. (upper) We considered states whose magnetization lies between $0.5400$ and $0.5600$ at $k_\mathrm{B} T = 0.0010$, and (lower) states whose magnetization lies between $0.7800$ and $0.8000$ at $k_\mathrm{B} T = 0.0010$. We set $K = 10$, $J = 1.0$, $q = 10^{-2.0}$, $t = 0$, $s = 0.9$, and $r_- = r_+ = 0.25$. Symbols on the horizontal axis are sorted by frequency and incremented by unity. The scales of the vertical and horizontal axes are logarithmic.}
	\label{supp_fig_relative_frequencies_K=10_q=10^(-2.0)_X_YZ_001_002}
\end{figure}

In Fig.~\ref{supp_fig_relative_frequencies_K=10_q=10^(-2.0)_X_YZ_001_001}, we plot the relative frequencies of letters in the alphabet.
Symbols on the horizontal axis are sorted by their frequencies.
In natural languages, relative frequencies of letters in the alphabet often show Zipf's law; that is, they form a straight line in the log-log plot.
The relative frequencies at $k_\mathrm{B} T = 0.3000$ seem to be a straight line in Fig.~\ref{supp_fig_relative_frequencies_K=10_q=10^(-2.0)_X_YZ_001_001}(upper).
However, the histogram of letters in the alphabet is almost uniformly random at $k_\mathrm{B} T = 0.1000$, as shown in Fig.~\ref{supp_fig_relative_frequencies_K=10_q=10^(-2.0)_X_YZ_001_001}(lower).
In Fig.~\ref{supp_fig_relative_frequencies_K=10_q=10^(-2.0)_X_YZ_001_002}, we also plot the relative frequencies of letters in the alphabet at $k_\mathrm{B} T = 0.0010$.
For characters whose relative frequencies are small, we can see Zipf's law.

\subsection{Case of $K=10$ and $X \to XX$}

Lastly, we consider the case of $K=10$ and $X \to XX$.

\subsubsection{Magnetization, susceptibilities, and Binder parameter}

\begin{figure}[t]
	\centering
	\includegraphics[scale=0.60]{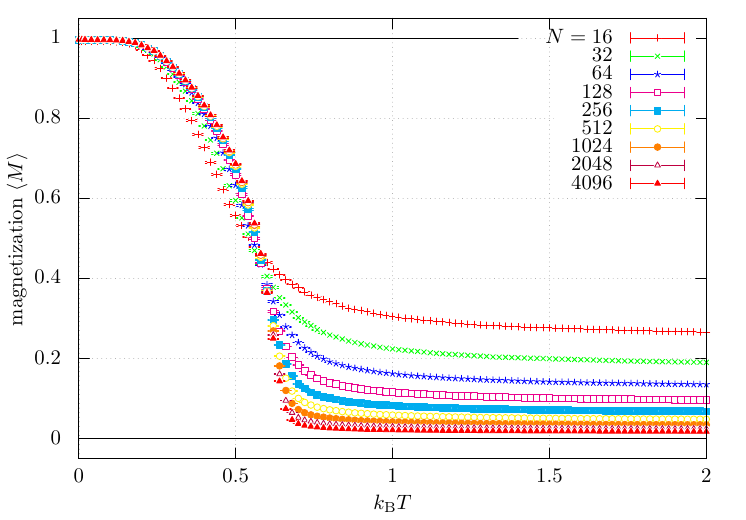}
	\includegraphics[scale=0.60]{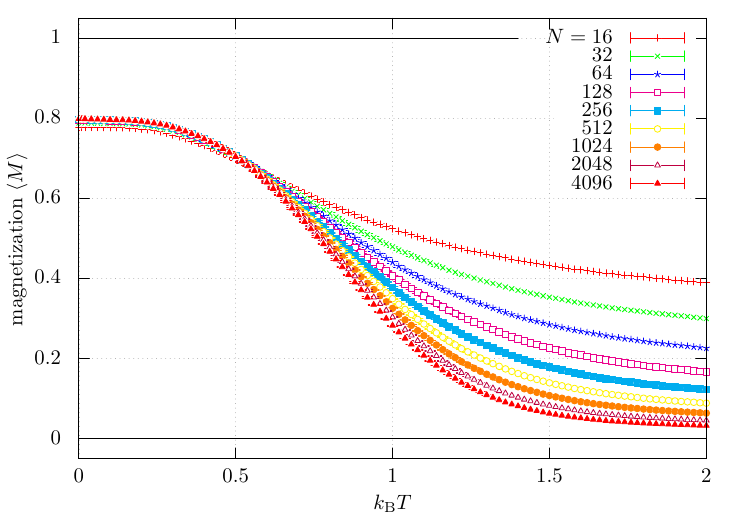}
	\caption{Temperature dependence of the magnetization, Eq.~\eqref{supp_eq_def_magnetization_001_001}, for $X \to XX$. We set (upper) $q = 10^{-2.0}$ and (lower) $q = 10^{-0.5}$. We also set $K = 10$, $J = 1.0$, $q = 10^{-0.5}$, $t = 0$, $s = 0.9$, and $r_- = r_+ = 0.25$. The length of the generated sentence by the language model, $N$, was varied from $16$ to $4096$.}
	\label{supp_fig_magnetization_K=10_X_XX_001_001}
\end{figure}

In Fig.~\ref{supp_fig_magnetization_K=10_X_XX_001_001}, we plot the magnetization, Eq.~\eqref{supp_eq_def_magnetization_001_001}.
In both cases, the magnetization, Eq.~\eqref{supp_eq_def_magnetization_001_001}, approaches a nontrivial value when $k_\mathrm{B} T$ is decreased.
The magnetization seems to jump near $k_\mathrm{B} T \sim 0.7$ for $q = 10^{-2.0}$, whereas it gradually increases for $q = 10^{-0.5}$.

\begin{figure}[t]
	\centering
	\includegraphics[scale=0.60]{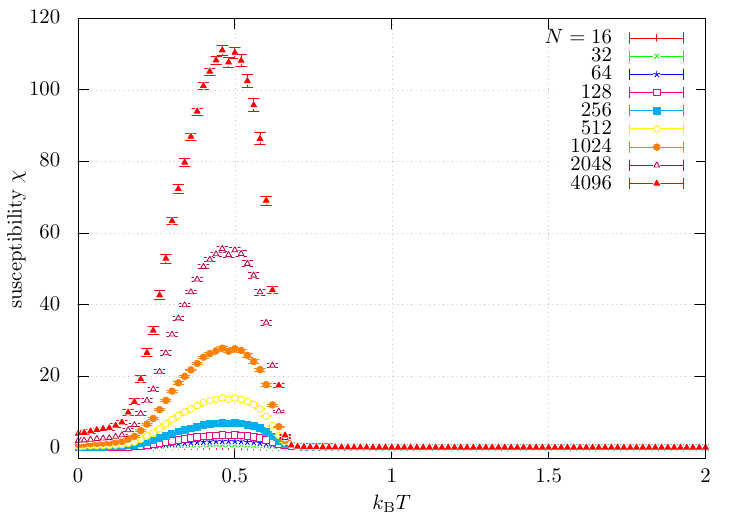}
	\includegraphics[scale=0.60]{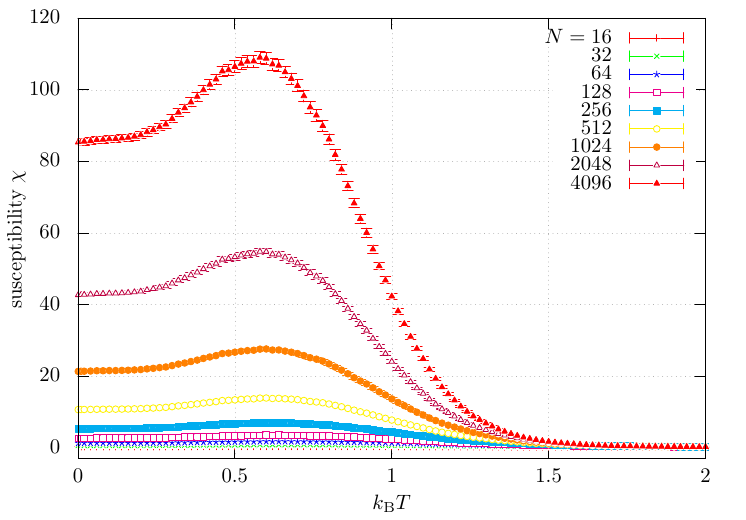}
	\caption{Temperature dependence of the susceptibility, Eq.~\eqref{supp_eq_def_specific_heat_001_001}, for $X \to XX$. We set (upper) $q = 10^{-2.0}$ and (lower) $q = 10^{-0.5}$. We also set $K = 10$, $J = 1.0$, $q = 10^{-0.5}$, $t = 0$, $s = 0.9$, and $r_- = r_+ = 0.25$. The length of the generated sentence by the language model, $N$, was varied from $16$ to $4096$.}
	\label{supp_fig_susceptibility_K=10_X_XX_001_001}
\end{figure}

In Fig.~\ref{supp_fig_susceptibility_K=10_X_XX_001_001}, the susceptibility, Eq.~\eqref{supp_eq_def_specific_heat_001_001} is plotted.
In Fig.~\ref{supp_fig_susceptibility_K=10_X_XX_001_001}, the susceptibility has a peak at a specific temperature.
This fact also supports the existence of a phase transition.

\begin{figure}[t]
	\centering
	\includegraphics[scale=0.60]{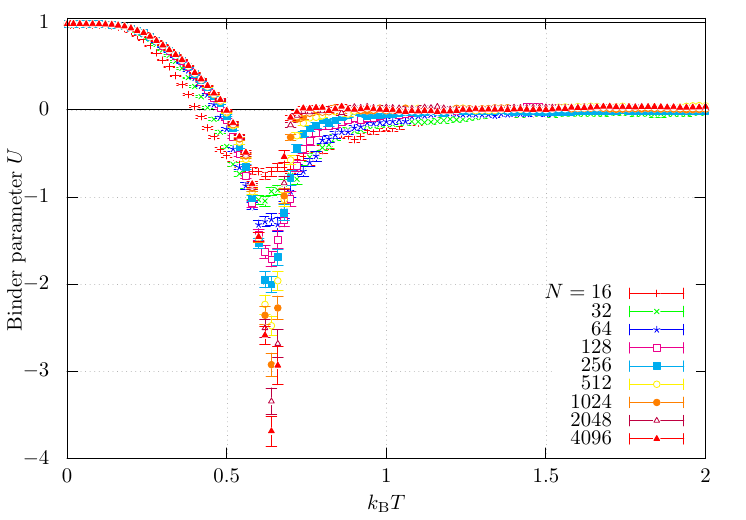}
	\includegraphics[scale=0.60]{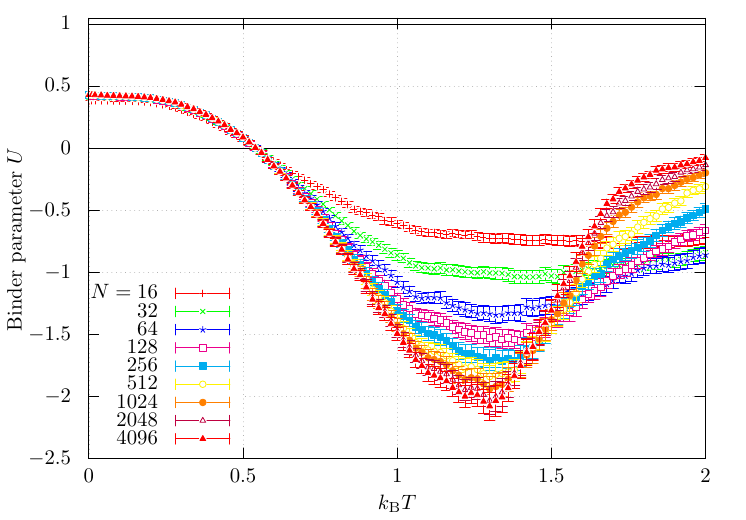}
	\caption{Temperature dependence of the Binder parameter, Eq.~\eqref{supp_eq_def_Binder_parameter_001_001}, for $X \to XX$. We set (upper) $q = 10^{-2.0}$ and (lower) $q = 10^{-0.5}$. We also set $K = 10$, $J = 1.0$, $q = 10^{-0.5}$, $t = 0$, $s = 0.9$, and $r_- = r_+ = 0.25$. The length of the generated sentence by the language model, $N$, was varied from $16$ to $4096$.}
	\label{supp_fig_Binder_parameter_K=10_X_XX_001_001}
\end{figure}

In Fig.~\ref{supp_fig_Binder_parameter_K=10_X_XX_001_001}, the Binder parameter, Eq.~\eqref{supp_eq_def_Binder_parameter_001_001}, is shown.
For $q = 10^{-2.0}$, the Binder parameter exhibits a sharp dip before reaching 1 as $k_\mathrm{B} T$ decreases.
This behavior is characteristic of a first-order phase transition.
Similarly, for $q = 10^{-0.5}$, the Binder parameter displays a similar trend, albeit without the width of the dip.
In both cases, the existence of a first-order phase transition appears evident.
However, the exact cause of the wide dip in the Binder parameter is not clear.

\subsubsection{Correlation functions, mutual information, and histogram of magnetization for $q = 10^{-2.0}$}

\begin{figure}[t]
	\centering
	\includegraphics[scale=0.60]{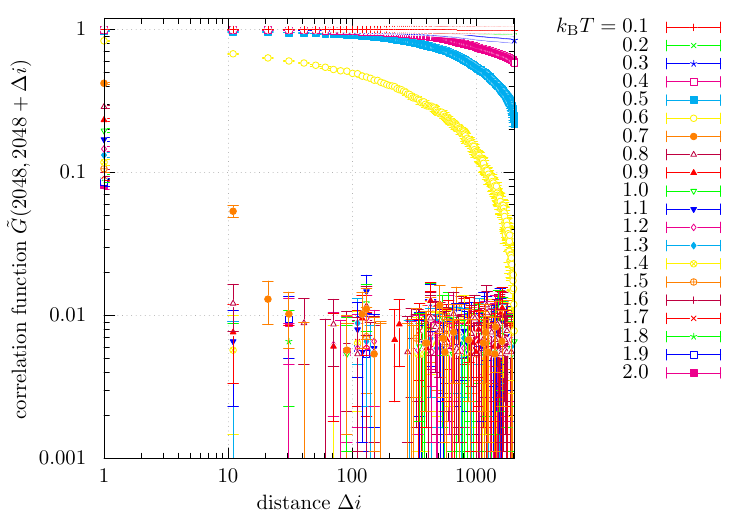}
	\includegraphics[scale=0.60]{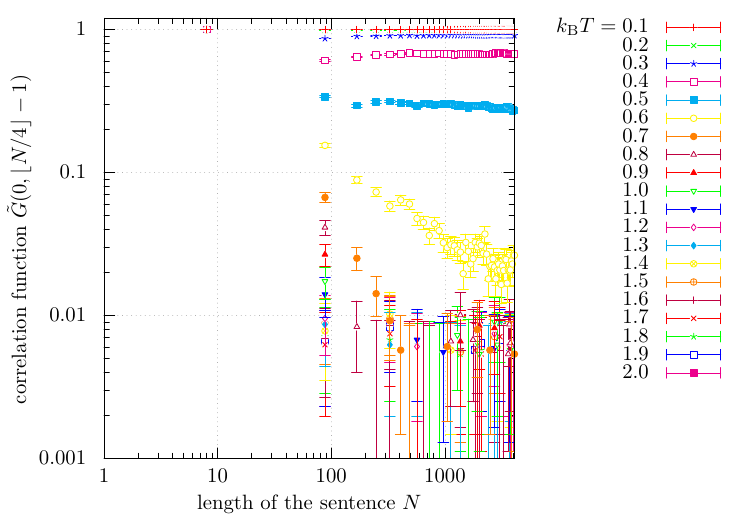}
	\caption{Correlation functions, Eq.~\eqref{supp_eq_correlation_function_with_disconnected_diagram_Potts_001_001}, (upper) with $i=2048$ and $j=2048 + \Delta i$ and (lower) with $i = 0$ and $j = \lfloor N / 4 \rfloor - 1$ for $X \to XX$. We set $K = 10$, $J = 1.0$, $q = 10^{-2.0}$, $t = 0$, $s = 0.9$, and $r_- = r_+ = 0.25$. The temperature $k_\mathrm{B} T$ was varied from $0.1$ to $2.0$.}
	\label{supp_fig_correlation_function_K=10_q=10^(-2.0)_X_XX_001_001}
\end{figure}

In Fig.~\ref{supp_fig_correlation_function_K=10_q=10^(-2.0)_X_XX_001_001}, the correlation function, Eq.~\eqref{supp_eq_correlation_function_with_disconnected_diagram_Potts_001_001}, is shown.
In Fig.~\ref{supp_fig_correlation_function_K=10_q=10^(-2.0)_X_XX_001_001}(upper), the correlation functions, Eq.~\eqref{supp_eq_correlation_function_with_disconnected_diagram_Potts_001_001}, transition from concave to convex functions as $k_\mathrm{B} T$ decreases.
This observation supports the existence of a first-order phase transition.
However, Fig.~\ref{supp_fig_correlation_function_K=10_q=10^(-2.0)_X_XX_001_001}(lower) displays straight lines over a wide range of $k_\mathrm{B} T$.

\begin{figure}[t]
	\centering
	\includegraphics[scale=0.60]{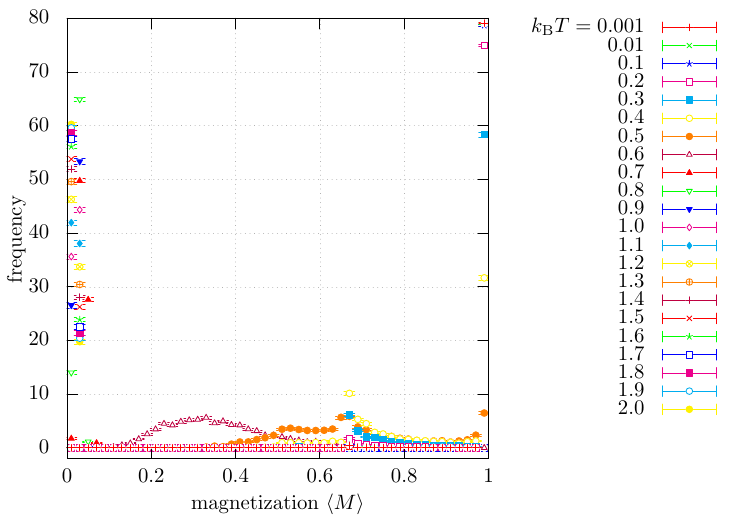}
	\caption{Histogram of the magnetization, Eq.~\eqref{supp_eq_def_magnetization_001_001}, for $X \to XX$. We set $K = 10$, $J = 1.0$, $q = 10^{-2.0}$, $t = 0$, $s = 0.9$, and $r_- = r_+ = 0.25$. The temperature $k_\mathrm{B} T$ was varied from $0.1$ to $2.0$.}
	\label{supp_fig_histogram_mag_K=10_q=10^(-2.0)_X_XX_001_001}
\end{figure}

In Fig.~\ref{supp_fig_histogram_mag_K=10_q=10^(-2.0)_X_XX_001_001}, we plot the histogram of the magnetization, Eq.~\eqref{supp_eq_histogram_magnetization_001_001}.
At $k_\mathrm{B} T \sim 0.4$, there are two peaks.
This fact strongly supports the existence of a phase transition similar to that of the first-order phase transition.

\subsubsection{Correlation functions, mutual information, and histogram of magnetization for $q = 10^{-0.5}$}

\begin{figure}[t]
	\centering
	\includegraphics[scale=0.60]{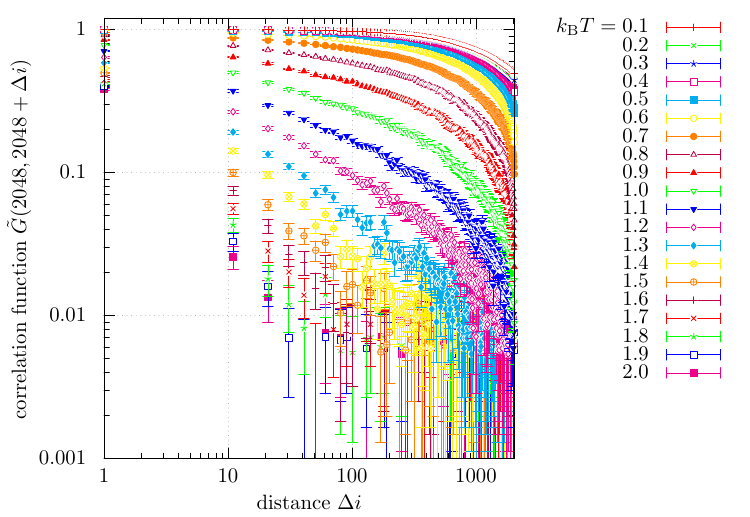}
	\includegraphics[scale=0.60]{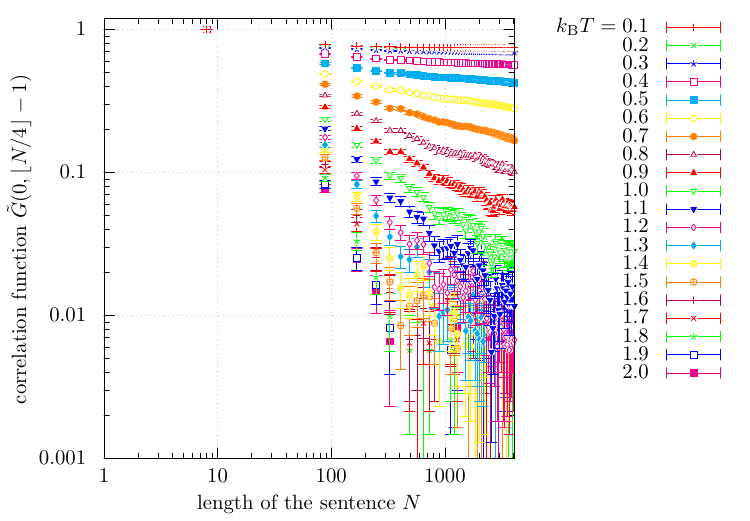}
	\caption{Correlation functions, Eq.~\eqref{supp_eq_correlation_function_with_disconnected_diagram_Potts_001_001}, (upper) with $i=2048$ and $j=2048 + \Delta i$ and (lower) with $i = 0$ and $j = \lfloor N / 4 \rfloor - 1$ for $X \to XX$. We set $K = 10$, $J = 1.0$, $q = 10^{-0.5}$, $t = 0$, $s = 0.9$, and $r_- = r_+ = 0.25$. The temperature $k_\mathrm{B} T$ was varied from $0.1$ to $2.0$.}
	\label{supp_fig_correlation_function_K=10_q=10^(-0.5)_X_XX_001_001}
\end{figure}

In Fig.~\ref{supp_fig_correlation_function_K=10_q=10^(-0.5)_X_XX_001_001}, the correlation functions, Eq.~\eqref{supp_eq_correlation_function_with_disconnected_diagram_Potts_001_001}, are shown.
For $q = 10^{-0.5}$, the correlation function varies similar to that of the second-order phase transition.

\begin{figure}[t]
	\centering
	\includegraphics[scale=0.60]{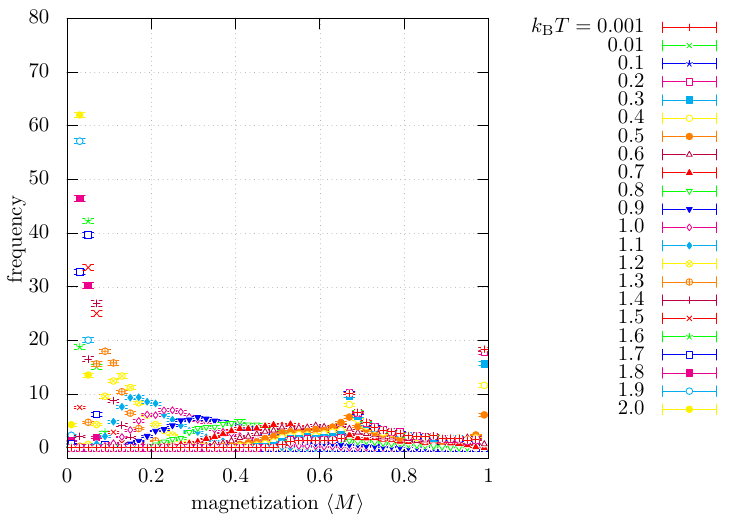}
	\caption{Histogram of the magnetization, Eq.~\eqref{supp_eq_def_magnetization_001_001}, for $X \to XX$. We set $K = 10$, $J = 1.0$, $q = 10^{-0.5}$, $t = 0$, $s = 0.9$, and $r_- = r_+ = 0.25$. The temperature $k_\mathrm{B} T$ was varied from $0.1$ to $2.0$.}
	\label{supp_fig_histogram_mag_K=10_q=10^(-0.5)_X_XX_001_001}
\end{figure}

In Fig.~\ref{supp_fig_histogram_mag_K=10_q=10^(-0.5)_X_XX_001_001}, the histogram of the magnetization, Eq.~\eqref{supp_eq_histogram_magnetization_001_001}, is shown.
Figure~\ref{supp_fig_histogram_mag_K=10_q=10^(-0.5)_X_XX_001_001} shows two peaks toward $k_\mathrm{B} T = 0$.
This fact implies that a phase transition akin to that of the first-order phase transition occurs at $k_\mathrm{B} T = 0$.

\subsubsection{System-size dependence of the magnetization, the susceptibilities, and the Binder parameter}

\begin{figure}[t]
	\centering
	\includegraphics[scale=0.60]{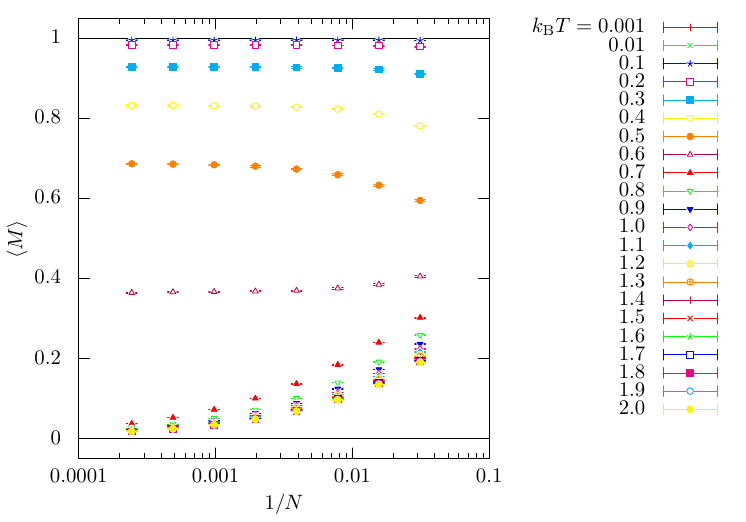}
	\caption{System-size dependence of the magnetization, Eq.~\eqref{supp_eq_def_magnetization_001_001}, at $q = 10^{-2.0}$. We consider $X \to XX$ and set $K = 10$, $J = 1.0$, $t = 0$, $s = 0.9$, and $r_- = r_+ = 0.25$. We varied $N = 32, 64, 128, 256, 512, 1024, 2048, 4096$.}
	\label{supp_fig_system-size-dependence_magnetization_002_002}
\end{figure}

\begin{figure}[t]
	\centering
	\includegraphics[scale=0.60]{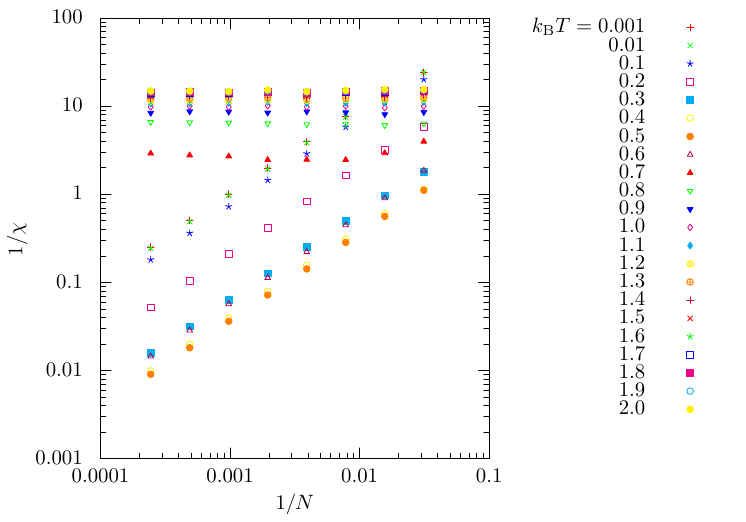}
	\includegraphics[scale=0.60]{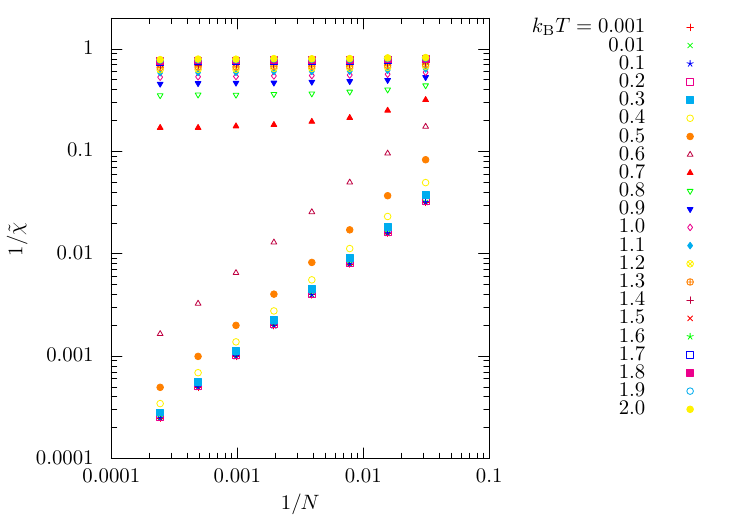}
	\caption{System-size dependence of the susceptibilities, (upper) Eq.~\eqref{supp_eq_def_specific_heat_001_001} and (lower) Eq.~\eqref{supp_eq_def_specific_heat_001_002} at $q = 10^{-2.0}$. We consider $X \to XX$ and set $K = 10$, $J = 1.0$, $t = 0$, $s = 0.9$, and $r_- = r_+ = 0.25$. We varied $N = 32, 64, 128, 256, 512, 1024, 2048, 4096$.}
	\label{supp_fig_system-size-dependence_susceptibility_002_002}
\end{figure}

\begin{figure}[t]
	\centering
	\includegraphics[scale=0.60]{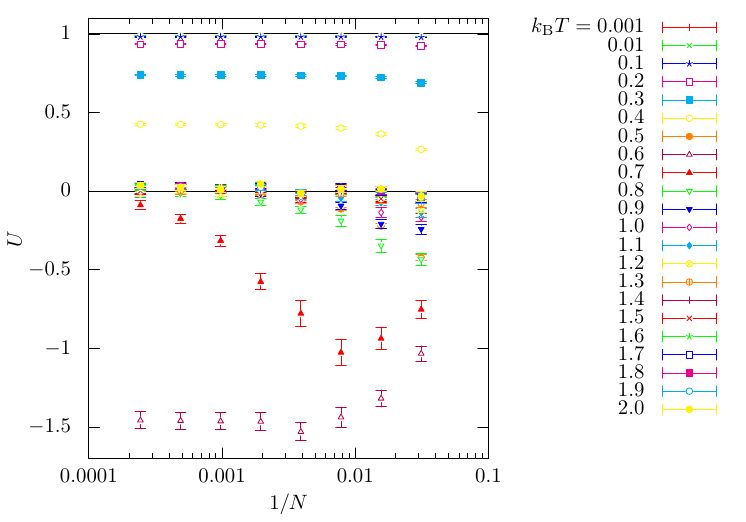}
	\caption{System-size dependence of the Binder parameter, Eq.~\eqref{supp_eq_def_Binder_parameter_001_001}, at $q = 10^{-2.0}$. We consider $X \to XX$ and set $K = 10$, $J = 1.0$, $t = 0$, $s = 0.9$, and $r_- = r_+ = 0.25$. We varied $N = 32, 64, 128, 256, 512, 1024, 2048, 4096$.}
	\label{supp_fig_system-size-dependence_Binder_002_002}
\end{figure}

In Figs.~\ref{supp_fig_system-size-dependence_magnetization_002_002}, \ref{supp_fig_system-size-dependence_susceptibility_002_002}, and \ref{supp_fig_system-size-dependence_Binder_002_002}, we show the system-size dependence of the magnetization, Eq.~\eqref{supp_eq_def_magnetization_001_001}, the two susceptibilities, Eqs.~\eqref{supp_eq_def_specific_heat_001_001} and \eqref{supp_eq_def_specific_heat_001_002}, and the Binder parameter, Eq.~\eqref{supp_eq_def_Binder_parameter_001_001}, at $q = 10^{-2.0}$, respectively.
The critical temperature was again determined by assessing whether the line is linear or not.
This estimate is consistent with the temperature at which the Binder parameter takes a non-zero value in the thermodynamic limit, as shown in Fig.~\ref{supp_fig_system-size-dependence_Binder_002_002}.
Figure~\ref{supp_fig_system-size-dependence_Binder_002_002} also suggests that phase transitions occur twice, but we do not discuss the details in this paper.

\subsubsection{Phase diagram, finite-size scaling, and $q$-dependence of critical exponents}

\begin{figure}[t]
	\centering
	\includegraphics[scale=0.60]{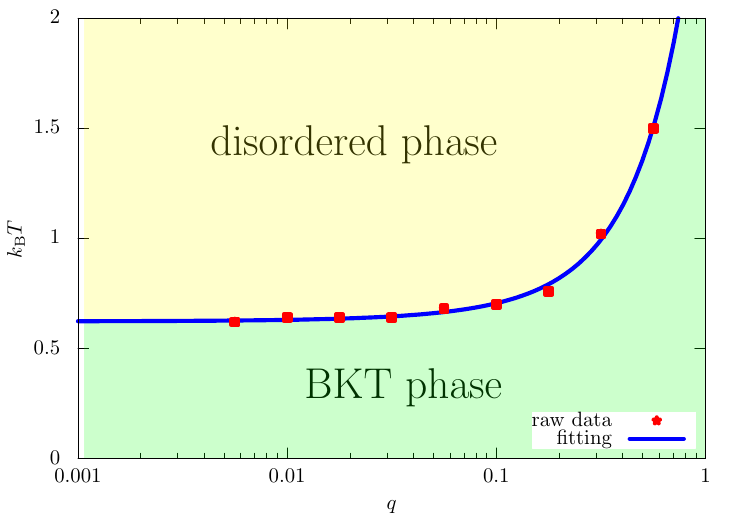}
	\caption{Phase diagram of the context-sensitive random language model, where the horizontal and vertical axes are the growth rate of a sentence $q$ and temperature $k_\mathrm{B} T$, respectively. We consider $X \to XX$ and set $K = 10$, $J = 1.0$, $t = 0$, $s = 0.9$, and $r_- = r_+ = 0.25$.}
	\label{supp_fig_phase_diagram_004_001}
\end{figure}

In Fig.~\ref{supp_fig_phase_diagram_004_001}, we plot the phase diagram of our language model with $K = 10$ and $X \to XX$.
We determined the boundary between the two phases by fitting a quadratic function via least squares regression.

\begin{figure}[t]
	\centering
	\includegraphics[scale=0.60]{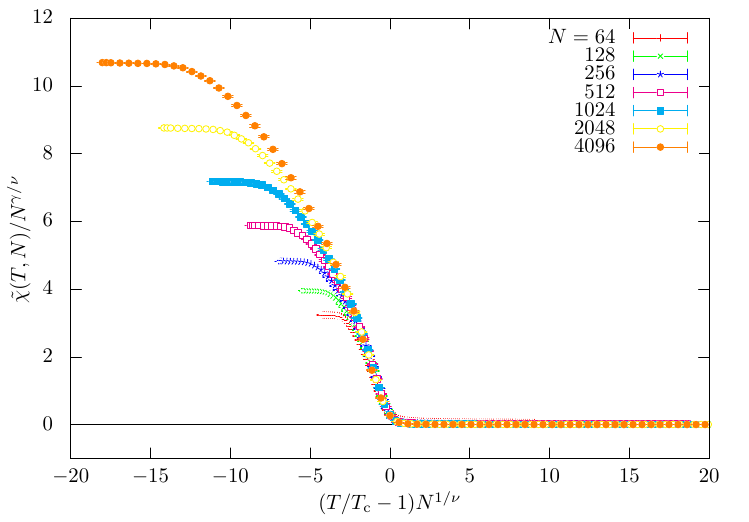}
	\includegraphics[scale=0.60]{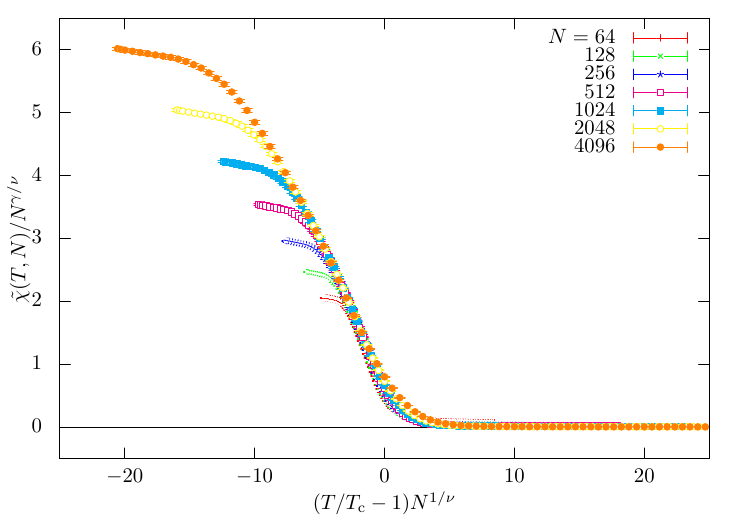}
	\caption{Finite-size scaling of $\tilde{\chi}$ at (upper) $q = 10^{-2.0}$ and (lower) $q = 10^{-1.0}$. We set $T_\mathrm{c} = 0.640$, $\nu = 2.8750$, and $\gamma = 2.05$ for $q = 10^{-2.0}$ and $T_\mathrm{c} = 0.70$, $\nu = 2.75$, and $\gamma = 2.05$ for $q = 10^{-1.0}$, where the values of $\nu$ and $\gamma$ are determined such that the scaling assumption is satisfied. We consider $X \to XX$ and set $K = 10$, $J = 1.0$, $t = 0$, $s = 0.9$, and $r_- = r_+ = 0.25$. We varied $N = 64, 128, 256, 512, 1024, 2048, 4096$.}
	\label{supp_fig_finite-size-scaling_004_001}
\end{figure}

In Fig.~\ref{supp_fig_finite-size-scaling_004_001}, we carry out finite-size scaling for $\tilde{\chi}$ at $q = 10^{-2.0}$.
We set $T_\mathrm{c} = 0.640$, $\nu = 2.8750$, and $\gamma = 2.05$, where $\nu$ and $\gamma$ are determined so that the scaling assumption holds.

\begin{figure}[t]
	\centering
	\includegraphics[scale=0.60]{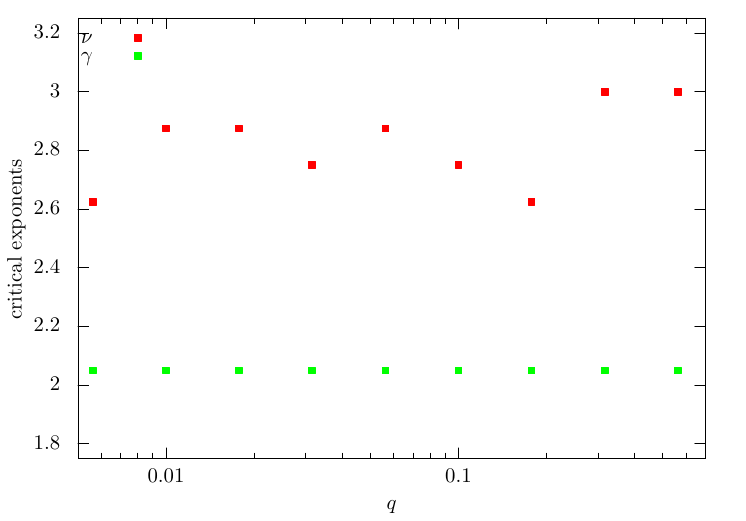}
	\caption{$q$-dependence of the critical exponents $\nu$ and $\gamma$. We consider $X \to XX$ and set $K = 10$, $J = 1.0$, $t = 0$, $s = 0.9$, and $r_- = r_+ = 0.25$.}
	\label{supp_fig_q_dependence_critical_exponents_004_001}
\end{figure}

In Fig.~\ref{supp_fig_q_dependence_critical_exponents_004_001}, we plot the $q$-dependence of the critical exponents $\nu$ and $\gamma$.
We have again observed the robust critical behavior, concluding that the BKT transition occurs.

\subsubsection{Configurations and relative frequencies for $q = 10^{-2.0}$}

\begin{figure}[t]
	\centering
	\includegraphics[scale=0.60]{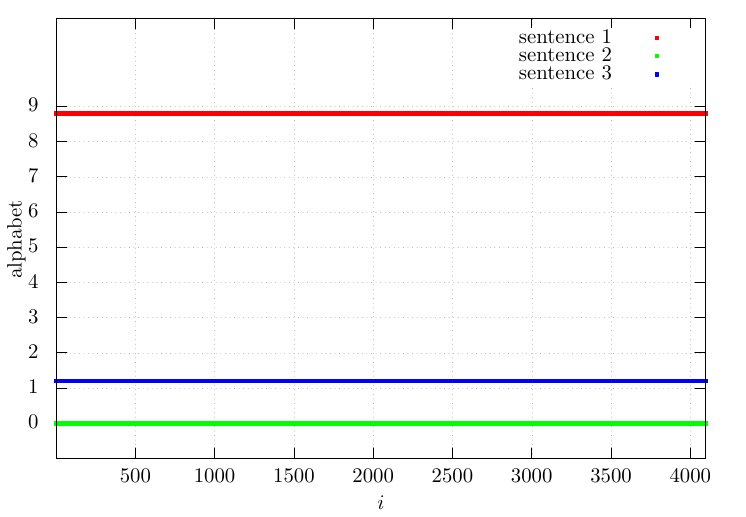}
	\includegraphics[scale=0.60]{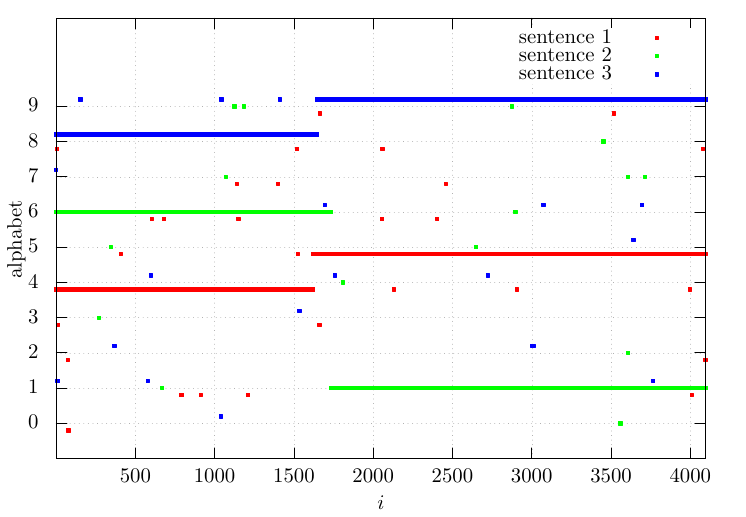}
	\caption{Typical configurations of symbols for $X \to XX$. (upper) We considered states whose magnetization lies between $0.6600$ and $0.6800$ at $k_\mathrm{B} T = 0.440$, and (lower) states whose magnetization lies between $0.9800$ and $1.0000$ at $k_\mathrm{B} T = 0.0010$. We set $K = 10$, $J = 1.0$, $q = 10^{-2.0}$, $t = 0$, $s = 0.9$, and $r_- = r_+ = 0.25$.}
	\label{supp_fig_configurations_K=10_q=10^(-2.0)_X_XX_001_001}
\end{figure}

In Fig.~\ref{supp_fig_configurations_K=10_q=10^(-2.0)_X_XX_001_001}, we plot typical configurations whose magnetization value is $0.670$ at $k_\mathrm{B} T = 0.440$ and configurations whose magnetization value is $0.990$ at $k_\mathrm{B} T = 0.0010$.
At $k_\mathrm{B} T = 0.440$, the configuration with $M=0.670$ is chosen because it corresponds to the peak of the magnetization histogram at $\langle M \rangle = 0.670$.
When the magnetization is around $0.670$, two characters out of ten are dominant and two states coexist.
As expected, the system is ordered when the magnetization is near $0.990$.

\begin{figure}[t]
	\centering
	\includegraphics[scale=0.60]{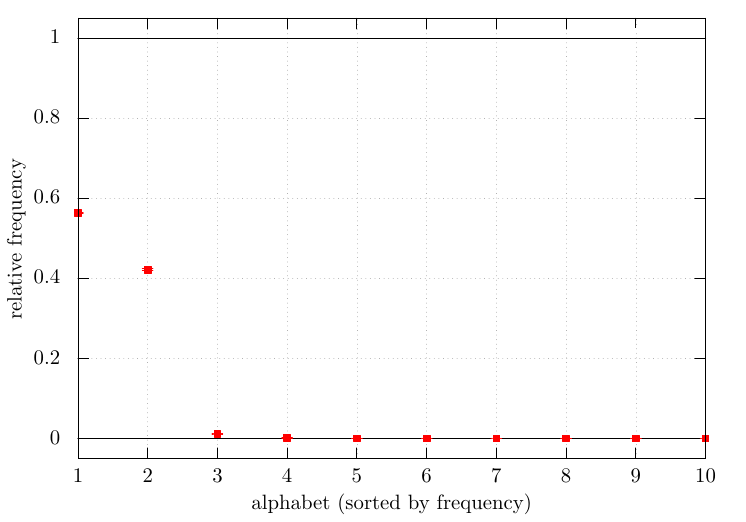}
	\includegraphics[scale=0.60]{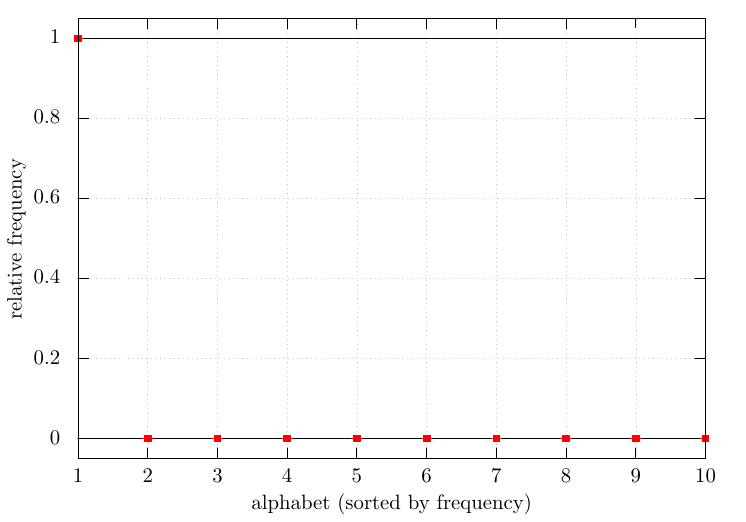}
	\caption{Relative frequencies of letters in the alphabet for $X \to XX$. (upper) We considered states whose magnetization lies between $0.6600$ and $0.6800$ at $k_\mathrm{B} T = 0.440$, and (lower) states whose magnetization lies between $0.9800$ and $1.0000$ at $k_\mathrm{B} T = 0.0010$. We set $K = 10$, $J = 1.0$, $q = 10^{-2.0}$, $t = 0$, $s = 0.9$, and $r_- = r_+ = 0.25$. Symbols on the horizontal axis are sorted by frequency and incremented by unity. The scales of the vertical and horizontal axes are linear.}
	\label{supp_fig_relative_frequencies_K=10_q=10^(-2.0)_X_XX_001_001}
\end{figure}

In Fig.~\ref{supp_fig_relative_frequencies_K=10_q=10^(-2.0)_X_XX_001_001}, relative frequencies of letters in the alphabet are shown.
When the magnetization is around $0.670$, two major characters are dominant, and other characters rarely appear in sentences.
As a result, the relative frequencies do not fit Zipf's law.
In the case where the magnetization is near $0.990$, that is, the ordered state is realized, only a single character appears in a sentence.

\clearpage

\bibliography{paper_language_model_long_999_001}

\end{document}